\documentclass[10pt,journal,compsoc]{IEEEtran}

\usepackage{times}
\usepackage{epsfig}
\usepackage{graphicx}
\usepackage{amsmath}
\usepackage{amssymb}
\usepackage{pifont}% http://ctan.org/pkg/pifont
\newcommand{\cmark}{\ding{51}}%
\newcommand{\xmark}{\ding{55}}%
\usepackage[utf8]{inputenc} % allow utf-8 input
\usepackage[T1]{fontenc}    % use 8-bit T1 fonts
\usepackage{booktabs}       % professional-quality tables
\usepackage{amsfonts}       % blackboard math symbols
\usepackage{nicefrac}       % compact symbols for 1/2, etc.
\usepackage{microtype}      % microtypography
\usepackage{color}
\usepackage{bigstrut}
\usepackage{hhline}% http://ctan.org/pkg/hhline
\usepackage{colortbl}
\usepackage{tabularx}
\usepackage{subcaption}
\usepackage{arydshln}
\usepackage{array,multirow}
\usepackage{xfrac}
\usepackage{bm}
\usepackage{soul}
\usepackage[export]{adjustbox}
\usepackage{graphicx}
\usepackage{verbatim}
\usepackage{mwe}
\usepackage[shortcuts]{extdash}
\usepackage[ruled,lined,linesnumberedhidden]{algorithm2e}
\usepackage{enumitem}
\usepackage{bbold}
\usepackage[normalem]{ulem}
\usepackage[dvipsnames]{xcolor}
\usepackage{bbm}
\usepackage{ulem}
\usepackage{float}
\usepackage{enumitem} % noindent for itemize

\newcommand{\eg}[1]{{e.g.#1}}
\newcommand{\ie}[1]{{i.e.#1}}
\newcommand{\etal}[1]{{et al.#1}}

\newcolumntype{L}[1]{>{\raggedright\let\newline\\\arraybackslash\hspace{0pt}}m{#1}}
\newcolumntype{R}[1]{>{\raggedleft\let\newline\\\arraybackslash\hspace{0pt}}m{#1}}
\newcolumntype{C}[1]{>{\centering\let\newline\\\arraybackslash\hspace{0pt}}m{#1}}
\newcommand{\rb}{\rotatebox{90}}%

\ifCLASSOPTIONcompsoc
  \usepackage[nocompress]{cite}
\else
  \usepackage{cite}
\fi

\usepackage[implicit=false]{hyperref}

\begin{document}
%\title{Learning Object-aware Semantic Correspondence}
\title{Learning Semantic Correspondence \\ Exploiting an Object-level Prior}
\author{Junghyup~Lee,~Dohyung~Kim,~Wonkyung~Lee,\\
		Jean Ponce,~\IEEEmembership{Fellow,~IEEE,}
		and~Bumsub~Ham,~\IEEEmembership{Member,~IEEE}        
\IEEEcompsocitemizethanks{
\IEEEcompsocthanksitem J. Lee,~D. Kim,~W. Lee,~and B. Ham are with the School of Electrical and Electronic Engineering, Yonsei University, Seoul 03722, Korea.\protect\\
E-mail: \{junghyup.lee, do.hyung, wonkyung.lee, bumsub.ham\}@yonsei.ac.kr% <-this % stops an unwanted space
\IEEEcompsocthanksitem J. Ponce is with Inria and DI-ENS, D{\'e}partement d'Informatique de l'ENS, CNRS, PSL University, Paris 75005, France. \protect \\E-mail: jean.ponce@inria.fr}% <-this % stops an unwanted space
\thanks{The first two authors contributed equally. Corresponding author: Bumsub Ham.}
%\thanks{(Corresponding author: Bumsub Ham)}
}

\markboth{SUBMISSION TO IEEE TRANSACTIONS ON PATTERN ANALYSIS AND MACHINE INTELLIGENCE, 2019}%
{SUBMISSION TO IEEE TRANSACTIONS ON PATTERN ANALYSIS AND MACHINE INTELLIGENCE, 2019}

\IEEEtitleabstractindextext{%
\begin{abstract}
We address the problem of semantic correspondence, that is, establishing a dense flow field between images depicting different instances of the same object or scene category. We propose to use images annotated with binary foreground masks and subjected to synthetic geometric deformations to train a convolutional neural network (CNN) for this task. Using these masks as part of the supervisory signal provides an object-level prior for the semantic correspondence task and offers a good compromise between semantic flow methods, where the amount of training data is limited by the cost of manually selecting point correspondences, and semantic alignment ones, where the regression of a single global geometric transformation between images may be sensitive to image-specific details such as background clutter. We propose a new CNN architecture, dubbed SFNet, which implements this idea. It leverages a new and differentiable version of the argmax function for end-to-end training, with a loss that combines mask and flow consistency with smoothness terms. Experimental results demonstrate the effectiveness of our approach, which significantly outperforms the state of the art on standard benchmarks.
\end{abstract}

\begin{IEEEkeywords}
Semantic correspondence, object-level prior, differentiable argmax function.
\end{IEEEkeywords}}

\maketitle

\IEEEdisplaynontitleabstractindextext

% \fi
\IEEEpeerreviewmaketitle

\IEEEraisesectionheading{\section{Introduction}\label{sec:introduction}}
\IEEEPARstart{E}{stablishing} dense correspondences across images is one of the fundamental tasks in computer vision~\cite{okutomi1993multiple,brox2004high,liu2011sift}. Early works have focused on handling different views of the same scene~(stereo matching~\cite{okutomi1993multiple,hosni2013fast}) or successive frames~(optical flow~\cite{brox2004high,brox2009large}) in a video sequence. Semantic correspondence algorithms~(\eg,~SIFT Flow~\cite{liu2011sift}) go one step further, finding a dense flow field between images depicting different instances of the same object or scene category, which has proven useful in various computer vision tasks including object recognition~\cite{liu2011sift,duchenne2011graph}, semantic segmentation~\cite{kim2013deformable}, co-segmentation~\cite{taniai2016joint}, image editing~\cite{dale2009image}, and scene parsing~\cite{kim2013deformable,zhou2015flowweb}. Establishing dense semantic correspondences is very challenging especially in the presence of large changes in appearance or scene layout and background clutter. Classical approaches to semantic correspondence~\cite{liu2011sift,kim2013deformable,hur2015generalized,bristow2015dense,yang2014daisy} typically use an objective function involving fidelity and regularization terms. The fidelity term encourages hand-crafted features~(\eg,~SIFT~\cite{lowe2004distinctive}, HOG~\cite{dalal2005histograms}, DAISY~\cite{tola2010daisy}) to be matched along a dense flow field between images, and the regularization term makes it smooth while aligning discontinuities to object boundaries. Hand-crafted features, however, do not capture high-level semantics~(\eg,~appearance and shape variations), and they are not robust to image-specific details~(\eg,~texture, background clutter, occlusion). 

Convolutional neural networks~(CNNs) have allowed remarkable advances in semantic correspondence in the past few years. Recent methods using CNNs~\cite{han2017scnet,choy2016universal,novotny2017anchornet,kanazawa2016warpnet,kim2017fcss,zhou2016learning,rocco2017convolutional,rocco2018end,Seo2018AttentiveSA,Jeon2018PARN} benefit from rich semantic features invariant to intra-class variations, achieving state-of-the-art results. Semantic flow approaches~\cite{han2017scnet,choy2016universal,novotny2017anchornet,kim2017fcss,zhou2016learning} attempt to find correspondences for individual pixels or patches. They are not seriously affected by non-rigid deformations, but are easily distracted by background clutter. They also require a large amount of data with ground-truth correspondences for training. Although pixel-level semantic correspondences impose very strong constraints, manually annotating them is extremely labor-intensive and somewhat subjective, which limits the amount of training data available~\cite{ham2016proposal}. An alternative is to learn feature descriptors only~\cite{choy2016universal,novotny2017anchornet,kim2017fcss} or to exploit 3D CAD models together with rendering engines~\cite{zhou2016learning}. 
Semantic alignment methods~\cite{kanazawa2016warpnet,rocco2017convolutional,rocco2018end,Seo2018AttentiveSA,Jeon2018PARN}, on the other hand, formulate semantic correspondence as a geometric alignment problem and directly regress parameters of a global transformation model~(\eg,~affine deformation or thin plate spline) between images. They leverage self-supervised learning where ground-truth parameters are generated synthetically using random transformations with, however, a higher sensitivity to non-rigid deformations. Moreover, background clutter prevents focusing on individual objects and interferes with the estimation of the transformation parameters. To overcome this problem, recent methods reduce the influence of distractors by inlier counting~\cite{rocco2018end} or an attention process~\cite{Seo2018AttentiveSA}.

\begin{figure}[t]
\begin{center}
   \includegraphics[width=\columnwidth]{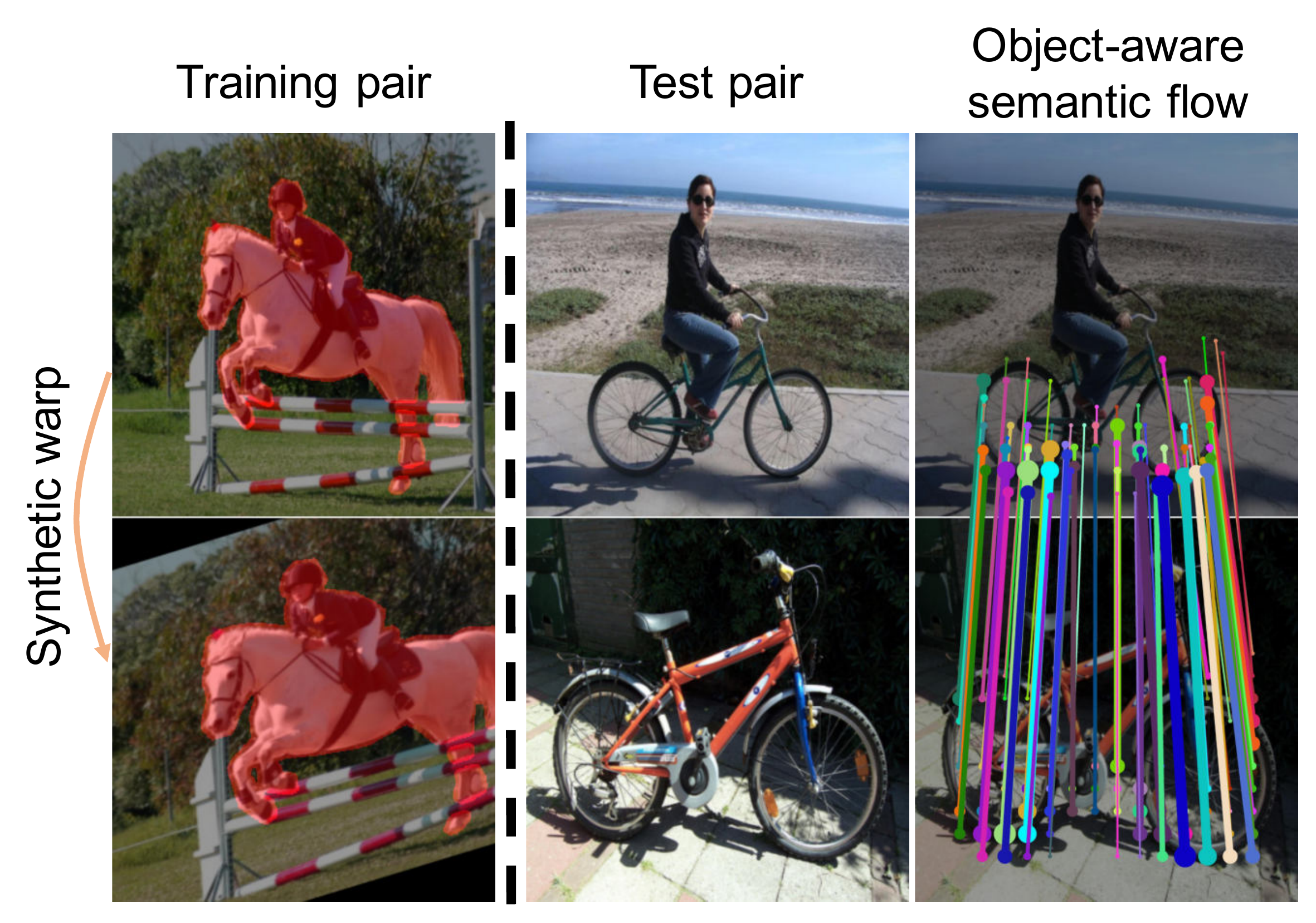}
\end{center}
   \caption{We use pairs of warped foreground masks obtained from a single image~(left) as a supervisory signal to train our model. This allows us to establish object-aware semantic correspondences across images depicting different instances of the same object or scene category~(right). No masks are required at test time.~(Best viewed in color.)}
\label{fig:teaser}
\end{figure}

In this paper, we present a new approach to establishing an object-aware semantic flow and propose to exploit binary foreground masks as a supervisory signal during training~(Fig.~\ref{fig:teaser}). Our approach builds upon the insight that correspondences of high quality between images allow to segment common objects from background. To implement this idea, we introduce a new CNN architecture, dubbed SFNet, that outputs a semantic flow field at a sub-pixel level. We leverage a new and differentiable version of the argmax function, the kernel soft argmax, together with mask/flow consistency and smoothness terms to train SFNet end to end, establishing object-aware correspondences while filtering out distracting details. Our approach has the following advantages: First, it is a good compromise between current semantic flow and alignment methods, since foreground masks are available for large datasets, and they provide an object-level prior for the semantic correspondence task. Exploiting these masks during training makes it possible to focus on learning correspondences between prominent objects and scene elements (masks are of course not used at test time). Second, our method establishes a dense non-parametric flow field~(\ie,~semantic flow), which is more robust to non-rigid deformations than parametric regression~(\ie,~semantic alignment). Finally, using the kernel soft argmax allows us to train the whole network end to end, and hence our approach further benefits from high-level semantics specific to the task of semantic correspondence.
The main contributions of this paper can be summarized as follows:
\begin{itemize}[leftmargin=*]
\item[$\bullet$] We propose to exploit binary foreground masks, that are widely available and can be annotated more easily than individual point correspondences, to learn semantic flow by incorporating an object-level prior in the learning task.
\item[$\bullet$] We introduce a kernel soft argmax function, making our model quite robust to multi-modal distributions while providing a differentiable flow field at a sub-pixel level.
\item[$\bullet$] We set a new state of the art on standard benchmarks for semantic correspondence, mask transfer, pose keypoint propagation, and object co-segmentation, clearly demonstrating the effectiveness of our approach. We also provide an extensive experimental analysis with ablation studies.
\end{itemize}

A preliminary version of this work appeared in~\cite{lee2019sfnet}. This version adds (1) a detailed description of related works exploiting object priors for semantic correspondence; (2) an in-depth presentation of SFNet including the kernel soft argmax and loss terms; (3) more comparisons with the state of the art on different benchmarks including the TSS~\cite{taniai2016joint} and recent SPair-71k~\cite{min2019hyperpixel} datasets; (4) an evaluation on the tasks of pose keypoint propagation and object co-segmentation with the JHMDB~\cite{jhuang2013towards} and TSS~\cite{taniai2016joint} datasets, respectively; and (5) an extensive experimental evaluation including a runtime comparison and a performance analysis on SFNet trained using noisy labels~(\ie,~bounding boxes) or with different dataset. To encourage comparison and future work, our code and model are available online:~\url{https://cvlab.yonsei.ac.kr/projects/SFNet}.

%\todo{A preliminary version of this work published in ~\cite{lee2019sfnet}. This version adds (1) an in-depth presentation of SFNet including a detailed description of the kernel soft argmax and loss terms; (2) an additional comparative study with the state-of-the-art methods using different datasets (TSS~\cite{taniai2016joint}, SPair-71k~\cite{}); (3) a further study on video propagation task using JHMDB dataset~\cite{jhuang2013towards} d; (4) an extensive experimental evaluation including performance analysis with the model trained with noisy labels (VOC 2012 bounding box~\cite{everingham2010pascal}) or larger datasets (MS COCO dataset~\cite{lin2014microsoft})} To encourage comparison and future work, our code and models are available online: \todo{https://cvlab-yonsei.github.io/projects/SFNet}.

\section{Related Work}
\label{sec:related}
Correspondence problems cover a broad range of topics in computer vision including stereo, motion analysis, object recognition and shape matching. Giving a comprehensive review on these topics is beyond the scope of this paper. We thus focus on representative works related to ours.

\subsection{Semantic Flow}
Classical approaches focus on finding sparse correspondences,~\eg,~for instance matching~\cite{lowe2004distinctive}, or establishing dense matches between nearby views of the same scene/object,~\eg,~for stereo matching~\cite{hosni2013fast,okutomi1993multiple} and optical flow estimation~\cite{brox2009large,brox2004high}. Unlike these, semantic correspondence methods estimate dense matches across pictures containing different instances of the same object or scene category. Early works on semantic correspondence focus on matching local features from hand-crafted descriptors, such as SIFT~\cite{liu2011sift,kim2013deformable,hur2015generalized,bristow2015dense}, DAISY~\cite{yang2014daisy} and HOG~\cite{ham2016proposal,taniai2016joint,yang2017object}, together with spatial regularization using graphical models~\cite{liu2011sift,kim2013deformable,taniai2016joint,hur2015generalized} or random sampling~\cite{barnes2009patchmatch,yang2014daisy}.
However, hand-crafting features capturing high-level semantics is extremely hard, and similarities between them are easily distracted,~\eg,~by clutter, texture, occlusion and appearance variations. There have been many attempts to estimate correspondences robust to background clutter or scale changes between objects/object parts. These use object proposals as candidate regions for matching~\cite{ham2016proposal,yang2017object} or perform matching in scale space~\cite{qiu2014scale}.

Recently, image features from CNNs have demonstrated a capacity to both representing high-level semantics and being robust to appearance and shape variations~\cite{krizhevsky2012imagenet,simonyan2014very,he2016deep}. Long~\etal~\cite{long2014convnets} apply CNNs to establish semantic correspondences between images. They follow the same procedure as the SIFT Flow~\cite{liu2011sift} method, but exploit off-the-shelf CNN features trained for the ImageNet classification task~\cite{deng2009imagenet} due to a lack of training datasets with pixel-level annotations. This problem can be alleviated by synthesizing ground-truth correspondences from 3D models~\cite{zhou2016learning} or augmenting the number of match pairs in a sparse keypoint dataset using interpolation~\cite{taniai2016joint}. More recently, new benchmarks for semantic correspondence have been released. PF-PASCAL~\cite{ham2017proposal} provides 1300+ image pairs of 20 image categories with ground-truth annotations from the PASCAL 2011 keypoint dataset~\cite{BourdevMalikICCV09}. SPair-71k~\cite{min2019hyperpixel} consists of over 70k of image pairs from PASCAL 3D+~\cite{xiang2014beyond} and PASCAL VOC 2012~\cite{everingham2010pascal} with rich annotations including keypoints, segmentation masks and bounding boxes. These enable learning local features~\cite{han2017scnet,kim2017fcss,rocco2018neighbourhood,min2019hyperpixel} specific to the task of semantic correspondence. FCSS~\cite{kim2017fcss} introduces a learnable local self-similarity descriptor robust to intra-class variations. SCNet~\cite{han2017scnet} and HPF~\cite{min2019hyperpixel} present region descriptors exploiting geometric consistency among object parts. NCN~\cite{rocco2018neighbourhood} analyzes neighborhood consensus patterns in the 4D space of all possible correspondences in order to find spatially consistent matches, disambiguating feature matches on repetitive patterns. Although these approaches using CNN features outperform early methods by large margins, the loss functions they use for training typically do not involve a spatial regularizer mainly due to a lack of differentiability of the flow field. In contrast, our flow field is differentiable, allowing us to train the whole network end to end with a spatial regularizer.

%\todo{FCSS~\cite{kim2017fcss} formulates a local self-similarity within a fully convolutional network, providing robustness to intra-class variations. SCNet~\cite{han2017scnet} presents a trainable region descriptor incorporated with geometric consistency for computing geometrically plausible semantic correspondence. In addition to local feature learning, analyzing neighbourhood consensus patterns in 4D space of all possible correspondences~\cite{rocco2018neighbourhood} further disambiguates a match on a repetitive pattern.}

\begin{figure*} % Overview figure
\centering
\includegraphics[width=0.85\textwidth]{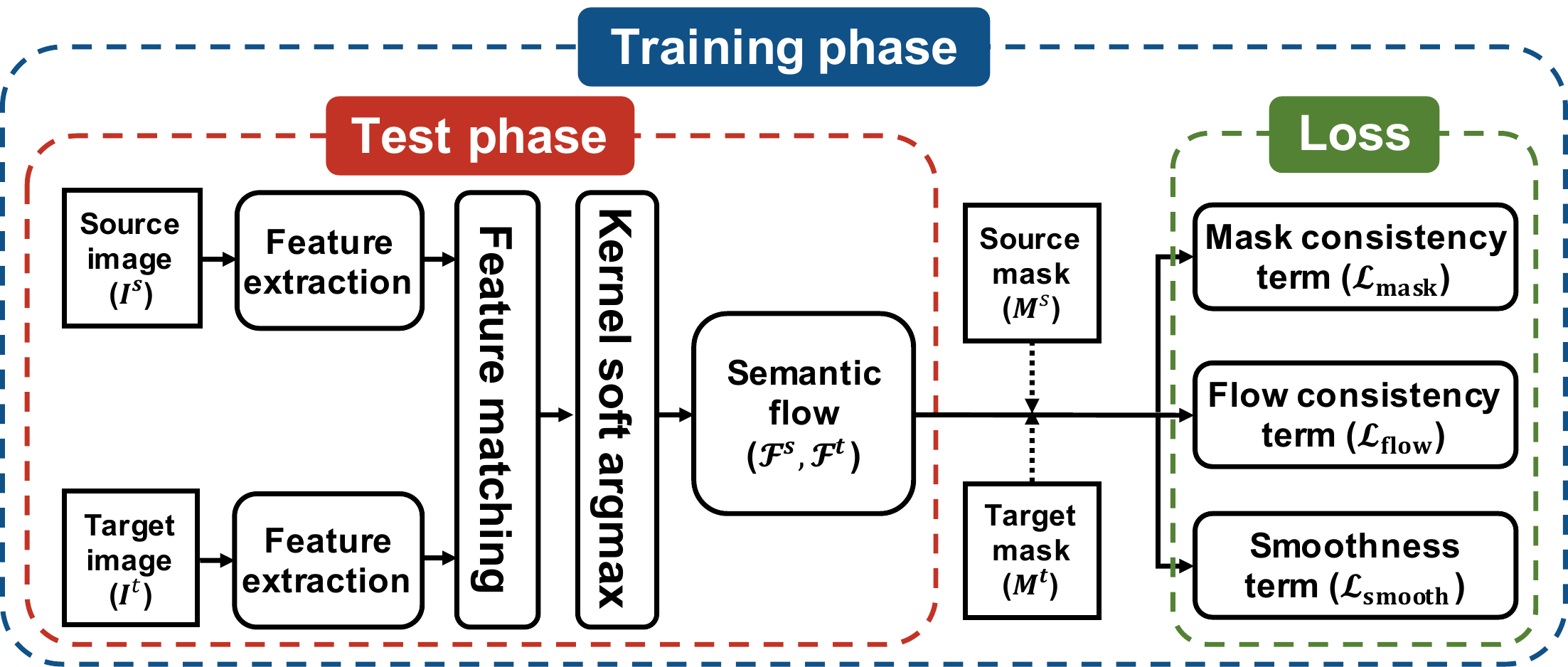} 
\caption{Overview of SFNet. SFNet takes an input pair of source and target images, and extracts local features using a siamese network. It then computes pairwise matching scores between features and establishes semantic flow for source and target images using the kernel soft argmax. At training time, corresponding foreground masks for the two images are used to compute mask consistency, flow consistency, and smoothness terms. See text for details.}
\label{fig:proposed-met}
\end{figure*}

\subsection{Semantic Alignment}
Several recent methods~\cite{kanazawa2016warpnet,rocco2017convolutional,rocco2018end,Seo2018AttentiveSA,Jeon2018PARN} formulate semantic correspondence as a geometric alignment problem using parametric models. In particular, these methods first compute feature correlations between images. The feature correlations are then fed into a regression layer to estimate parameters of a global transformation model~(\eg,~affine, homography, and thin plate spline) to align images. This makes it possible to leverage self-supervised learning~\cite{kanazawa2016warpnet,rocco2017convolutional,rocco2018end,Seo2018AttentiveSA} using synthetically-generated data, and to train the entire CNNs end to end. These approaches apply the same transformation to all pixels, which has the effect of an implicit spatial regularization, providing smooth matches and often outperforming semantic flow methods~\cite{choy2016universal,ham2016proposal,han2017scnet,kim2017fcss,zhou2016learning}. However, they are easily distracted by background clutter and occlusion~\cite{kanazawa2016warpnet,rocco2017convolutional}, since correlations between pairs of features are noisy and include outliers~(\eg,~between different backgrounds). Although this can be alleviated by using attention models~\cite{Seo2018AttentiveSA} or suppressing outlier matches~\cite{rocco2018end}, global transformation models are highly sensitive to non-rigid deformations or local geometric variations. Alternative approaches include estimating local transformation models in a coarse-to-fine scheme~\cite{Jeon2018PARN} or applying the geometric transformation recursively~\cite{kim2018recurrent}, but they are computationally expensive. In contrast, our method avoids the problem efficiently by establishing semantic correspondences directly from feature correlations.

%\todo{Estimating locally varying affine transformation fields across images in a coarse-to-fine scheme~\cite{Jeon2018PARN} or recursively computing the geometric transformations~\cite{kim2018recurrent} avoids this problem, but they are computationally expensive. Unlike them, our method efficiently addresses this problem by establishing semantic correspondences directly from feature correlation rather than the coarse-to-fine or the iterative process.}
 
\subsection{An Object-level Prior for Semantic Correspondence}
Several methods~\cite{ham2016proposal,han2017scnet,yang2017object,Jeon2018PARN,kim2017fcss,zhou2015flowweb,zhou2016learning} leverage object priors~(\eg,~object proposals, bounding boxes or foreground masks) to learn semantic correspondence. Proposal flow~\cite{ham2016proposal} and its CNN version~\cite{han2017scnet} use object proposals as matching primitives, and consider appearance and geometric consistency constraints to establish region correspondences. OADSC~\cite{yang2017object} also exploits object proposals, but leverages hierarchical graphs built on the proposals in a coarse-to-fine manner, allowing pixel-level correspondences. Similar to ours, other methods leverage bounding boxes or foreground masks for semantic correspondence. They, however, do not incorporate the object location prior explicitly into loss functions, and use the prior for pre-processing training samples instead. For example, PARN~\cite{Jeon2018PARN} and FCSS~\cite{kim2017fcss} use bounding boxes or foreground masks to generate positive/negative matches within object regions at training time. In~\cite{zhou2015flowweb,zhou2016learning}, bounding boxes are used to limit the candidate regions for matching at both training and test time. Contrary to these methods, we incorporate this prior~(\eg,~bounding boxes or foreground masks) directly into the 	loss functions to train the network, and outperform the state of the art by a significant margin.

\section{Approach}
In this section, we describe our approach to establishing object-aware semantic correspondences including the network architecture~(Section~\ref{sec:archi}) and loss functions~(Section~\ref{sec:loss}). An overview of our method is shown in Fig.~\ref{fig:proposed-met}.

\subsection{Network Architecture}
\label{sec:archi}
Our model consists of three main parts~(Fig.~\ref{fig:proposed-met}): We first extract features from source and target images, $I^s$~and~$I^t$, respectively, using a fully convolutional siamese network, where each of the two branches has the same structure with shared parameters. We then compute matching scores between all pairs of local features in the two images, and assign the best match for each feature using the kernel soft argmax function defined in Sec~\ref{sec:kernel}. All components are differentiable, allowing us to train the whole network end to end. In the following, we describe the network architecture for source to target matching in detail. A target to source match is computed in the same manner.

\begin{figure*}[ht] % kernel soft argmax figure
\begin{center}
   \includegraphics[width=0.95\textwidth]{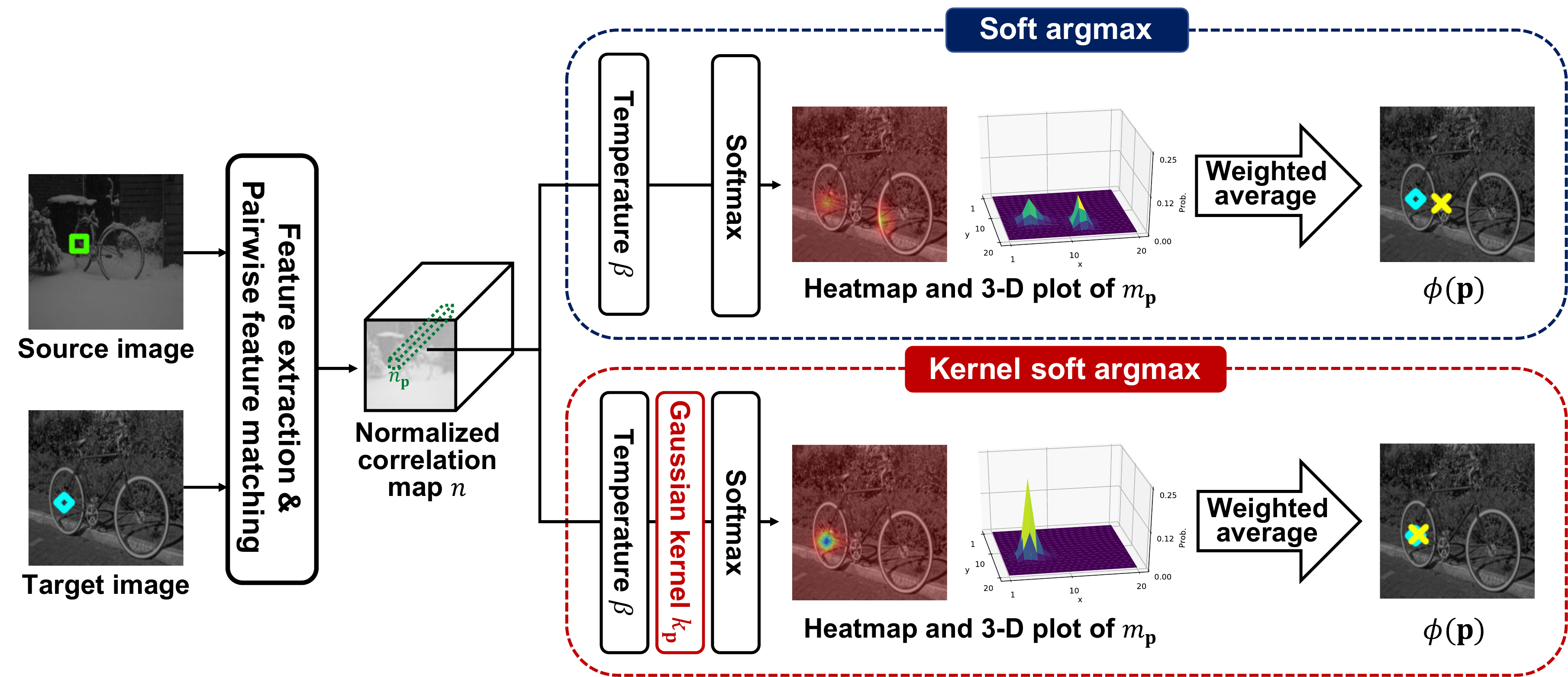}
\end{center}
  \caption{Visualization of soft and kernel soft argmax operations. A point in the source image and its ground-truth correspondence in the target image are shown as the square and the diamond, respectively. A matching point computed by either the soft or kernel soft argmax operators is shown as the cross. When multiple features are highly correlated, the soft argmax often gives incorrect matches. The kernel soft argmax avoids this problem while maintaining differentiability.~(Best viewed in color.)}
\label{fig:KS_argmax}
\end{figure*}

\subsubsection{Feature Extraction}
For feature extraction, we exploit a ResNet-101~\cite{he2016deep} pretrained for the ImageNet classification task~\cite{deng2009imagenet}. Although such CNN features give rich semantics, they typically fire on highly discriminative parts for classification~\cite{zhou2016cam}. This may be less adequate for feature matching that requires capturing a spatial deformation for fine-grained localization. We thus use additional adaptation layers to extract features specific to the task of semantic correspondence, making them highly discriminative w.r.t both appearance and spatial context. This gives a feature map of size $h \times w \times d$ for each image that corresponds to $h \times w$ grids of $d$-dimensional local features. We then apply L2 normalization to the individual $d$-dimensional features. As will be seen in our experiments, the adaptation layers boost the matching performance drastically.

\subsubsection{Feature Matching}
Matching scores are computed as the dot product between local features, resulting in a 4-dimensional correlation map of size~$h \times w \times h \times w$ as follows:
\begin{equation}
c({\bf{p}},{\bf{q}}) = f^s({\bf{p}})^\top f^t({\bf{q}}),
\label{eq:correlation_map}
\end{equation}
where we denote by $f^s({\bf{p}})$ and $f^t({\bf{q}})$~$d$-dimensional features at positions ${\bf{p}}=(p_x, p_y)$ and ${\bf{q}}=(q_x, q_y)$ in the source and target images, respectively. 

\begin{figure*}[t]
	\captionsetup[subfigure]{aboveskip=-0.5pt,belowskip=-0.5pt}
	\centering
	\begin{tabular}{cc}
		\begin{tabular}{c}
			\begin{subfigure}{0.14\linewidth}
			\includegraphics[width=\linewidth, frame]{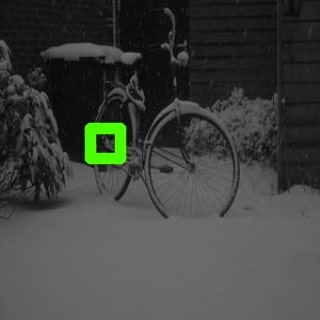}
			\vspace{-0.3cm}
			\caption*{Source image.}
			\vspace{0.1cm}
			\end{subfigure} \\
			\begin{subfigure}{0.14\linewidth}
			\includegraphics[width=\linewidth, frame]{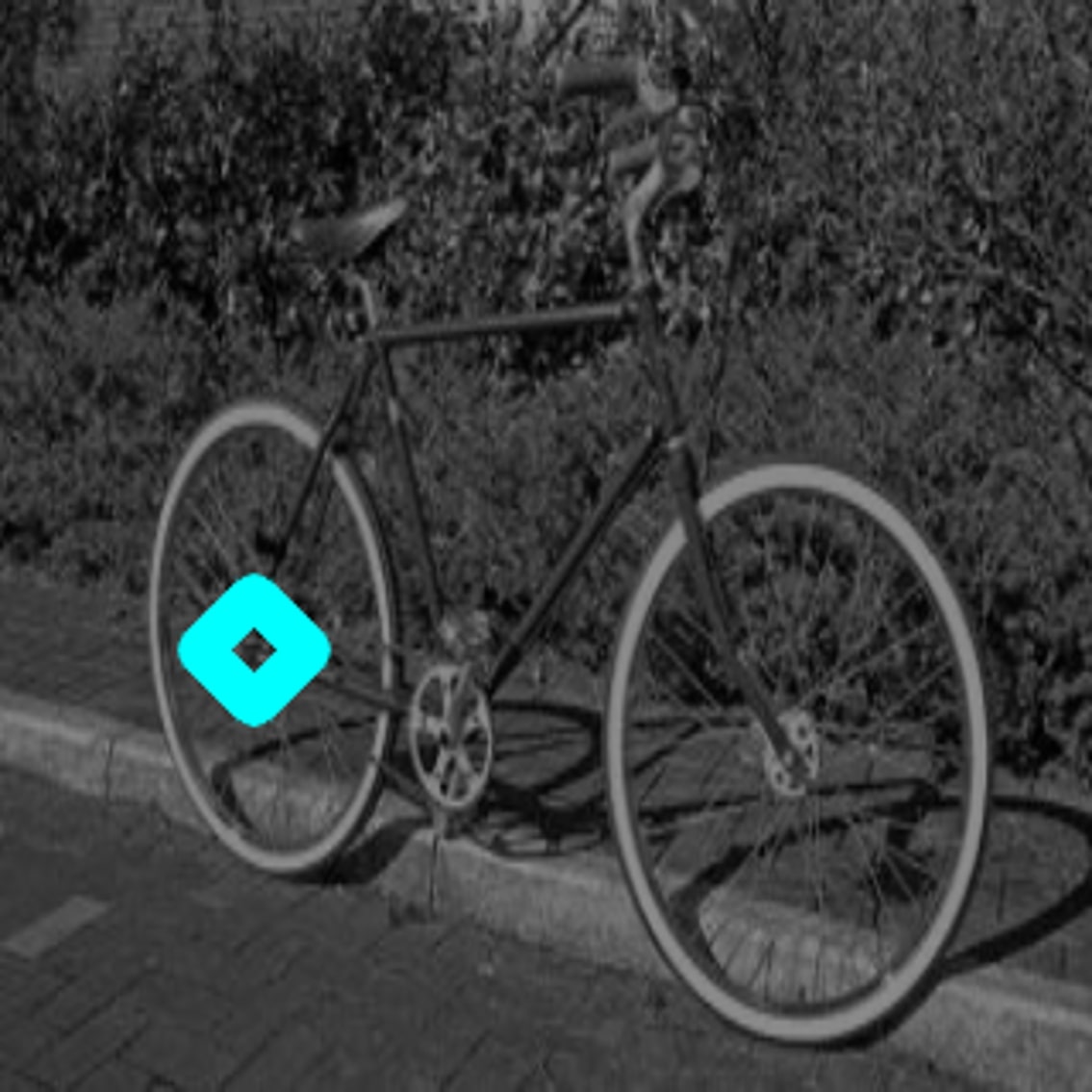}
			\vspace{-0.3cm}
			\caption*{Target image.}
			\end{subfigure} \\
		\end{tabular}
		\vline
		\hspace{0.05cm}
		\begin{tabular}{cccc}
			\adjustbox{valign=m,vspace=1.6pt}{\includegraphics[width=.14\linewidth, frame]{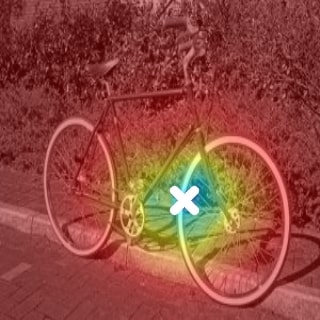}} &
			\adjustbox{valign=m,vspace=1.6pt}{\includegraphics[width=.14\linewidth, frame]{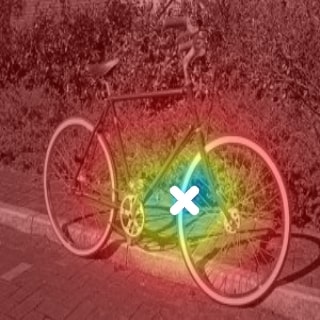}} &
			\adjustbox{valign=m,vspace=1.6pt}{\includegraphics[width=.14\linewidth, frame]{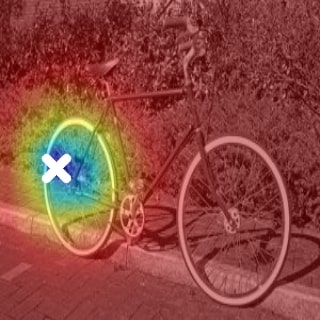}} &
			\adjustbox{valign=m,vspace=1.6pt}{\includegraphics[width=.14\linewidth, frame]{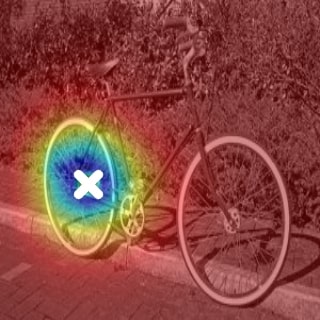}} \\

			\adjustbox{valign=m,vspace=1.6pt}{\includegraphics[width=.14\linewidth]{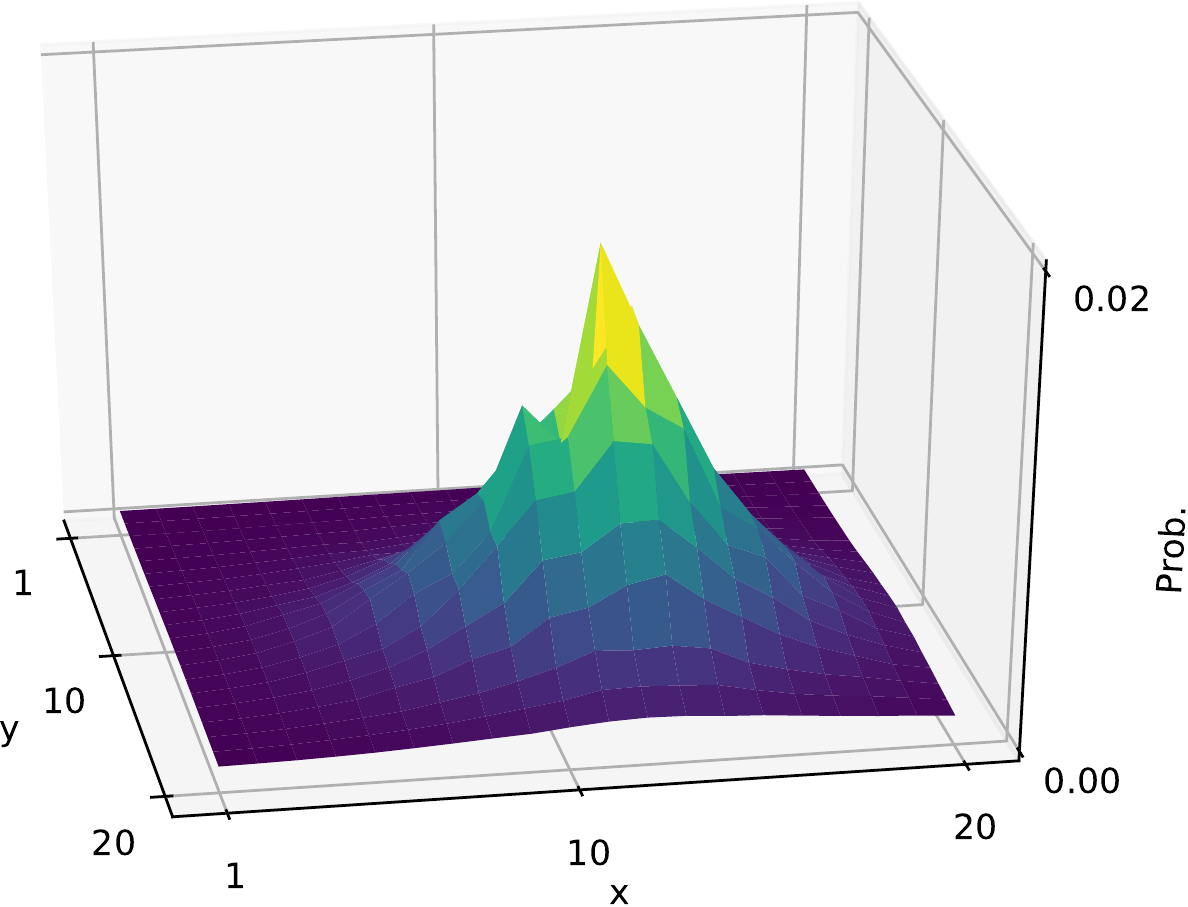}} & 
			\adjustbox{valign=m,vspace=1.6pt}{\includegraphics[width=.14\linewidth]{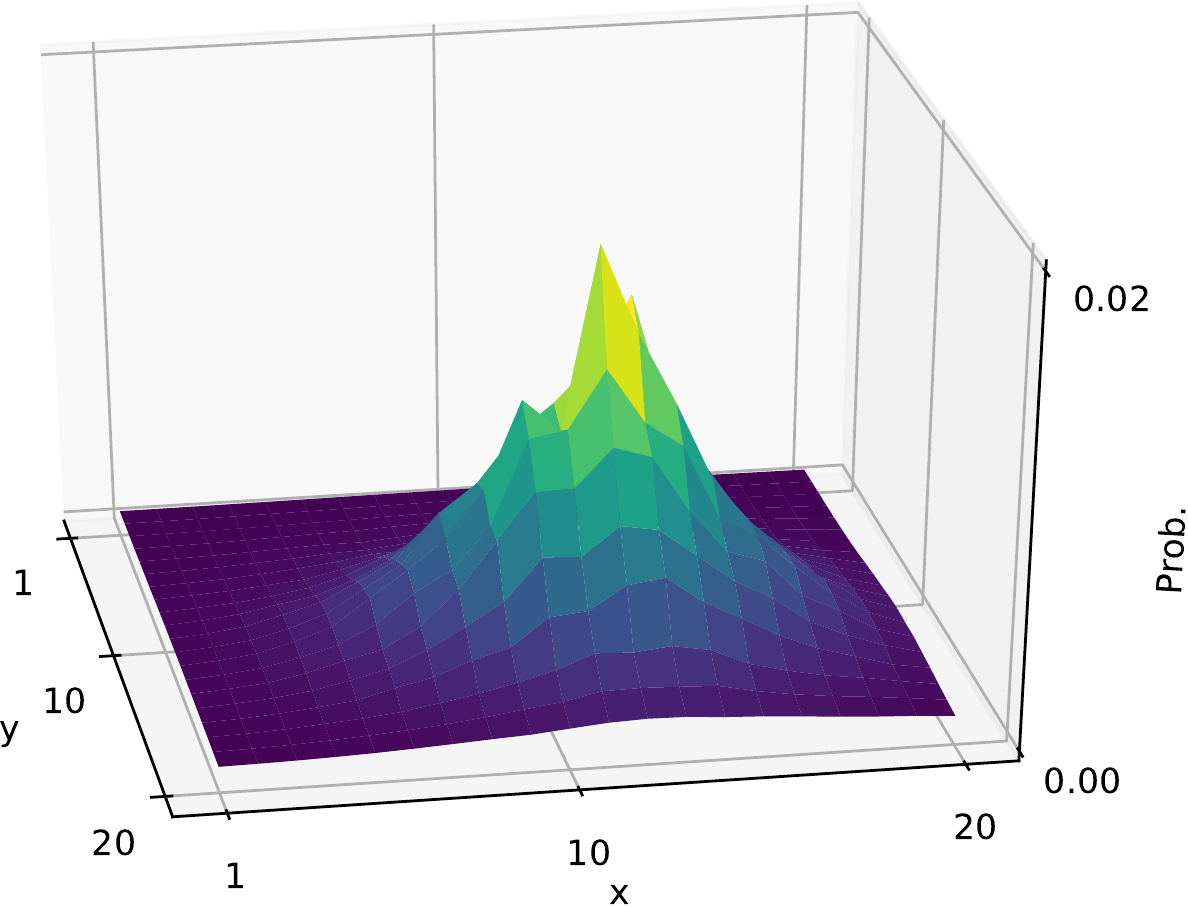}} &
			\adjustbox{valign=m,vspace=1.6pt}{\includegraphics[width=.14\linewidth]{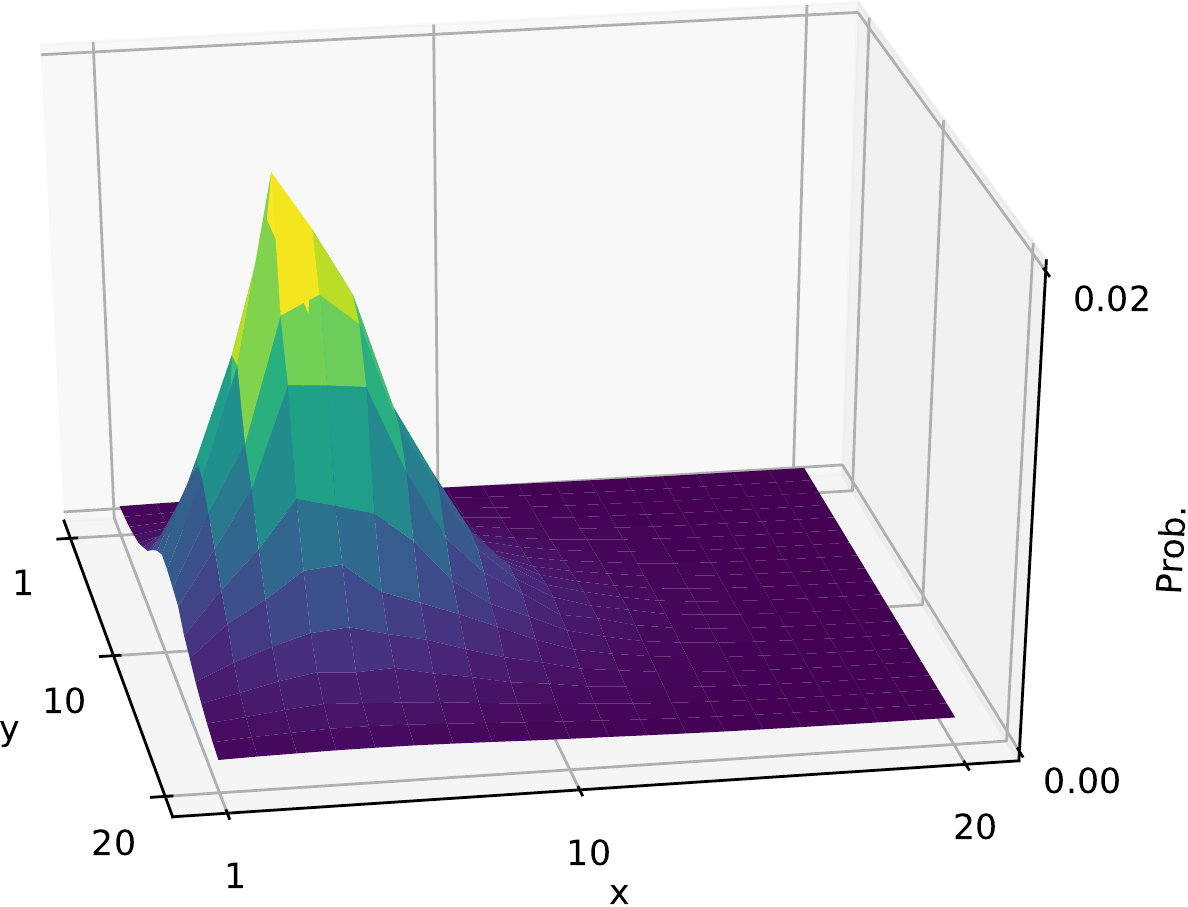}} &
			\adjustbox{valign=m,vspace=1.6pt}{\includegraphics[width=.14\linewidth]{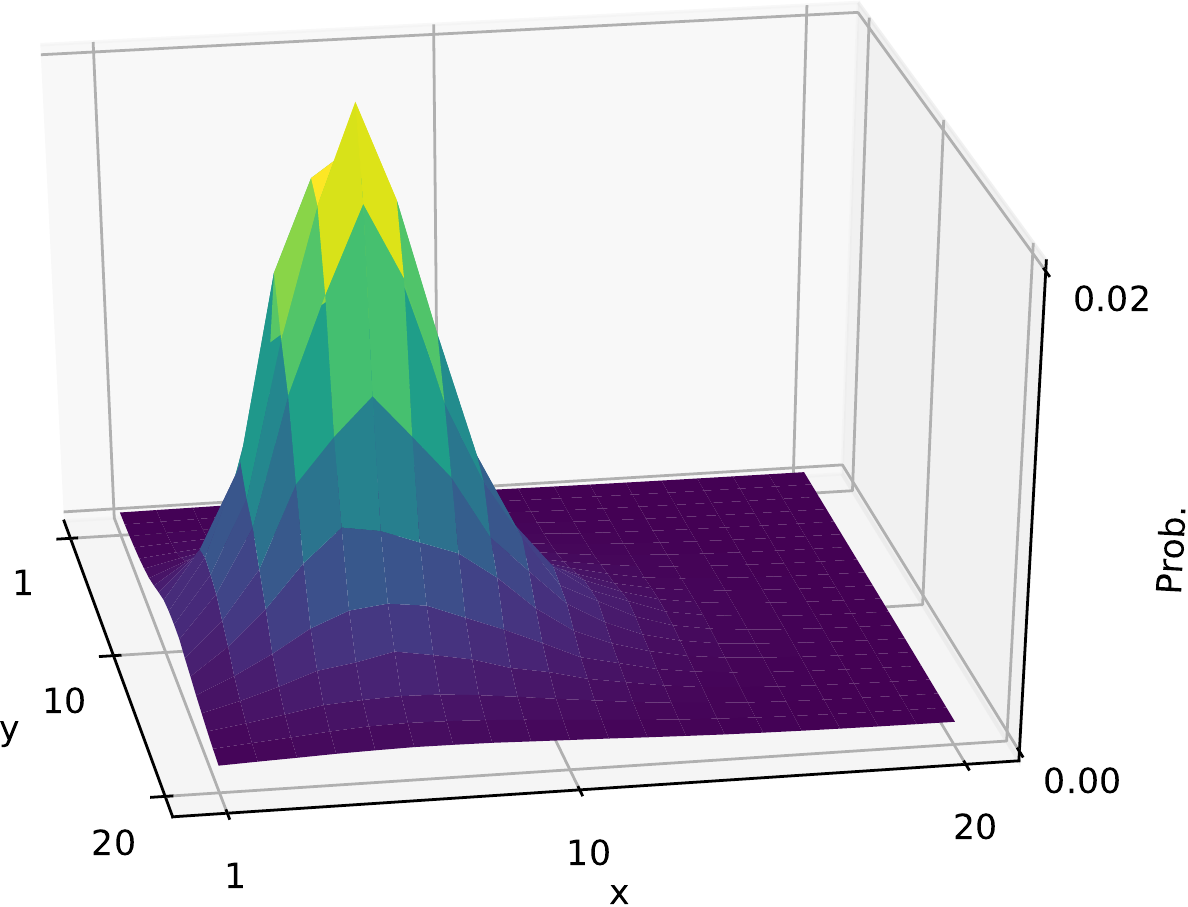}} \\
			
			Initial state. & Iteration 1. & Iteration 2. & Iteration 4.
		\end{tabular}
	\end{tabular}
	\vspace{-0.2cm}
	\caption{Visualization of matching probabilities~$m_{\bf{p}}$ and shifts of a dominant mode during training. A point in the source and its ground-truth correspondence in the target are shown as the square and diamond, respectively. The modes selected by the discrete argmax are shown as the cross. We can see that the dominant mode shifts toward a correct match after the second iteration.}
	\label{fig:incorrect_mode}	
	\vspace{-0.1cm}
\end{figure*}

\subsubsection{Kernel Soft Argmax Layer} 
\label{sec:kernel}
We could assign the best matches by applying the argmax function over a 2-dimensional correlation map~$c_{\bf{p}}({\bf{q}})=c({\bf{p}},{\bf{q}})$,~w.r.t all features~$f^t({\bf{q}})$~at each spatial location~${\bf{p}}$,~\ie,~$\operatorname*{argmax}_{{\bf{q}}} c_{\bf{p}}({\bf{q}})$. However, argmax is not differentiable. The soft argmax function~\cite{honari2018improving,kendall2017end} computes an output by a weighted average of all spatial positions with corresponding matching probabilities~(\ie,~an expected value of all spatial coordinates weighted by corresponding probabilities). Although it is differentiable and enables fine-grained localization at a sub-pixel level, its output is influenced by all spatial positions, which is problematic especially in the case of multi-modal distributions~(Fig.~\ref{fig:KS_argmax}). In other words, the soft argmax best approximates the discrete argmax when the matching probability is uni-modal having one clear peak.

We introduce a hybrid version, the \emph{kernel soft argmax}, that takes advantage of both the soft and discrete argmax. Concretely, it computes correspondences~$\phi({\bf{p}})$ for individual locations~${\bf{p}}$ as an average of all coordinate pairs~${\bf{q}}=(q_x,q_y)$ weighted by matching probabilities~$m_{\bf{p}}({\bf{q}})$ as follows. 
\begin{equation}\label{eq:kernel_soft_max}
	\phi({\bf{p}}) = \sum_{{\bf{q}}}  m_{\bf{p}}({\bf{q}})  {\bf{q}}.
\end{equation}
The matching probability~$m_{\bf{p}}$ is computed by applying a spatial softmax function to a L2-normalized version~$n_{\bf{p}}$ of the correlation map~$c_{\bf{p}}$: 
\begin{equation}\label{eq:matching_prop}
	m_{\bf{p}}({\bf{q}}) = \frac{\exp (\beta k_{\bf{p}}({\bf{q}}) n_{\bf{p}}({\bf{q}}))}{\sum_{{\bf{q}}^\prime \in n_{\bf{p}} } \exp (\beta k_{\bf{p}}({\bf{q}}^\prime) n_{\bf{p}}({\bf{q}}^\prime)) }.
\end{equation}
We perform L2 normalization on the 2-dimensional correlation map~$c_{\bf{p}}$, adjusting the matching scores~$f^s({\bf{p}})^\top f^t({\bf{q}})$ to a common scale before applying the softmax function. $\beta$~is a ``temperature'' parameter adjusting the distribution of the softmax output. As the temperature parameter~$\beta$ increases, the softmax function approaches the discrete one with one clear peak, but this may cause an unstable gradient flow at training time. $k_{\bf{p}}$ is a 2-dimensional Gaussian kernel centered on the position obtained by applying the discrete argmax to the correlation map,~\ie,~$\operatorname*{argmax}_{{\bf{q}}} n_{\bf{p}}({\bf{q}})$. The Gaussian kernel allows us to retain the scores~$n_{\bf{p}}$ near the output of the discrete argmax while suppressing others. That is, the kernel~$k_{\bf{p}}$ has the effect of restricting the range of averaging in~\eqref{eq:kernel_soft_max}, and makes the kernel soft argmax less susceptible to multi-modal distributions~(\eg,~from ambiguous matches in clutter and repetitive patterns) while maintaining differentiability. Note that the center position of the Gaussian kernel is changed at every iteration during training, and the matching probability~$m_{\bf{p}}$ is differentiable, since we do not train the Gaussian kernel itself and no gradients are propagated through the discrete argmax. Note also that the normalization of the correlation map is particularly important for semantic alignment methods~\cite{rocco2017convolutional,rocco2018end,Seo2018AttentiveSA,Jeon2018PARN}~(see, for example, Table~2 in~\cite{rocco2017convolutional}) but its purpose is different from ours. They use the normalization to penalize features having multiple highly-correlated matches, boosting the scores of discriminative matches.

We visualize soft and kernel soft argmax operators in Fig.~\ref{fig:KS_argmax}, which shows that the soft argmax yields an incorrect correspondence in the presence of multiple highly correlated features, since a weighted average of matching probabilities~$m_{\bf{p}}$ having multi-modal distributions accumulates positional errors. The kernel soft argmax instead suppresses matching probabilities~$m_{\bf{p}}$ except for the ones around the highest mode, making them have an (approximately) uni-modal distribution and favoring correct correspondences. The discrete argmax for the Gaussian kernel~$k_{\bf{p}}$ can select incorrect correspondences from a correlation map~$n_{\bf{p}}$ during training. This can be handled by flow consistency and smoothness terms in our loss, which will be described in the following section. They penalize inconsistent flows and outlier matches, making it possible to learn feature descriptors in such a way that correlation scores of incorrect matches become smaller. We visualize in Fig.~\ref{fig:incorrect_mode} matching probabilities~$m_{\bf{p}}$ and shifts of a dominant mode during training. We can see that the discrete argmax initially selects an incorrect match, but the dominant mode shifts toward a correct match after the second iteration.

\begin{figure}[t]
	\begin{center}
	   \includegraphics[width=0.98\columnwidth]{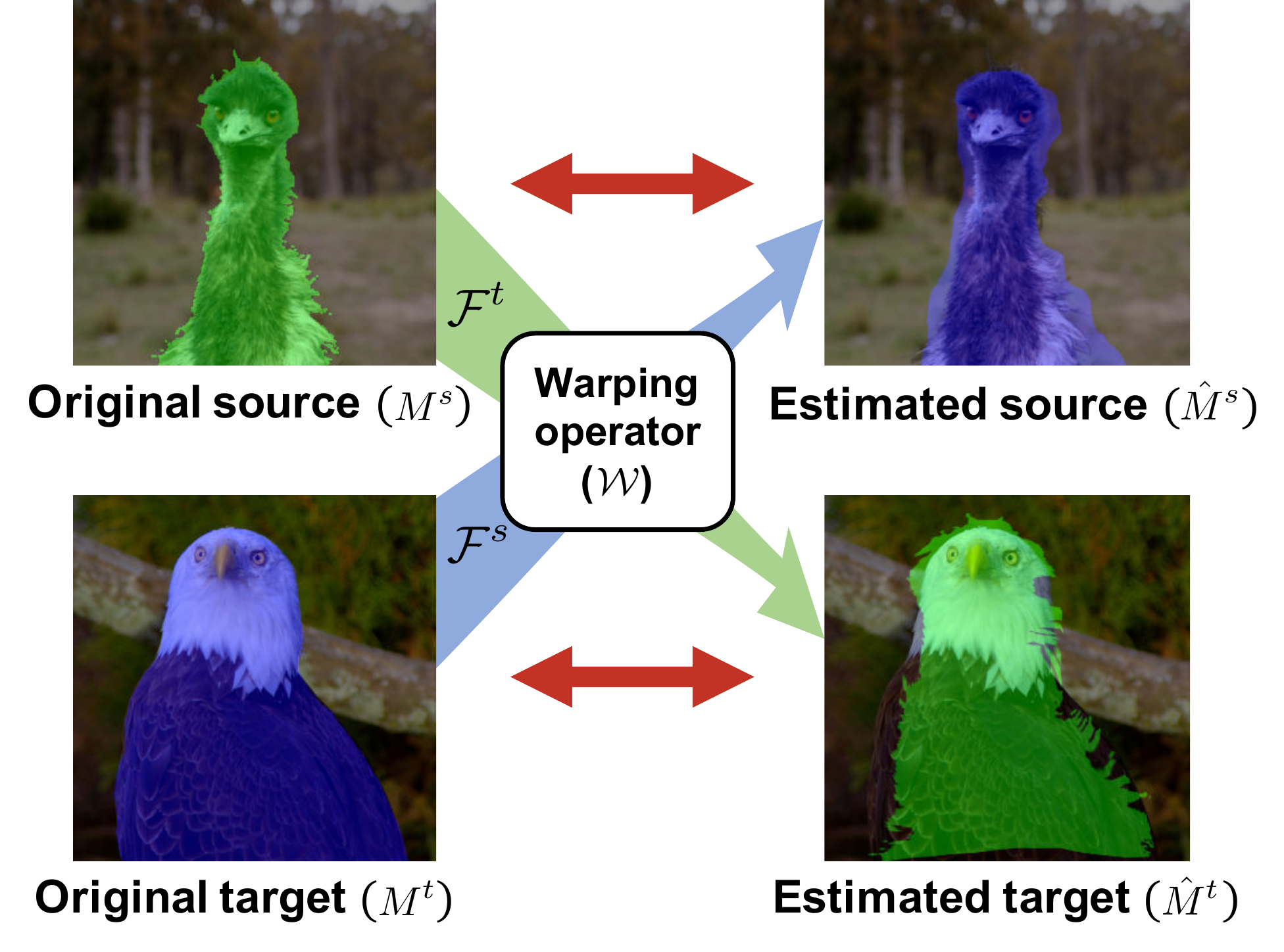}
	   \vspace{-0.3cm} 
	\end{center}
	   \caption{Illustration of the mask consistency loss. We estimate the binary source mask~$\hat{M}^s$ by warping the target one~$M^t$ using the flow field~$\mathcal{F}^{s}$. The target mask~$\hat{M}^t$ is similarly estimated. We then compute the average error between original and estimated masks to compute the mask consistency loss. This penalizes the correspondences between foreground and background regions, and vice versa. We show foreground parts only for the purpose of visualization.~(Best viewed in color.)}
	\label{fig:mask_consist}
	\vspace{-0.2cm}
\end{figure}

\begin{figure*}[t] % loss discussion figure1
\captionsetup[subfigure]{aboveskip=-0.5pt,belowskip=-0.5pt}
\centering
	\begin{subfigure}{0.33\textwidth}
	\includegraphics[width=\textwidth]{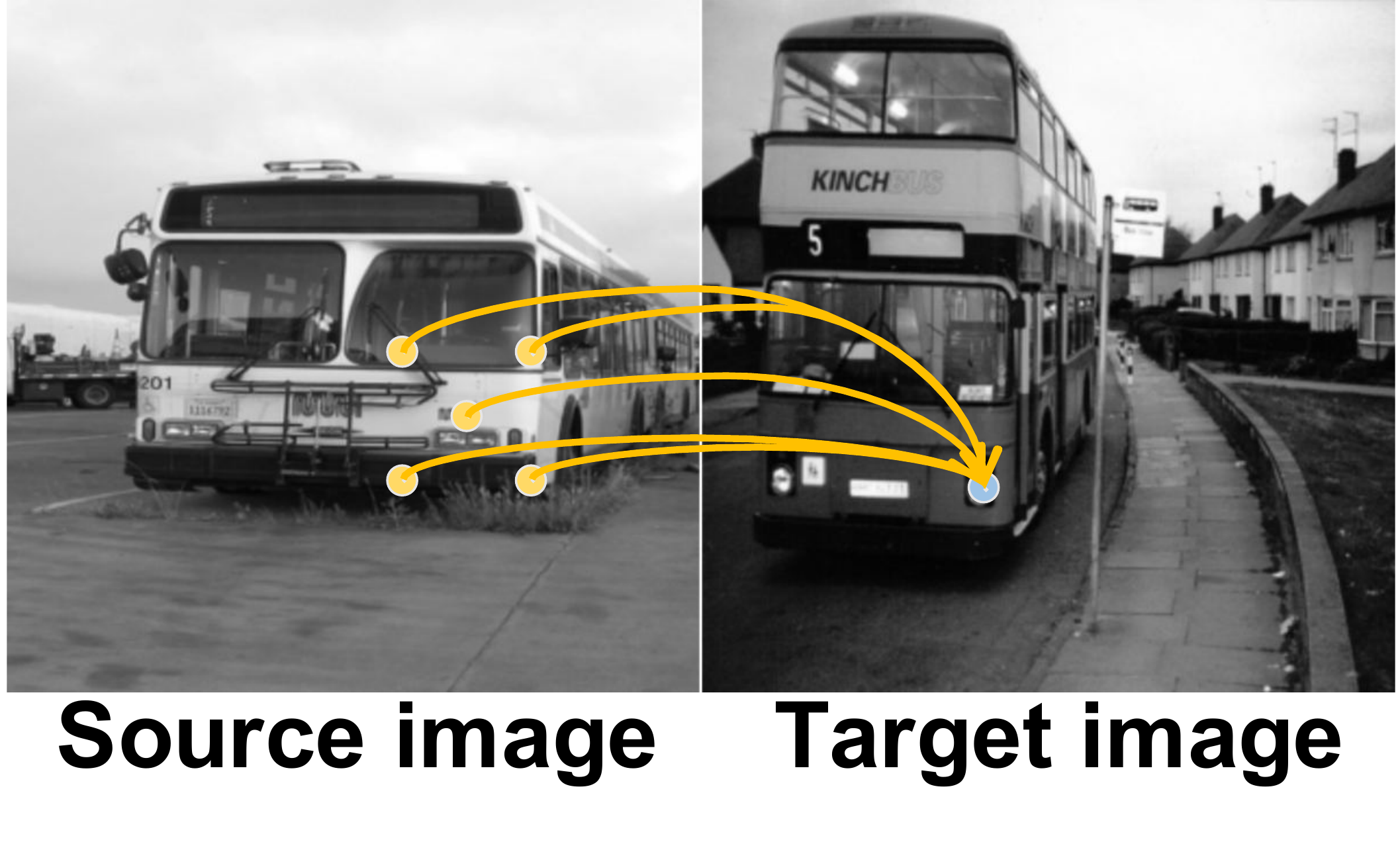}
	\vspace{-0.4cm}
	\caption*{(a) Many-to-one matching.}
	\end{subfigure}
	\begin{subfigure}{0.33\textwidth}
	\includegraphics[width=\textwidth]{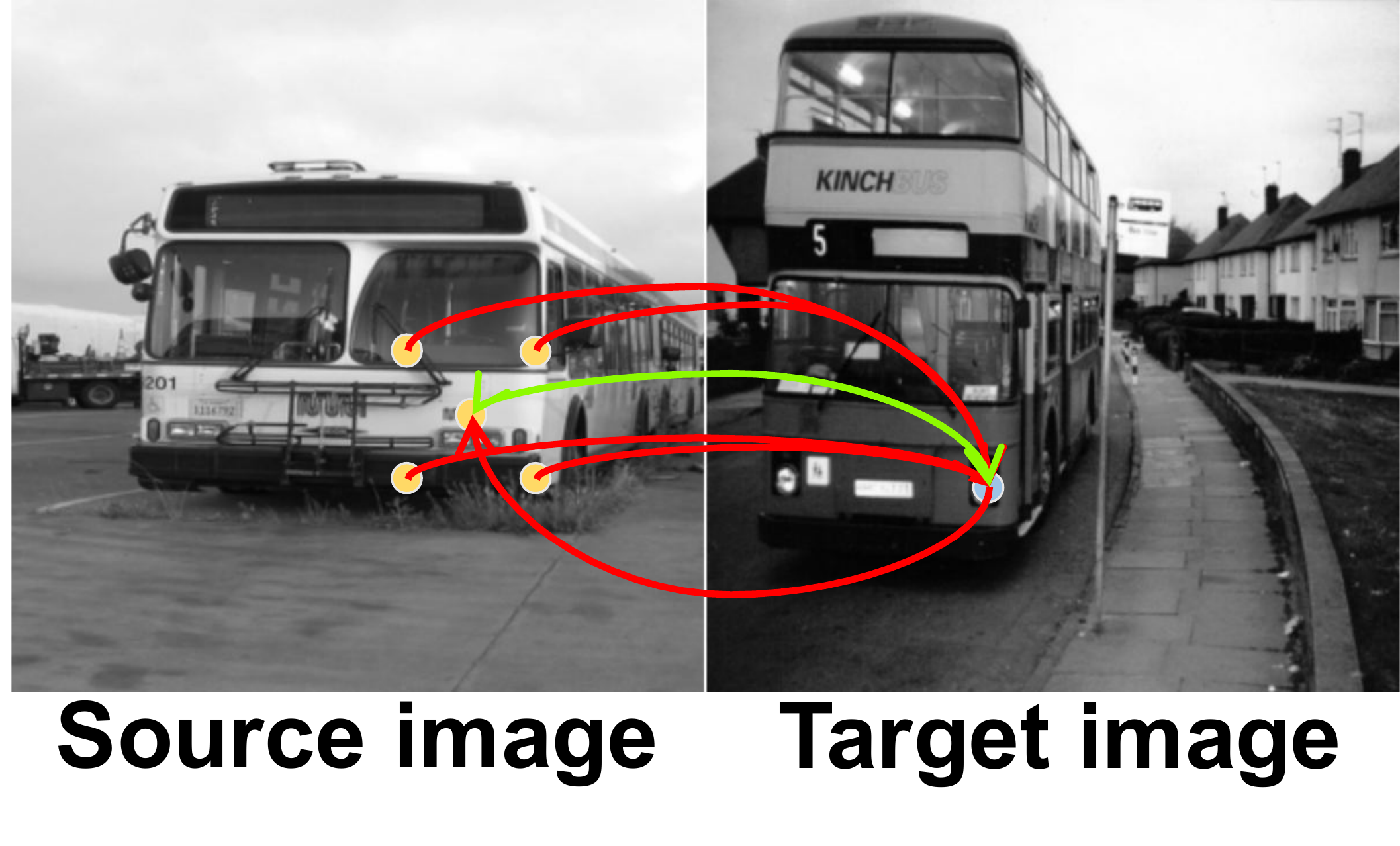}
	\vspace{-0.4cm}
	\caption*{(b) Consistent and inconsistent flows.}
	\end{subfigure}
	\begin{subfigure}{0.33\textwidth}
	\includegraphics[width=\textwidth]{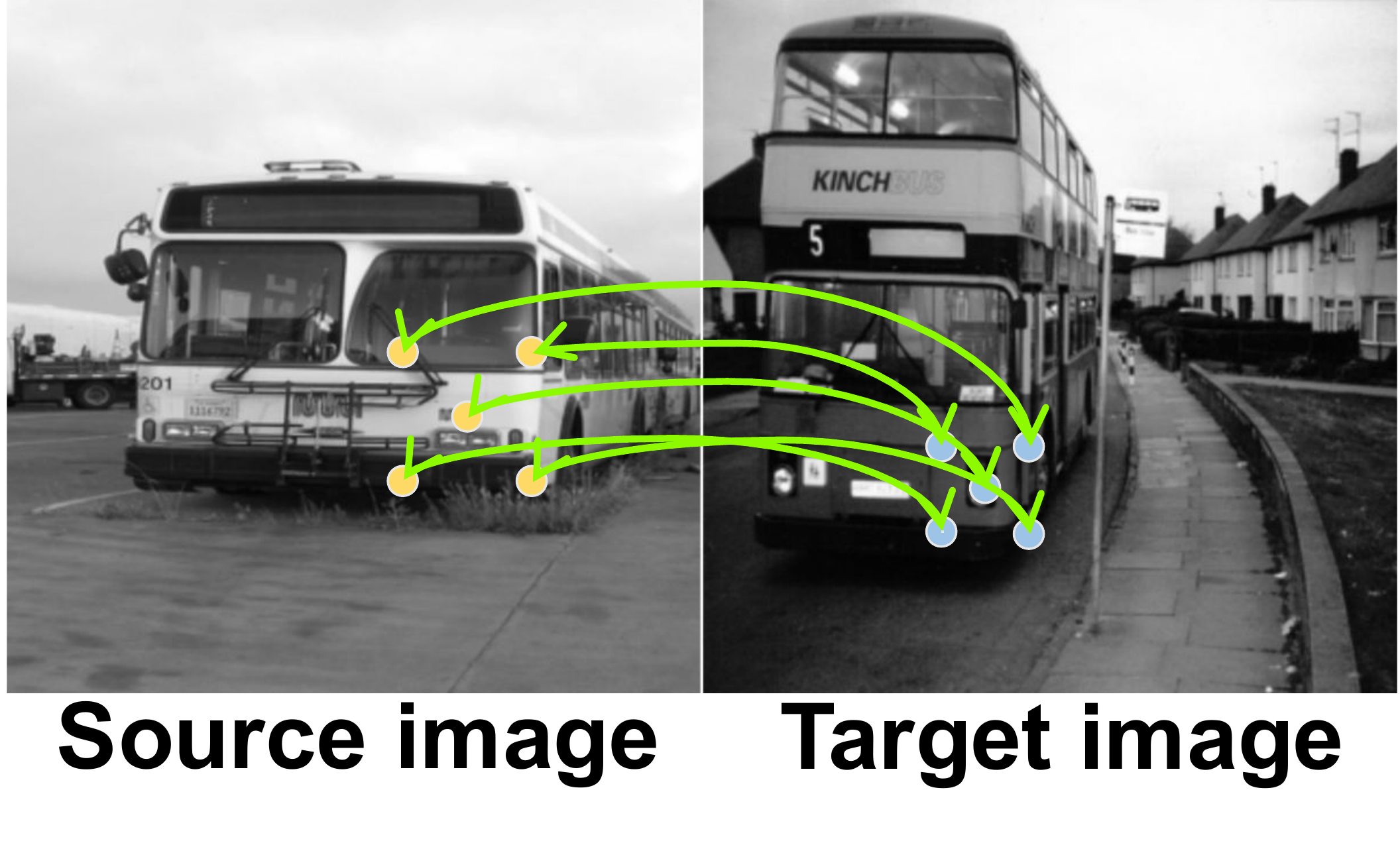}
	\vspace{-0.4cm}
	\caption*{(c) One-to-one matching.}
	\end{subfigure}
	\vfill
	\vspace{-0.2cm}
\caption{Using the mask consistency term alone may cause a many-to-one matching problem: (a) multiple yellow points in the source image can be matched to the single blue one in the target image. The flow consistency term (b) penalizes inconsistent correspondences and (c) favors a one-to-one matching. We denote by green and red arrows consistent and inconsistent matches, respectively.~(Best viewed in color.)}
\vspace{-0.2cm}
\label{fig:loss}
\end{figure*}

\begin{figure*}[h] % loss discussion figure2
\captionsetup[subfigure]{aboveskip=-0.5pt,belowskip=-0.5pt}
\centering
	\begin{subfigure}{0.33\textwidth}
	\includegraphics[width=\textwidth]{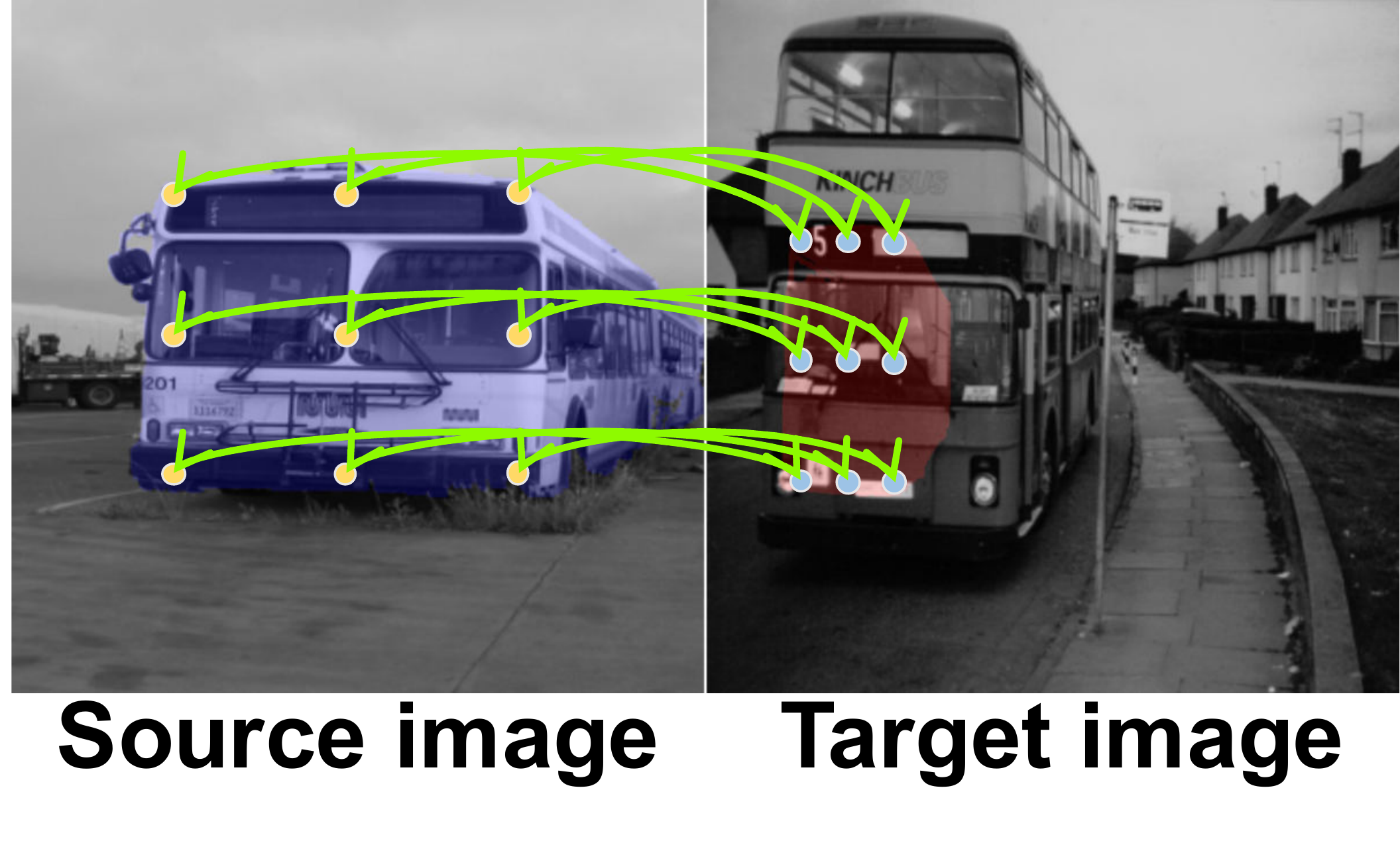}
	\vspace{-0.4cm}
	\caption*{(a) Flow consistency for the source image.}
	\end{subfigure}
	\begin{subfigure}{0.33\textwidth}
	\includegraphics[width=\textwidth]{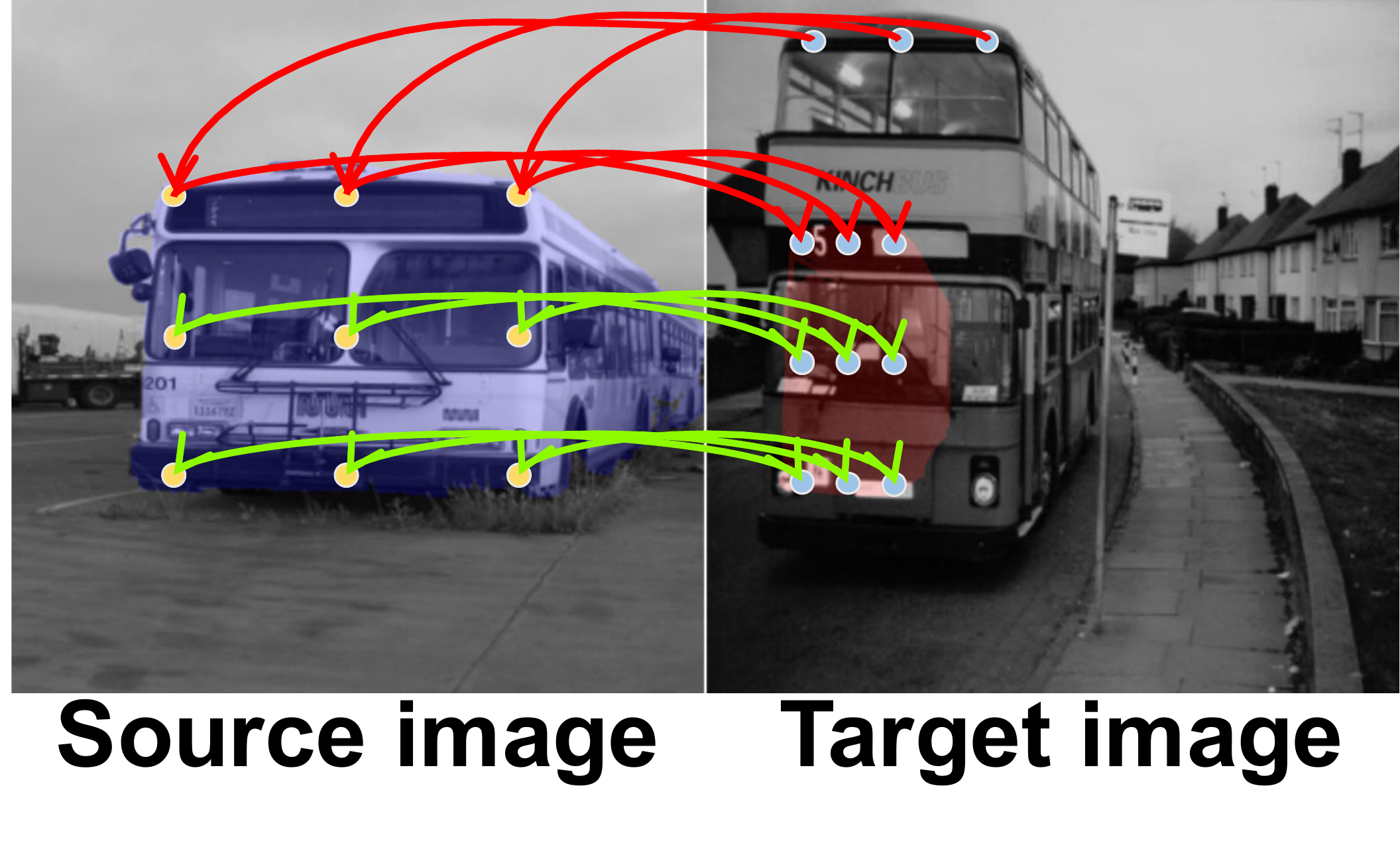}
	\vspace{-0.4cm}
	\caption*{(b) Flow consistency for the target image.}
	\end{subfigure}
	\begin{subfigure}{0.33\textwidth}
	\includegraphics[width=\textwidth]{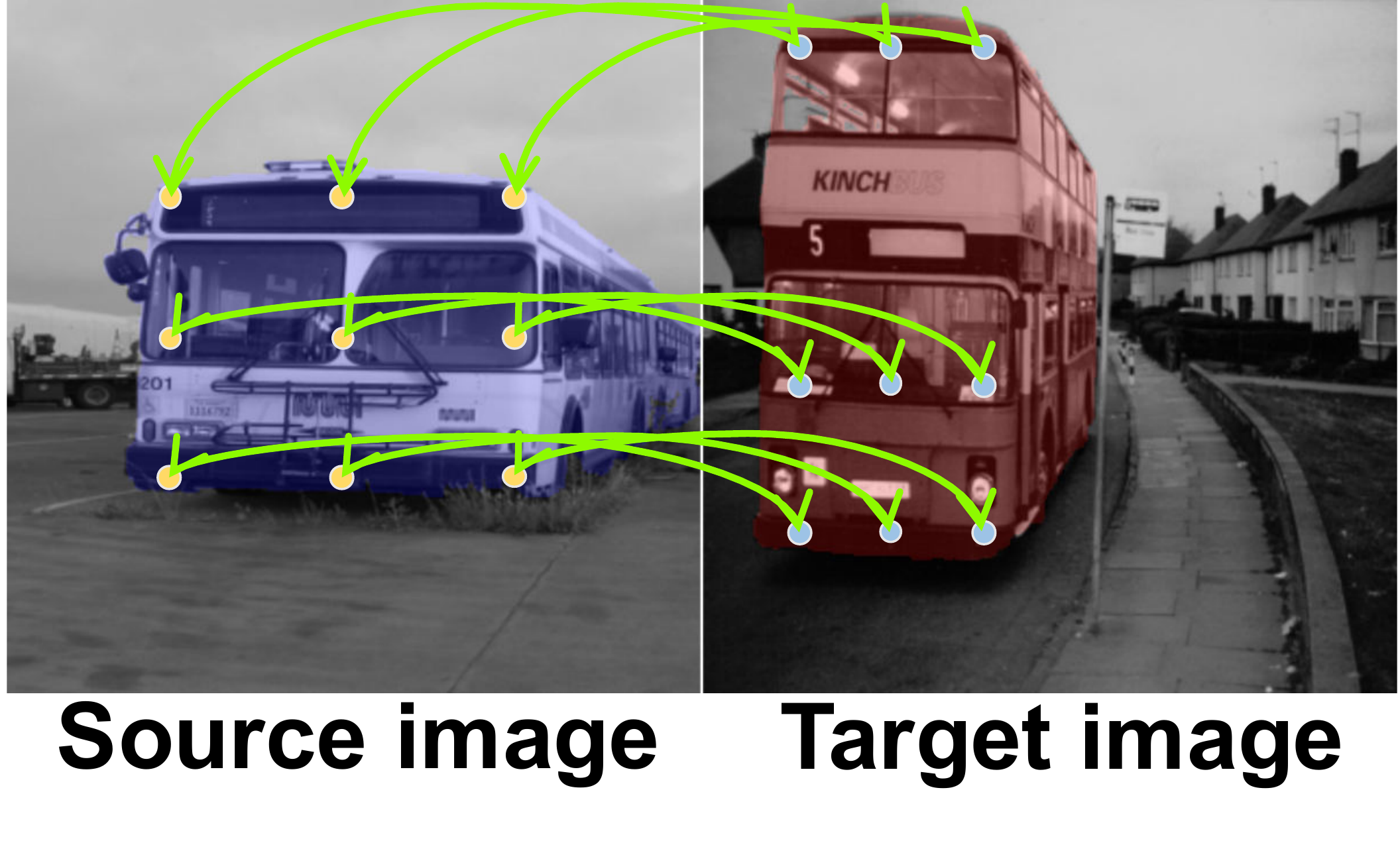}
	\vspace{-0.4cm}
	\caption*{(c) Flow consistency for both images.}
	\end{subfigure}
	\vfill
	\vspace{-0.2cm}
\caption{Using a symmetric loss: (a) considering the flow consistency loss w.r.t a source image only may cause a flow shrinkage problem; (b) we can overcome this problem by computing the loss w.r.t a target image as well and penalizing inconsistent matches; (c) this symmetric loss allows us to perform object-level matching. We use green and red arrows to show consistent and inconsistent matches, respectively.~(Best viewed in color.)}
\label{fig:pair_loss}
\end{figure*}

\subsection{Loss}
\label{sec:loss}
We exploit binary foreground masks as a supervisory signal to train the network, which gives a strong object prior. To this end, we define three losses that guide the network to learn object-aware correspondences without pixel-level ground truth as
\begin{equation}\label{eq:total_loss}
	\mathcal{L} = \lambda_{\mathrm{mask}} \mathcal{L}_{\mathrm{mask}} + 
								   \lambda_{\mathrm{flow}} \mathcal{L}_{\mathrm{flow}} + 
								   \lambda_{\mathrm{smooth}} \mathcal{L}_{\mathrm{smooth}},
\end{equation}
which consists of mask consistency~$\mathcal{L}_{\mathrm{mask}}$, flow consistency~$\mathcal{L}_{\mathrm{flow}}$ and smoothness~$\mathcal{L}_{\mathrm{smooth}}$ terms, balanced by the weight parameters~($\lambda_{\mathrm{mask}}$, $\lambda_{\mathrm{flow}}$, $\lambda_{\mathrm{smooth}}$). In the following, we describe each term in detail.  

\subsubsection{Mask Consistency Term}
We define a flow field~$\mathcal{F}^s$ from source to target images as 
\begin{equation}
	\mathcal{F}^{s} ({\bf{p}}) = \phi({\bf{p}}) - {\bf{p}}.
\end{equation}
Similarly, a flow field~$\mathcal{F}^t ({\bf{q}})$ from target to source images is defined as~$\phi({\bf{q}}) - {\bf{q}}$. We denote by~$M^s$ and $M^t$ the binary masks of the source and target images, respectively. Values of 0 and 1 in the masks respectively indicate background and foreground regions. We assume that the binary mask in the source images can be reconstructed by warping the mask in the target image and vice versa, if we have discriminative features and correct dense correspondences. To implement this idea, we transfer the target mask~$M^t$ by warping~\cite{jaderberg2015spatial} using the flow field~$\mathcal{F}^s$ and obtain an estimate of the source mask~$\hat M^s$ as follows.
\begin{equation}
	\hat M^s = \mathcal{W}(M^t ; \mathcal{F}^s) 
\end{equation}
where~$\mathcal{W}$ denotes a warping operator using the flow field,~\eg,
\begin{equation}
\begin{aligned}
\mathcal{W}(M^t ; \mathcal{F}^s)({\bf{p}}) & = M^t({\mathbf{p}} + \mathcal{F}^{s} ({\mathbf{p}})) \\
& = M^t(\phi(\mathbf{p})).
\end{aligned}
\end{equation} 	
Since the sampling positions from the correspondences~$\phi(\mathbf{p})$ are typically fractional, we use a bilinear kernel~\cite{jaderberg2015spatial} to compute corresponding values of $M^t$:
\begin{equation}
\begin{aligned}
M^t(\phi(\mathbf{p}))=\sum_{q_x,q_y} M^t(q_x, q_y) &\max(0, 1-|\phi(\mathbf{p})_x-q_x|) \\
&\max(0, 1-|\phi(\mathbf{p})_y-q_y|).
\end{aligned}
\end{equation}               
We then compute the difference between the source mask~$M^s$ and its estimate~$\hat M^s$. Similarly, we reconstruct the target mask~$\hat M^t$ from $M^s$ using the flow field~$\mathcal{F}^t$ and compute the difference between~$\hat M^t$ and $M^t$.  Accordingly, we define the mask consistency loss~(Fig.~\ref{fig:mask_consist}) as
\begin{equation}
	\mathcal{L}_{\mathrm{mask}} = \sum_{i \in \{s,t \}} \Biggl(\frac{1}{|N^i|} \sum_{{\bf{p}}} ({M}^{i}({\bf{p}}) - \hat {M}^{i}({\bf{p}}))^2\Biggr),
\end{equation}
where~$|N^i|$ is the number of pixels in the mask~$M^{i}$. Although the mask consistency loss does not constrain the background itself, it prevents matches from foreground to background regions and vice versa by penalizing them. This encourages correspondences to be established between features within foreground and background masks, guiding our model to learn object-aware correspondences. Note that the mask consistency loss does not guarantee that our model establishes more accurate correspondences at the object boundaries. Note also that this loss using binary masks does not prevent a many-to-one matching~(Fig.~\ref{fig:loss}(a)). That is, it does not penalize a case when many foreground features in an image are matched to a single one in other image. For example, the foreground mask in the source image even can be reconstructed, when all points in the foreground region are matched to a single foreground point in the target image.

\subsubsection{Flow Consistency Term}
To address the many-to-one matching problem, we propose to use a flow consistency loss. It measures consistency between flow fields~$\mathcal{F}^s$ and~$\mathcal{F}^t$ within foreground masks as  
\begin{equation}\label{eq:flow_consistency}
	\mathcal{L}_{\mathrm{flow}} = \sum_{i \in \{s,t \}} \Biggl(\frac{1}{|N^i_F|} \sum_{{\bf{p}}} ||(\mathcal{F}^i({\bf{p}}) + \hat{\mathcal{F}}^i({\bf{p}})) \odot M^i({\bf{p}})||^{2}_{2}\Biggr) ,
\end{equation}
where~$|N^i_F|$ is the number of foreground pixels in the mask~$M^i$, and
\begin{equation}\label{eq:flow_warp}
	\hat{\mathcal{F}}^s = \mathcal{W}(\mathcal{F}^t; \mathcal{F}^s),
\end{equation}
which aligns the flow field~$\mathcal{F}^t$ with respect to~$\mathcal{F}^s$ by warping. $\hat{\mathcal{F}}^t$ is computed similar to~\eqref{eq:flow_warp}. We denote by $\left\lVert \cdot \right\rVert_{2}$ and $\odot$ the L2 norm and element-wise multiplication, respectively. The multiplication is applied separately for each $x$ and $y$ component. 

The flow consistency term penalizes inconsistent correspondences~(Fig.~\ref{fig:loss}(b)), and favors one-to-one matching~(Fig.~\ref{fig:loss}(c)), alleviating the many-to-one matching problem in the mask consistency loss. For example, when the flow fields are consistent with each other,~$\mathcal{F}^s$ and~$\hat{\mathcal{F}}^s$ in \eqref{eq:flow_consistency}~have the same magnitude with opposite directions. Note that having multiple matches for individual points~(\ie,~a one-to-many matching) is impossible within our framework. Similar ideas have been explored in stereo fusion~\cite{zbontar2015computing,godard2017unsupervised} and optical flow~\cite{meister2018unflow,zou2018df}, but without appearance and shape variations. It is hard to incorporate this term in current semantic flow methods based on CNNs~\cite{choy2016universal,han2017scnet,kim2017fcss,rocco2018neighbourhood,min2019hyperpixel}, mainly due to a lack of differentiability of the flow field. Recently, Zhou~\etal~\cite{zhou2016learning} exploit cycle consistency between flow fields, but they regress correspondences directly from concatenated features from source and target images and do not consider background clutter. In contrast, our method establishes a differentiable flow field by computing feature similarities explicitly while considering background clutter.

Although the flow consistency term relieves the many-to-one matching problem, computing this term for a source or a target image only may cause a flow shrinkage problem~(Fig.~\ref{fig:pair_loss}(a)). To address this problem, we compute this term w.r.t both source and target images in~\eqref{eq:flow_consistency}. This penalizes inconsistent matches, \eg,~between the entire foreground region in the source image and small parts of the target image~(Fig.~\ref{fig:pair_loss}(b)). Note that spreading the flow fields over the entire regions is particularly important to handle scale changes between objects~(Fig.~\ref{fig:pair_loss}(c)).

%\todo{In Fig.~\ref{fig:loss}~(b), all points except the center one are penalized by this term.} \todo{Accordingly,} the flow consistency term favors a one-to-one matching\todo{~(Fig.~\ref{fig:loss}~(c))}, spreading flow fields over foreground regions and alleviating the many-to-one matching problem in the mask consistency loss. For example, when the flow fields are consistent with each other,~$\mathcal{F}^s$ and~$\hat{\mathcal{F}}^s$~have the same magnitude with opposite directions. Similar ideas have been explored in stereo matching~\cite{zbontar2015computing,godard2017unsupervised} and optical flow~\cite{meister2018unflow,zou2018df}, but without considering appearance and shape variations. It is hard to incorporate this term in current semantic flow methods based on CNNs~\cite{choy2016universal,han2017scnet,kim2017fcss} mainly due to a lack of differentiability of the flow field. Recently, Zhou~\etal~\cite{zhou2016learning} exploit cycle consistency between flow fields, but they regress correspondences directly from concatenated features from source and target images and do not consider background clutter. In contrast, our method establishes a differentiable flow field by computing feature similarities explicitly while considering background clutter.

\subsubsection{Smoothness Term}
The differentiable flow field also allows us to exploit a smoothness term, as widely used in classical energy-based approaches~\cite{liu2011sift,kim2013deformable,hur2015generalized}. We define this term using the first-order derivative of the flow fields~$\mathcal{F}^s$ and~$\mathcal{F}^t$ as
\begin{equation}\label{eq:smoothness}
	\mathcal{L}_{\mathrm{smooth}} = \sum_{i \in \{s,t \}} \Biggl(\frac{1}{|N^i_F|} \sum_{{\bf{p}}}
	|| \nabla \mathcal{F}^{i} ({\bf{p}}) \odot M^i({\bf{p}}) ||_1 \Biggr) ,
\end{equation}
where $\left\lVert \cdot \right\rVert_1$ and~$\nabla$ are the L1 norm and the gradient operator, respectively. This regularizes~(or smooths) flow fields within foreground regions without being affected by (incorrect) correspondences at background. 

%\todo{
%\subsubsection{Discussion}
%Although the flow consistency loss gives a one-to-one matching, using this loss for a source or a target image only may cause a flow shrinkage problem~(Fig.~\ref{fig:pair_loss}(a)). Computing this loss in a symmetric way alleviates this problem. We compute the flow consistency term w.r.t both source and target images \todo{in~\eqref{eq:flow_consistency}}. It penalizes the inconsistent matches, \eg,~between the entire foreground region in the source image and small parts of the target image~(Fig.~\ref{fig:pair_loss}(b)). \sout{Using the first or the second term in~\eqref{eq:flow_consistency} only does not handle such cases.} Note that spreading the flow fields over the entire regions is particularly important to handle scale changes between objects~(Fig.~\ref{fig:pair_loss}(c)).
%}

\section{Experiments}
\label{sec:exp}
In this section, we give experimental details~(Secs.~\ref{sec:exp_details}), and present a detailed analysis and evaluation of our approach on the tasks of semantic correspondence~(Section~\ref{sec:semantic_correspondence_results}), mask transfer~(Section~\ref{sec:mask_transfer_results}), pose keypoint propagation~(Section~\ref{sec:tracking_results}) and object co-segmentation~(Section~\ref{sec:coseg}). We then present ablation studies for different losses and network architectures, as well as a performance analysis for different training datasets~(Section~\ref{sec:ablation}).

%\todo{
%\sout{In this section we present a detailed analysis and evaluation of our approach including ablation studies on different losses and network architectures.}
%In this section, we first introduce our implementation details~(Sec.~\ref{sec:imple}) and training setup~(Sec.~\ref{sec:training}). Next, we present a detailed analysis and evaluation of our approach on semantic correspondence~(Sec.~\ref{sec:semantic_correspondence_results}) and pose tracking~(Sec.~\ref{sec:tracking_results}) tasks. We then show ablation studies on different losses and network architectures~(Sec.~\ref{sec:ablation}). Lastly, we investigate how our model is affected by different training datasets~(Sec.~\ref{sec:different_datasets}).
%}

\subsection{Experimental Details} \label{sec:exp_details}
\subsubsection{Implementation} \label{sec:imple}
Following~\cite{kim2018recurrent,rocco2018neighbourhood,rocco2018end,Seo2018AttentiveSA,min2019hyperpixel}, we use CNN features from ResNet-101~\cite{he2016deep} trained for ImageNet classification~\cite{deng2009imagenet}. Specifically, we use the networks cropped at $\texttt{conv4-23}$ and $\texttt{conv5-3}$ layers, respectively. This results in two feature maps of size $20 \times 20 \times 1024$ and $10 \times 10 \times 2048$, respectively, for a pair of input images of size~$320 \times 320$, and gives a good compromise between localization accuracy and high-level semantics. Adaptation layers are trained with random initialization, separately for each feature map in a residual fashion~\cite{he2016deep}. To compute residuals, we add two blocks of convolutional, batch normalization~\cite{Ioffe2015BatchNA} and ReLU~\cite{krizhevsky2012imagenet} layers, with padding on top of each feature map, where the sizes of convolutional kernels for $\texttt{conv4-23}$ and $\texttt{conv5-3}$ features are $5 \times 5$ and $3 \times 3$, respectively. Each block outputs a residual, which is then added to the corresponding input features. Adaptation layers aggregate the features nonlinearly from large receptive fields~(\eg,~the receptive field of size $9 \times 9$ on a feature map of size $20 \times 20 \times 1024$), transforming them, guided by semantic correspondences and the corresponding loss terms in~\eqref{eq:total_loss}, to be highly discriminative w.r.t both appearance and spatial context. With the resulting two feature maps of size~$20 \times 20 \times 1024$ and $20 \times 20 \times 2048$\footnote{We upsample the features adapted from $\texttt{conv5-3}$ using bilinear interpolation.}, we compute pairwise matching scores and then combine them by element-wise multiplication, resulting in a correlation map of size~$20 \times 20 \times 20 \times 20$. Following~\cite{rocco2017convolutional,Seo2018AttentiveSA}, we fix the feature extractor, and train adaptation layers only. At test time, we upsample a flow field of size $20 \times 20$ using bilinear interpolation.

\begin{figure}[t]
	\begin{center}
	   \includegraphics[width=1.1\columnwidth]{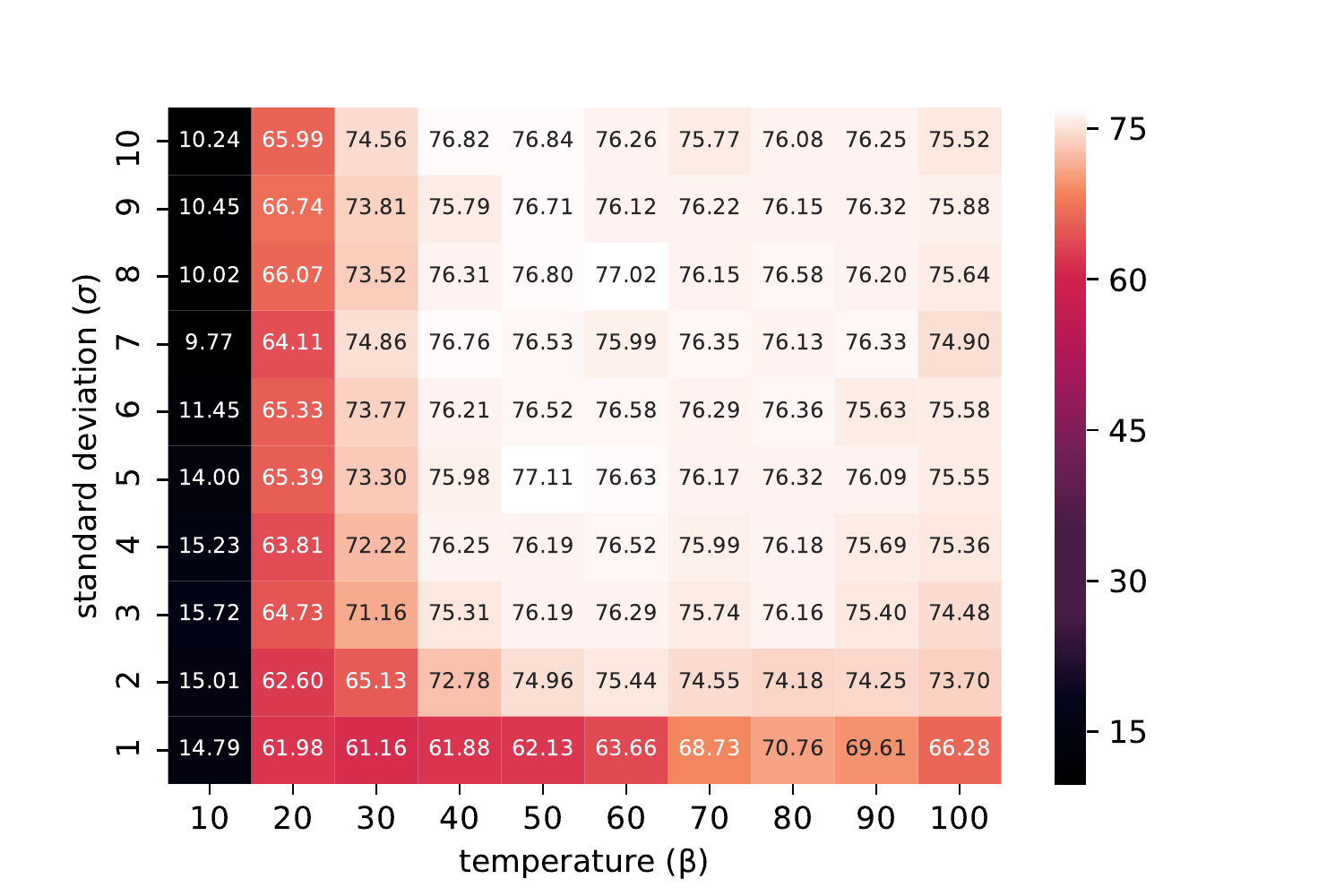}
	   \vspace{-0.3cm} 
	\end{center}
	   \vspace{-0.3cm} 	
	\caption{Average PCK scores~($\alpha_\mathrm{bbox}=0.1$) for ($\beta$,~$\sigma$) pairs on the validation split of PF-PASCAL~\cite{ham2017proposal}.}
	\label{fig:heatmap}	
\end{figure}

%We empirically set the temperature parameter~$\beta$ to 50 and standard deviation~$\sigma$ of Gaussian kernel $k_{\bf{p}}$ to 5. 
We determine the temperature parameter~$\beta$ and standard deviation~$\sigma$ of Gaussian kernel~$k_{\bf{p}}$ using a grid search over ($\beta$,~$\sigma$) pairs, where the maximum search ranges for~$\beta$ and $\sigma$ are 100 and 10 with intervals of 10 and 1, respectively. We show in Fig.~\ref{fig:heatmap} average PCK scores for ($\beta$,~$\sigma$) pairs on the validation split of PF-PASCAL~\cite{ham2017proposal}. We can see that the PCK performance is robust over a wide range of parameters. We can also see that the PCK scores decrease significantly, when the temperature~$\beta$ is too small, since the matching probability~$m_\mathbf{p}$ approaches a uniform distribution. We select the parameters~($\beta=50$,~$\sigma=5$) that give the best performance. Other parameters~($\lambda_{\mathrm{mask}}=3$, $\lambda_{\mathrm{flow}}=16$, $\lambda_{\mathrm{smooth}}=0.5$) are chosen similarly using the validation split of the PF-PASCAL dataset. Note that we perform a grid search on one validation set and fix these parameters in all experiments. As will be seen in experimental results, they generalize well to the test sets of all the datasets used (\eg,~PF-WILLOW~\cite{ham2016proposal} and TSS~\cite{taniai2016joint}) and tasks (\eg,~mask transfer and pose keypoint propagation).

%We fix these parameters in all experiments. 

%We determine those values using a grid search over ($\beta$,~$\sigma$) pairs, where the maximum search ranges for~$\beta$ and $\sigma$ are 100 and 10 with intervals of 10 and 1, respectively.  \changed{We show in Fig.~\ref{fig:heatmap} average PCK scores for ($\beta$,~$\sigma$) pairs on the validation split of PF-PASCAL~\cite{ham2017proposal}. We can see that the PCK performance is robust over a wide range of parameters. However, the significant performance degradation is observed when the temperature~$\beta$ is too small. This is because the matching probabilities~$m_\mathbf{p}$ approach the uniform distribution when we decrease the temperature significantly even with the Gaussian kernel. The kernel soft argmax function in Eq.~(2) cannot approximate the discrete argmax correctly, since its output is affected by all spatial positions. This problem is also observed in the soft argmax function~[46],~[47].} We select the parameters that give the best performance on the validation split of the PF-PASCAL dataset~\cite{ham2017proposal,rocco2018end}. Other parameters~($\lambda_{\mathrm{mask}}=3$, $\lambda_{\mathrm{flow}}=16$, $\lambda_{\mathrm{smooth}}=0.5$) are chosen similarly using the validation split of the PF-PASCAL dataset. We fix these parameters in all experiments. 

%To compute residuals, we add \todo{two successive} $5 \times 5$ and \todo{two successive} $3\times 3$ convolutional layers with padding on top of $\texttt{conv4-23}$ and $\texttt{conv5-3}$, respectively, with batch normalization~\cite{Ioffe2015BatchNA} and the ReLU~\cite{krizhevsky2012imagenet}. The residuals are then added to the corresponding input features.

\subsubsection{Training}  \label{sec:training}
Training our network requires pairs of foreground masks for source and target images depicting different instances of the same object category. Although the TSS~\cite{taniai2016joint} and Caltech-101~\cite{fei2006one} datasets provide such pairs, the number of masks in TSS~\cite{taniai2016joint} is not sufficient to train our network, and images in Caltech-101~\cite{fei2006one} lack background clutter. Our model trained with these datasets suffers from overfitting and may not generalize well for other images containing clutter. Motivated by~\cite{kanazawa2016warpnet,rocco2017convolutional,Seo2018AttentiveSA,novotny2018self}, we generate pairs of source and target images synthetically from single images by applying random affine transformations and use the synthetically warped pairs as training samples. The corresponding foreground masks are transformed with the same parameters. Contrary to~\cite{kanazawa2016warpnet,rocco2017convolutional,Seo2018AttentiveSA}, our model does not perform parametric regression, and thus does not require ground-truth transformation parameters for training. We use the PASCAL VOC 2012 segmentation dataset~\cite{everingham2010pascal} that consists of 1,464, 1,449, and 1,456 images for training, validation and test, respectively. We exclude 122 images from train/validation sets that overlap with the test split in PF-PASCAL~\cite{ham2017proposal,rocco2018end}, and train our model with the corresponding 2,791 images. We augment the training dataset by horizontal flipping and color jittering. Note that we do not use the segmentation masks provided by the PASCAL VOC 2012 dataset, that specify the class of the object at each pixel. We instead generate binary foreground masks using all labeled objects, regardless of image categories and the number of objects, at training time. We train our model with a batch size of 16 for about 7k iterations, giving roughly 40 epochs over the training data. We use the Adam optimizer~\cite{kingma2014adam} with~$\beta_1 = 0.9$ and~$\beta_2=0.999$. A learning rate initially set to 3e-5 is divided by 5 after 30 epochs. All networks are trained end to end using $\texttt{PyTorch}$\cite{paszke2017automatic}.

\subsubsection{Evaluation Metric}\label{sec:metric}
We use the probability of correct keypoint~(PCK)~\cite{yang2013articulated} to measure the precision of overall assignment, particularly at sparse keypoints of semantic relevance. We compute the Euclidean distance between warped keypoints using the estimated dense flow and ground truth, and count the number of keypoints whose distances lie within $\alpha_n \text{max}(h_n,w_n)$ pixels, where $n \in \{\mathrm{img}, \mathrm{bbox}\}$, $\alpha$ typically set to $0.1$, is a tolerance value, and $h$ and $w$ are the height and width, respectively, of an image or an object bounding box. Following~\cite{rocco2017convolutional,rocco2018end}, we divide keypoint coordinates by the height and width of the image size in case of $\alpha_{\mathrm{img}}$, such that they are normalized in a range of $[0,1]$ and $h_{\mathrm{img}}=w_{\mathrm{img}}=1$.

%\todo{We use the probability of correct keypoint~(PCK)~\cite{yang2013articulated} to measure the precision of overall assignment, particularly at sparse keypoints of semantic relevance. To compute PCK scores, we use two different threshold values $\alpha_{\mathrm{bbox}}$ and $\alpha_{\mathrm{img}}$. For $\alpha_{\mathrm{bbox}}$, we compute the Euclidean distances between warped keypoints using an estimated dense flow and ground truth, and count the number of keypoints whoses distances lie within $\alpha_{\mathrm{bbox}}\text{max}(h,w)$ pixels, where $h$ and $w$ are the height and width of the bounding box, respectively. For $\alpha_{\mathrm{img}}$, the keypoint coordinates are normalized in the range of $[0,1]$ by dividing them with the height and width of the image size, respectively, and compute the Euclidean distances between warped keypoints and ground truth. We then count the number of keypoints whose distances lie within threshold values~($\alpha_{\mathrm{img}}$).}

\subsection{Semantic Correspondence}  \label{sec:semantic_correspondence_results}
We compare our model to the state of the art on semantic correspondence including hand-crafted and CNN-based methods with the following four benchmark datasets:~PF-WILLOW~\cite{ham2016proposal}, PF-PASCAL~\cite{ham2017proposal}, SPair-71k~\cite{min2019hyperpixel}, and TSS~\cite{taniai2016joint}. Following the experimental protocol in~\cite{rocco2017convolutional,rocco2018end,min2019hyperpixel}, we use $\alpha_{\mathrm{img}}$ for PF-PASCAL and TSS, and $\alpha_{\mathrm{bbox}}$ for PF-WILLOW and SPair-71k, respectively. Results for all comparisons have been obtained from the source code or models provided by the authors, unless otherwise specified.

\setlength{\tabcolsep}{0.3em}
\begin{table}[t]
\caption{Quantitative comparison with the state of the art on the PF-WILLOW~\cite{ham2016proposal} and the test split of the PF-PASCAL~\cite{ham2017proposal,rocco2018end} in terms of average PCK. Numbers in bold indicate the best performance and underscored ones are the second best. The subscript for each method indicates the corresponding feature extractor. We denote by ``F'' and ``A'', respectively, semantic flow and semantic alignment methods. The characters in parentheses correspond to the type of supervisory signal used in training:~T: transformation parameters; P: image pairs depicting different instances of the same object category; B: bounding boxes; C: ground-truth correspondences; M: foreground masks. All numbers in PF-WILLOW are taken from~\cite{min2019hyperpixel,kim2018recurrent}. The results of~\cite{kim2018recurrent,ham2016proposal,kim2019fcss} in PF-PASCAL are taken from~\cite{kim2018recurrent}.}
\vspace{-0.1cm}
\small
\begin{center}
\begin{tabular}{@{}C{0.7cm}@{} | L{3.7cm}@{\hspace{1mm}} | @{}C{2.0cm}@{} | @{}C{2.0cm}@{}}

\multicolumn{1}{c|}{\multirow{2}{*}{Type}} & \multicolumn{1}{c|}{\multirow{2}{*}{Methods}} & PF-WILLOW & PF-PASCAL  \\
\multicolumn{1}{c|}{} & \multicolumn{1}{c|}{} & ($\alpha_{\mathrm{bbox}}=0.1$)    & ($\alpha_{\mathrm{img}}=0.1$)     \\ 
\hline				
A &(T)~A2Net$_\textrm{res101}$~\cite{Seo2018AttentiveSA} & 68.8 & 70.8\\
A &(T)~CNNGeo$_\textrm{res101}$~\cite{rocco2017convolutional} & 69.2 & 71.9\\
A &(T+P)~WS-SA$_\textrm{res101}$~\cite{rocco2018end} & 70.2 & 75.8 \\
A &(P)~RTN$_\textrm{res101}$~\cite{kim2018recurrent} & 71.9 & 75.9 \\
\hline
F &PF-LOM$_\textrm{HOG}$~\cite{ham2016proposal} & 56.8 & 62.5 \\ 
F &(B+P)~PF-LOM$_\textrm{CAT-FCSS}$~\cite{kim2019fcss} & 58.4 & 68.9 \\
F &(P)~NCN$_\textrm{res101}$~\cite{rocco2018neighbourhood} & 67.0 & 78.9 \\
F &(C+P)~HPF$_\textrm{res101}$~\cite{min2019hyperpixel} & \textbf{74.4} & \underline{80.4} \\
F &(M)~Ours$_\textrm{res50}$ & 71.7 & 78.3 \\
F &(M)~Ours$_\textrm{res101}$ & \underline{73.5} & \textbf{81.9} \\
\hline
\end{tabular}
\end{center}
\label{tab:pf_data_bb}
\vspace{-0.3cm}
\end{table}	

\begin{center}
    \begin{table*}[t]
    \caption{Per-class PCK~($\alpha_{\mathrm{img}}=0.1$) on PF-PASCAL~\cite{ham2017proposal}. The results of~\cite{ham2016proposal} are taken from~\cite{rocco2018end}.}

        \begin{center}
            \scalebox{0.96}{
            \begin{tabular}{c|l|cccccccccccccccccccc|c}
            \hline
            \multicolumn{1}{c|}{Type} & \multicolumn{1}{c|}{Methods} & aero &	bike &	bird &	boat &	 bot &	bus &	car & cat & cha & cow & tab & dog & hor & mbik & pers & plnt & she & sofa & trai & tv & all\\
            \hline
            \hline
            A & (T)~A2Net$_\textrm{res101}$~\cite{Seo2018AttentiveSA} & 83.2 & 82.8 & 83.8 & 44.4 & 57.8 & 81.3 & 89.4 & 86.1 & 40.1 & \underline{91.7} & 21.4 & 73.2 & 33.8 & 76.3 & \textbf{74.3} & 63.3 & \textbf{100.0} & 45.5 & 45.3 & 60.0 & 70.8\\
            A & (T)~CNNGeo$_\textrm{res101}$~\cite{rocco2017convolutional} & 82.4 & 80.9 & \textbf{85.9} & 47.2 & 57.8 & 83.1 & 92.8 & 86.9 & 43.8 & \underline{91.7} & 28.1 & 76.4 & 70.2 & 76.6 & 68.9 & 65.7 & \underline{80.0} & 50.1 & 46.3 & 60.6 & 71.9 \\
            A & (T+P)~WS-SA$_\textrm{res101}$~\cite{rocco2018end} & 83.7 & 88.0 & 83.4 & 58.3 & 68.8 & \underline{90.3} & 92.3 & 83.7 & 47.4 & \underline{91.7} & 28.1 & 76.3 & \textbf{77.0} & 76.0 & 71.4 & \textbf{76.2} & \underline{80.0} & 59.5 & 62.3 & 63.9 & 75.8 \\
\hline
			F & PF-LOM$_\textrm{HOG}$~\cite{ham2016proposal} & 73.3 & 74.4 & 54.4 & 50.9 & 49.6 & 73.8 & 72.9 & 63.6 & 46.1 & 79.8 & 42.5 & 48.0 & 68.3 & 66.3 & 42.1 & 62.1 & 65.2 & 57.1 & 64.4 & 58.0 & 62.5  \\
            F & (P)~NCN$_\textrm{res101}$~\cite{rocco2018neighbourhood} & \underline{86.8} & 86.7 & \underline{86.7} & 55.6 & \underline{82.8} & 88.6 & 93.8 & \underline{87.1} & 54.3 & 87.5 & 43.2 & \underline{82.0} & 64.1 & 79.2 & 71.1 & \underline{71.0} & 60.0 & 54.2 & \underline{75.0} & \textbf{82.8} & 78.9 \\
            F & (C+P)~HPF$_\textrm{res101}$~\cite{min2019hyperpixel} & 86.5 & \underline{88.9} & 81.6 & \textbf{75.0} & 81.3 & 89.7 & \underline{93.7} & \textbf{87.6} & \underline{62.2} & 87.5 & \underline{52.6} & \textbf{87.5} & \underline{74.2} & \textbf{83.5} & \underline{73.5} & 66.2 & 60.0 & \underline{66.2} & 68.5 & 66.7 & \underline{80.4} \\
            F & Ours$_\textrm{res101}$ & \textbf{89.5} & \textbf{89.2} & 83.1 & \underline{73.6} & \textbf{85.9} & \textbf{92.6} & \textbf{95.0} & 83.7 & \textbf{65.6} & \textbf{93.8} & \textbf{53.6} & 81.3 & 71.6 & \underline{80.6} & 72.3 & \underline{71.0} & \textbf{100.0} & \textbf{69.3} & \textbf{80.0} & \underline{79.5} & \textbf{81.9} \\
            \hline
            \end{tabular}}
        \end{center}

	 \label{tab:pf_data_2}
    \end{table*}
\end{center}

\begin{center}
    \begin{table*}[t]
    \caption{Per-class PCK~($\alpha_{\mathrm{bbox}}=0.1$) on SPair-71k~\cite{min2019hyperpixel}. All numbers but ours are taken from~\cite{min2019hyperpixel}.}

        \begin{center}
            \scalebox{0.95}{
            \begin{tabular}{c|l|cccccccccccccccccc|c}
            \hline
             \multicolumn{2}{c|}{Methods} & aero & bike & bird & boat & bottle & bus & car & cat & chair & cow & dog & horse & moto & person & plant & sheep & train & tv & all\\
            \hline
            \hline
            \multirow{5}{*}{\shortstack[1]{Transferred \\ \\ models}}
            & (T)~CNNGeo$_\textrm{res101}$~\cite{rocco2017convolutional} & 21.3 & 15.1 & 34.6 & 12.8 & 31.2 & 26.3 & 24.0 & 30.6 & 11.6 & 24.3 & 20.4 & 12.2 & 19.7 & 15.6 & 14.3 & 9.6 & 28.5 & 28.8 & 18.1  \\
            & (T)~A2Net$_\textrm{res101}$~\cite{Seo2018AttentiveSA} &  20.8 & \underline{17.1} & 37.4 & 13.9 & 33.6 & \textbf{29.4} & \underline{26.5} & 34.9 & 12.0 & 26.5 & 22.5 & 13.3 & \underline{21.3} & 20.0 & 16.9 & 11.5 & 28.9 & 31.6 & 20.1 \\
            & (T+P)~WS-SA$_\textrm{res101}$~\cite{rocco2018end} & 23.4 & 17.0 & 41.6 & \underline{14.6} & \textbf{37.6} & \underline{28.1} & \textbf{26.6} & 32.6 & 12.6 & 27.9 & 23.0 & 13.6 & \underline{21.3} & \textbf{22.2} & \underline{17.9} & 10.9 & 31.5 & 34.8 & 21.1 \\
            & (P)~NCN$_\textrm{res101}$~\cite{rocco2018neighbourhood} & \underline{24.0} & 16.0 & \underline{45.0} & 13.7 & 35.7 & 25.9 & 19.0 & \underline{50.4} & \textbf{14.3} & \underline{32.6} & \textbf{27.4} & \textbf{19.2} & \textbf{21.7} & \underline{20.3} & \textbf{20.4} & \underline{13.6} & \textbf{33.6} & \textbf{40.4} & \textbf{26.4} \\
            & (M)~Ours$_\textrm{res101}$ & \textbf{27.3} & \textbf{17.2} & \textbf{47.2} & \textbf{14.7} & \underline{36.7} & 21.4 & 16.5 & \textbf{56.4} & \underline{13.6} & \textbf{32.9} & \underline{25.4} & \underline{17.4} & 19.9 & 19.5 & 15.9 & \textbf{15.9} & \underline{33.2} & \underline{35.1} & \underline{26.0} \\
            \hline
            \multirow{6}{*}{\shortstack[1]{SPair-71k \\ \\ trained \\ \\  models}}
            & (T)~CNNGeo$_\textrm{res101}$~\cite{rocco2017convolutional} &  23.4 & 16.7 & 40.2 & 14.3 & 36.4 & \underline{27.7} & 26.0 & 32.7 & 12.7 & 27.4 & 22.8 & 13.7 & 20.9 & 21.0 & 17.5 & 10.2 & 30.8 & 34.1 & 20.6  \\
            & (T)~A2Net$_\textrm{res101}$~\cite{Seo2018AttentiveSA} & 22.6 & \underline{18.5} & 42.0 & \textbf{16.4} & 37.9 & \textbf{30.8} & \underline{26.5} & 35.6 & 13.3 & 29.6 & 24.3 & 16.0 & \underline{21.6} & \underline{22.8} & \underline{20.5} & 13.5 & 31.4 & \textbf{36.5} & 22.3 \\
            & (T+P)~WS-SA$_\textrm{res101}$~\cite{rocco2018end} &  22.2 & 17.6 & 41.9 & 15.1 & \textbf{38.1} & 27.4 & \textbf{27.2} & 31.8 & 12.8 & 26.8 & 22.6 & 14.2 & 20.0 & 22.2 & 17.9 & 10.4 & \underline{32.2} & 35.1 & 20.9 \\
            & (P)~NCN$_\textrm{res101}$~\cite{rocco2018neighbourhood} & 17.9 & 12.2 & 32.1 & 11.7 & 29.0 & 19.9 & 16.1 & 39.2 & 9.9 & 23.9 & 18.8 & 15.7 & 17.4 & 15.9 & 14.8 & 9.6 & 24.2 & 31.1 & 20.1   \\
            & (C+P)~HPF$_\textrm{res101}$~\cite{min2019hyperpixel} & \underline{25.2} & \textbf{18.9} & \textbf{52.1} & \underline{15.7} & \underline{38.0} & 22.8 & 19.1 & \underline{52.9} & \textbf{17.9} & \underline{33.0} & \textbf{32.8} & \textbf{20.6} & \textbf{24.4} & \textbf{27.9} & \textbf{21.1} & \textbf{15.9} & 31.5 & 35.6 & \textbf{28.2} \\
            & (M)~Ours$_\textrm{res101}$ & \textbf{26.9} & 17.2 & \underline{45.5} & 14.7 & \underline{38.0} & 22.2 & 16.4 & \textbf{55.3} & \underline{13.5} & \textbf{33.4} & \underline{27.5} & \underline{17.7} & 20.8 & 21.1 & 16.6 & \underline{15.6} & \textbf{32.3} & \underline{35.9} & \underline{26.3} \\
            \hline
            \end{tabular}}
        \end{center}
	\label{tab:eval_spair}
    \end{table*}
\end{center} 

\vspace{-0.3cm}
\subsubsection{PF-WILLOW and PF-PASCAL}
The PF-WILLOW~\cite{ham2016proposal} and PF-PASCAL~\cite{ham2017proposal} datasets provide 900 and 1,351 image pairs of 4 and 20 image categories, respectively, with corresponding ground-truth object bounding boxes and keypoint annotations. The PF-PASCAL dataset is more challenging than other datasets~\cite{ham2016proposal,taniai2016joint} for semantic correspondence evaluation, featuring different instances of the same object class in the presence of large changes in appearance and scene layout, clutter and scale changes between objects. To evaluate our model, we use PF-WILLOW and the test split of PF-PASCAL provided by~\cite{ham2017proposal,rocco2018end} corresponding roughly 900 and 300 image pairs, respectively. 

We show in Table~\ref{tab:pf_data_bb} the average PCK scores for the PF-WILLOW and PF-PASCAL datasets, and compare our method with the state of the art. From this table, we observe five things: (1)~Our model outperforms the state of the art by a significant margin in terms of PCK, especially for the PF-PASCAL dataset. In particular, it shows better performance than other object-aware methods~\cite{ham2016proposal,kim2019fcss} that focus on establishing region correspondences between prominent objects. A plausible explanation is that establishing correspondences between object proposals is susceptible to shape deformations. (2)~We can clearly see that our model gives better results than semantic alignment methods on both datasets, but performance gain for the PF-PASCAL dataset, which typically contains pictures depicting a non-rigid deformation and clutter~(\eg,~in cow and sofa classes), is more significant. For example, the PCK gain over RTN~\cite{kim2018recurrent} for the PF-PASCAL~(81.9 vs. 75.9) is about four times more than that for the PF-WILLOW~(73.5 vs. 71.9), indicating that our semantic flow method is more robust to non-rigid deformations and background clutter than semantic alignment approaches. (3)~By comparing our model with CNN-based semantic flow methods, we can see that involving a spatial regularizer is significant. These techniques focus on designing fidelity terms~(\eg,~using a contrastive loss~\cite{choy2016universal,kim2019fcss}) to learn a feature space preserving semantic similarities. This is because of a lack of differentiability of the flow field. In contrast, our model gives a differentiable flow field, allowing to exploit a spatial regularizer while further leveraging high-level semantics from CNN features more specific to semantic correspondence. (4)~Similar to other CNN-based methods~\cite{rocco2018end,min2019hyperpixel,Seo2018AttentiveSA,kim2018recurrent}, our models with ResNet-50 ~\cite{he2016deep} show lower performance than the original ones using ResNet-101. (5)~We confirm once more a finding in~\cite{long2014convnets} that CNN features trained for ImageNet classification~\cite{deng2009imagenet} clearly show a better ability to handle intra-class variations than hand-crafted ones~(HOG~\cite{dalal2005histograms} in PF-LOM~\cite{ham2016proposal}). 

Table~\ref{tab:pf_data_2} shows per-class PCK scores on the PF-PASCAL dataset~\cite{ham2017proposal}. Our model achieves state-of-the-art results for 11 object categories, and outperforms all methods on average by a large margin. The performance gain is significant especially in the presence of non-rigid deformations~(\eg,~in cow and sheep classes) or distractions such as clutter~(\eg,~in table and sofa classes). This demonstrates once again that our method is able to establish reliable semantic correspondences of keypoints, even for images with large shape variations and clutter by which semantic alignment methods are easily distracted.

\subsubsection{SPair-71k}
The SPair-71k dataset~\cite{min2019hyperpixel}, a large-scale benchmark for semantic correspondence, provides 70,958 image pairs of 18 object categories with ground-truth annotations for object bounding boxes, segmentation masks, and keypoints. The image pairs in SPair-71k feature various changes in viewpoint, scale, truncation, and occlusion. Following the experimental protocol of~\cite{min2019hyperpixel}, we evaluate our model on the test split of 12,234 image pairs, and compute PCK scores with $\alpha_{\mathrm{bbox}}=0.1$ We show in Table~\ref{tab:eval_spair} the per-class and average PCK scores, and compare our model with the state of the art~\cite{rocco2017convolutional,Seo2018AttentiveSA,rocco2018end,rocco2018neighbourhood,min2019hyperpixel}. The first five rows show the PCK scores for models provided by authors, without retraining or finetuning on the SPair-71k dataset. We can see that our model achieves the second best performance, demonstrating that it generalizes well to unseen images. It is slightly outperformed by NCN~\cite{rocco2018neighbourhood}~(0.4\% in terms of the average PCK) but runs about 11 times faster at test time (Table~\ref{tab:inference_time}). The last six rows show the PCK scores for the models trained with SPair-71k. For fair comparison, we train our model with the training set of SPair-71k (986 images). The results show that it performs best in the presence of non-rigid deformations (\ie, in cat and cow classes). For other object categories, our model outperforms other CNN-based methods except for HPF~\cite{min2019hyperpixel}. Note that HPF exploits ground-truth correspondences at training time, which gives strong constraints but is extremely labor-intensive. In contrast, our model uses binary foreground masks only, that are widely available and much cheaper to obtain.

\begin{figure*}[t] % visual comparisons
\captionsetup[subfigure]{aboveskip=-0.5pt,belowskip=-0.5pt}
\centering
  \begin{subfigure}{0.16\textwidth} % bird / non-rigid deformation (beak of a bird)
  \includegraphics[width=\textwidth, height=0.917\textwidth, frame]{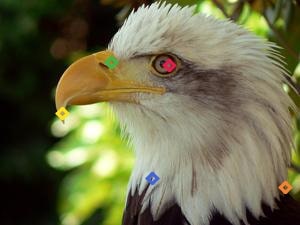}
  \end{subfigure}
  \begin{subfigure}{0.16\textwidth}
  \includegraphics[width=\textwidth, frame]{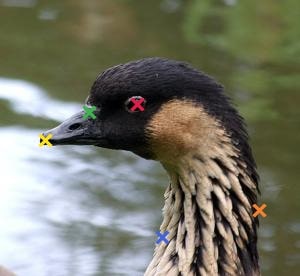}
  \end{subfigure}
  \begin{subfigure}{0.16\textwidth}
  \includegraphics[width=\textwidth, frame]{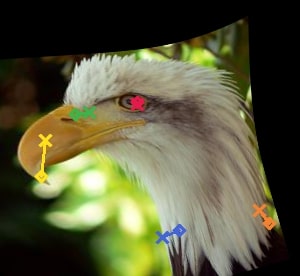}
  \end{subfigure}
% \begin{subfigure}{0.16\textwidth}
% \includegraphics[width=\textwidth, frame]{./fig/with_keypoints/PF-PASCAL_A2Net/0053.jpg}
% \end{subfigure}
  \begin{subfigure}{0.16\textwidth}
  \includegraphics[width=\textwidth, frame]{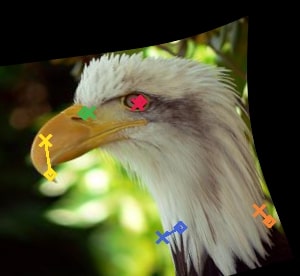}
  \end{subfigure}
  \begin{subfigure}{0.16\textwidth}
  \includegraphics[width=\textwidth, frame]{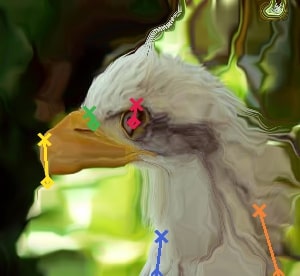}
  \end{subfigure}
  \begin{subfigure}{0.16\textwidth}
  \includegraphics[width=\textwidth, frame]{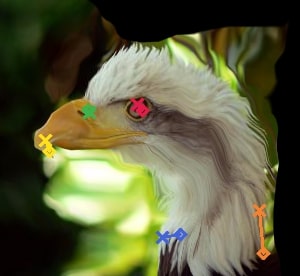}
  \end{subfigure}
  \vspace{0.05cm}

  \begin{subfigure}{0.16\textwidth} % horse / non-rigid deformation + DOF (head, forepaws)
  \includegraphics[width=\textwidth, height=0.685\textwidth, frame]{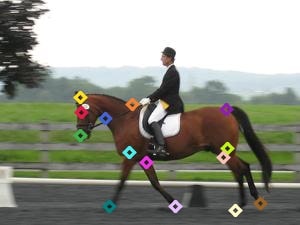}
  \end{subfigure}
  \begin{subfigure}{0.16\textwidth}
  \includegraphics[width=\textwidth, frame]{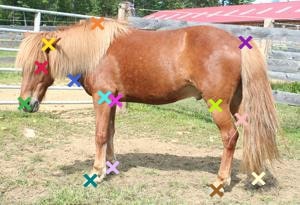}
  \end{subfigure}
  \begin{subfigure}{0.16\textwidth}
  \includegraphics[width=\textwidth, frame]{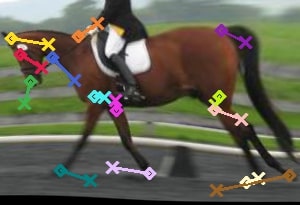}
  \end{subfigure}
% \begin{subfigure}{0.16\textwidth}s
% \includegraphics[width=\textwidth, frame]{./fig/with_keypoints/PF-PASCAL_A2Net/0198.jpg}
% \end{subfigure}
  \begin{subfigure}{0.16\textwidth}
  \includegraphics[width=\textwidth, frame]{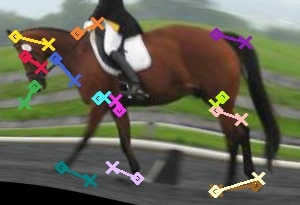}
  \end{subfigure}
  \begin{subfigure}{0.16\textwidth}
  \includegraphics[width=\textwidth, frame]{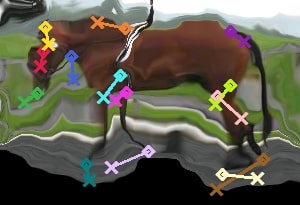}
  \end{subfigure}
  \begin{subfigure}{0.16\textwidth}
  \includegraphics[width=\textwidth, frame]{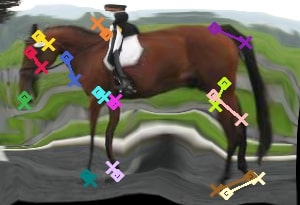}
  \end{subfigure}
  \vspace{0.05cm}
  
  \begin{subfigure}{0.16\textwidth} % bicycle / non-rigid deformation
  \includegraphics[width=\textwidth, height=1.25\textwidth, frame]{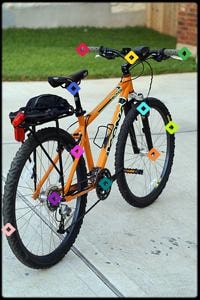}
  \end{subfigure}
  \begin{subfigure}{0.16\textwidth}
  \includegraphics[width=\textwidth, frame]{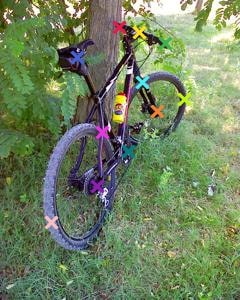}
  \end{subfigure}
  \begin{subfigure}{0.16\textwidth}
  \includegraphics[width=\textwidth, frame]{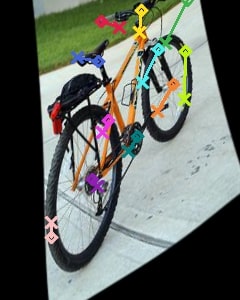}
  \end{subfigure}
% \begin{subfigure}{0.16\textwidth}
% \includegraphics[width=\textwidth, frame]{./fig/with_keypoints/PF-PASCAL_A2Net/0018.jpg}
% \end{subfigure}
  \begin{subfigure}{0.16\textwidth}
  \includegraphics[width=\textwidth, frame]{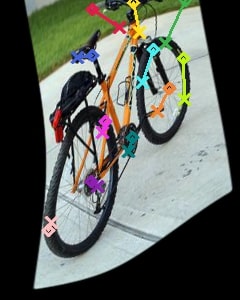}
  \end{subfigure}
  \begin{subfigure}{0.16\textwidth}
  \includegraphics[width=\textwidth, frame]{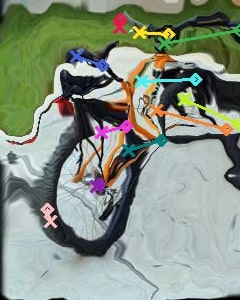}
  \end{subfigure}
  \begin{subfigure}{0.16\textwidth}
  \includegraphics[width=\textwidth, frame]{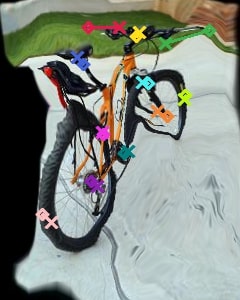}
  \end{subfigure}
  \vspace{0.05cm}
  
  \begin{subfigure}{0.16\textwidth} % bike
  \includegraphics[width=\textwidth, height=0.675\textwidth, frame]{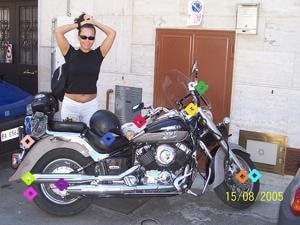}
  \end{subfigure}
  \begin{subfigure}{0.16\textwidth}
  \includegraphics[width=\textwidth, frame]{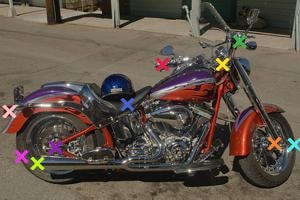}
  \end{subfigure}
  \begin{subfigure}{0.16\textwidth}
  \includegraphics[width=\textwidth, frame]{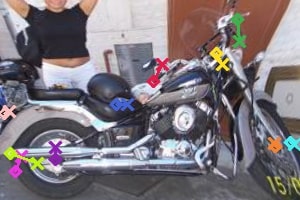}
  \end{subfigure}
% \begin{subfigure}{0.16\textwidth}
% \includegraphics[width=\textwidth, frame]{./fig/with_keypoints/PF-PASCAL_A2Net/0206.jpg}
% \end{subfigure}
  \begin{subfigure}{0.16\textwidth}
  \includegraphics[width=\textwidth, frame]{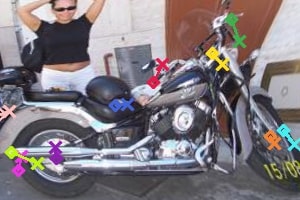}
  \end{subfigure}
  \begin{subfigure}{0.16\textwidth}
  \includegraphics[width=\textwidth, frame]{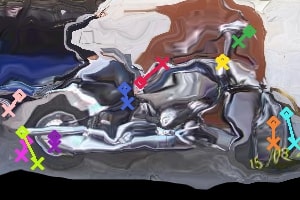}
  \end{subfigure}
  \begin{subfigure}{0.16\textwidth}
  \includegraphics[width=\textwidth, frame]{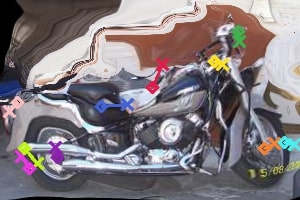}
  \end{subfigure}
  \vspace{0.05cm}

  \hspace{-0.185cm}
  \begin{subfigure}{0.16\textwidth} % bicycle
  \includegraphics[width=\textwidth, height=0.75\textwidth, frame]{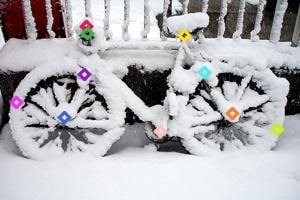}
  \vspace{-0.3cm}
  \caption*{Source image.}
  \end{subfigure}
  \begin{subfigure}{0.16\textwidth}
  \includegraphics[width=\textwidth, frame]{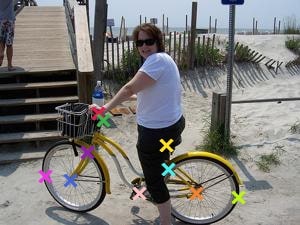}
  \vspace{-0.3cm}
  \caption*{Target image.}
  \end{subfigure}
  \begin{subfigure}{0.16\textwidth}
  \includegraphics[width=\textwidth, frame]{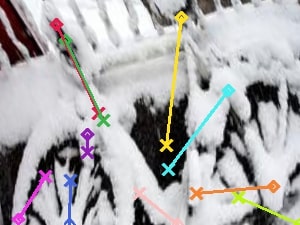}
  \vspace{-0.3cm}
  \caption*{CNNGeo~\cite{rocco2017convolutional}.}
  \end{subfigure}
% \begin{subfigure}{0.16\textwidth}
% \includegraphics[width=\textwidth, frame]{./fig/with_keypoints/PF-PASCAL_A2Net/0039.jpg}
% \vspace{-0.3cm}
% \caption*{A2Net~\cite{Seo2018AttentiveSA}.}
  \begin{subfigure}{0.16\textwidth}
  \includegraphics[width=\textwidth, frame]{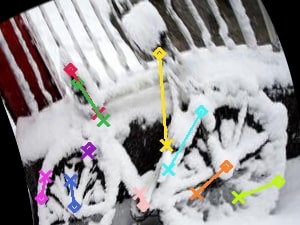}
  \vspace{-0.3cm}
  \caption*{WS-SA~\cite{rocco2018end}.}
  \end{subfigure}
  \begin{subfigure}{0.16\textwidth}
  \includegraphics[width=\textwidth, frame]{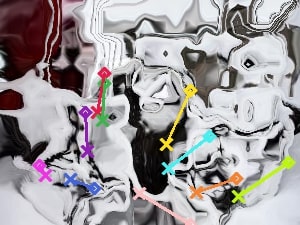}
  \vspace{-0.3cm}
  \caption*{HPF~\cite{min2019hyperpixel}.}
  \end{subfigure}
  \begin{subfigure}{0.16\textwidth}
  \includegraphics[width=\textwidth, frame]{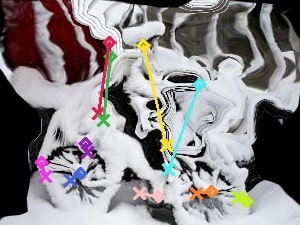}
  \vspace{-0.3cm}
  \caption*{Ours.}
  \end{subfigure}
    
  \vfill
  \vspace{-0.2cm}
\caption{Visual comparison of alignment results between source and target images on the PF-PASCAL dataset~\cite{ham2017proposal}. Keypoints in the source and target images are shown as diamonds and crosses, respectively, with a vector representing the matching error. All methods use ResNet-101 features. Compared to the state of the art, our method is more robust to local non-rigid deformations, scale changes between objects, and clutter. See text for details.~(Best viewed in color.)}
\label{fig:comparison_PF_PASCAL}  
\vspace{-0.3cm}
\end{figure*}

\setlength{\tabcolsep}{0.3em}
\begin{table}[t] % TSS
\caption{Quantitative comparison on the TSS dataset~\cite{taniai2016joint} in terms of the average PCK. We measure the PCK scores~($\alpha_{\mathrm{img}}=0.05$) on three subsets~(FG3DCar, JODS and PASCAL). All numbers except ours are taken from~\cite{rocco2018end,taniai2016joint,kim2018recurrent}.}
\small
\begin{center}
\begin{tabular}{@{}C{0.4cm}@{} | @{}C{0.4cm}@{} | L{3.8cm}@{\hspace{1mm}} | @{}C{1.05cm}@{} | @{}C{1.05cm}@{} | @{}C{1.05cm}@{}}
\multicolumn{2}{c|}{Type} & \multicolumn{1}{c|}{Methods}  & FG3D.    & JODS & PASC.  \\ 
\cmidrule{1-6}\morecmidrules
\hline
\multirow{6}{*}{\rb{Hand-crafted~}}
& F &DSP$_\textrm{SIFT}$~\cite{kim2013deformable} & 48.7 & 46.5 & 38.2\\
& F &DFF$_\textrm{DAISY}$~\cite{yang2014daisy} & 49.3 & 30.3 & 22.4\\
& F &SIFT Flow$_\textrm{SIFT}$~\cite{liu2011sift} & 63.4 & 52.2 & 45.3\\
& F &PF-LOM$_\textrm{HOG}$~\cite{ham2016proposal} & 78.6 & 65.3 & 53.1\\
& F &TSS$_\textrm{HOG}$~\cite{taniai2016joint} &	 83.0 &	59.5 & 48.3\\
& F &OADSC$_\textrm{HOG}$~\cite{yang2017object} &	87.5 &	70.8 & {\bf 72.9} \\
\hline
\multirow{7}{*}{\rb{CNN-based~~}}
& A &(T)~A2Net$_\textrm{vgg16}$~\cite{Seo2018AttentiveSA} & 87.0 & 67.0 & 55.0\\
& A &(T)~CNNGeo$_\textrm{res101}$~\cite{rocco2017convolutional} & 90.1 & 76.4 & 56.3\\
& A &(T+P)~WS-SA$_\textrm{res101}$~\cite{rocco2018end} &  \underline{90.3} & 76.4 & 56.5\\
& A &(P)~RTN$_\textrm{res101}$~\cite{kim2018recurrent} &	90.1 &	\underline{78.2} & \underline{63.3} \\
\cline{2-6}
& F &(B+P)~PF-LOM$_\textrm{FCSS}$~\cite{kim2017fcss} & 83.9 &63.5 & 58.2 \\
& F &(B+P)~DCTM$_\textrm{FCSS}$~\cite{kim2017dctm} &	89.1 &	72.1 & 61.0 \\
& F &(M)~Ours$_\textrm{res101}$ & {\bf 90.6} & {\bf 78.7} & 56.5 \\
\hline
\end{tabular}
%\captionsetup{font={small}}
\label{tab:eval_tss}	
\end{center}
\vspace{-0.3cm}
\end{table}

\subsubsection{TSS} \label{sec:tss}
The TSS dataset~\cite{taniai2016joint} consists of three subsets~(FG3DCar, JODS and PASCAL) that contain 400 image pairs of 7 object categories. It provides dense flow fields obtained by interpolating sparse keypoint matches with additional co-segmentation masks. Following the experimental protocol of~\cite{rocco2018end}, we compute the PCK scores~($\alpha_{\mathrm{img}}=0.05$) densely over the foreground object. Table~\ref{tab:eval_tss} compares the average PCK on each subset in the TSS dataset. Our method shows better performance than the state of the art for FG3DCar and JODS. We do not do as well on the PASCAL part of TSS, which contains many image pairs with different poses~(\eg,~cars captured with left- and right-side viewpoints). Current methods, except for OADSC~\cite{yang2017object} that is specially designed for handling changes in viewpoint, have a limited capability of finding matches between images with different poses. Ours is no exception.

\begin{figure}[h] % PF-WILLOW / PF-PASCAL / TSS
\captionsetup[subfigure]{aboveskip=-0.5pt,belowskip=-0.5pt}
\centering
	\begin{subfigure}{0.45\textwidth}
	\includegraphics[width=\textwidth, frame]{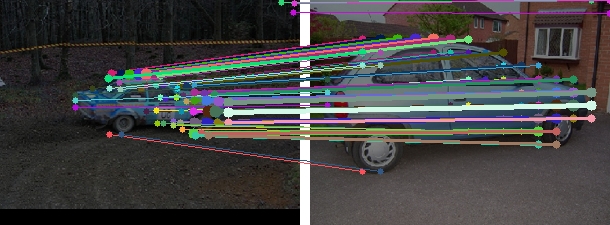}
	\end{subfigure}

%	\vspace{0.05cm}
%	\begin{subfigure}{0.45\textwidth}
%	\includegraphics[width=\textwidth, frame]{./fig/PF-WILLOW/0405corr.jpg}
%	\end{subfigure}

	\vspace{0.05cm}
	\begin{subfigure}{0.45\textwidth}
	\includegraphics[width=\textwidth, frame]{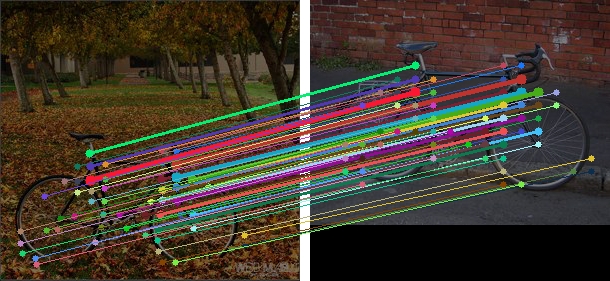}
	\end{subfigure}
	
%	\vspace{0.05cm}
%	\begin{subfigure}{0.45\textwidth}
%	\includegraphics[width=\textwidth, frame]{./fig/PF-PASCAL/0185corr.jpg}
%	\end{subfigure}
	
%	\vspace{0.05cm}
%	\begin{subfigure}{0.45\textwidth}
%	\includegraphics[width=\textwidth, frame]{./fig/TSS/0685corr.jpg}
%	\end{subfigure}
%	
	\vspace{0.05cm}
	\begin{subfigure}{0.45\textwidth}
	\includegraphics[width=\textwidth, frame]{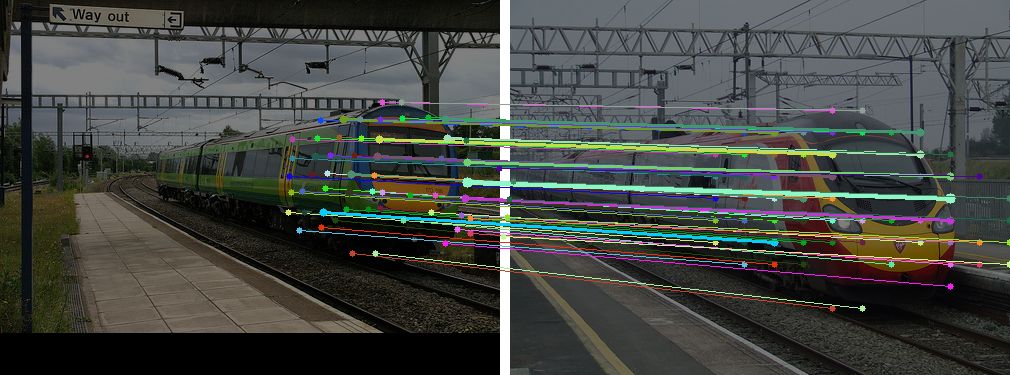}
	\vspace{-0.2cm}
	\end{subfigure}

	\vfill
	\vspace{-0.2cm}
\caption{Top matches on standard benchmarks. We visualize the top 60 matches according to matching probabilities. Each row shows a result on PF-WILLOW~\cite{ham2016proposal}, PF-PASCAL~\cite{ham2017proposal} and TSS~\cite{taniai2016joint}, respectively.~(Best viewed in color.)}
\label{fig:qualitative_PF_TSS}	
\end{figure}

\begin{figure}[t]
\captionsetup[subfigure]{aboveskip=-0.5pt,belowskip=-0.5pt}
\centering
	\begin{subfigure}{0.115\textwidth}
	\includegraphics[width=\textwidth, frame]{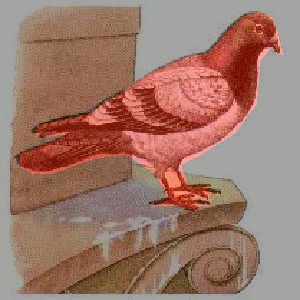}
	\end{subfigure}
	\begin{subfigure}{0.115\textwidth}
	\includegraphics[width=\textwidth, height=1.3\textwidth, frame]{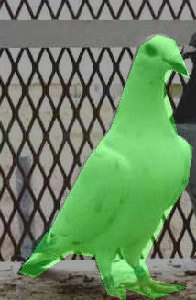}
	\end{subfigure}
	\begin{subfigure}{0.115\textwidth}
	\includegraphics[width=\textwidth, height=1.3\textwidth, frame]{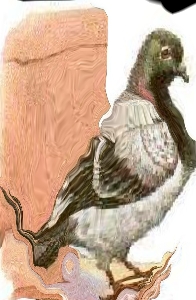}
	\end{subfigure}
	\begin{subfigure}{0.115\textwidth}
	\includegraphics[width=\textwidth, height=1.3\textwidth, frame]{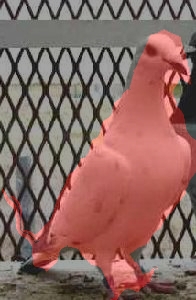}
	\end{subfigure}
	
	\vspace{0.05cm}
	\begin{subfigure}{0.115\textwidth}
	\includegraphics[width=\textwidth, frame]{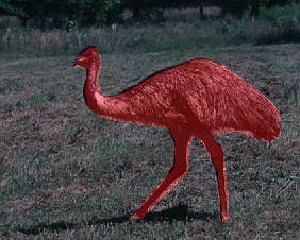}
	\end{subfigure}
	\begin{subfigure}{0.115\textwidth}
	\includegraphics[width=\textwidth, frame]{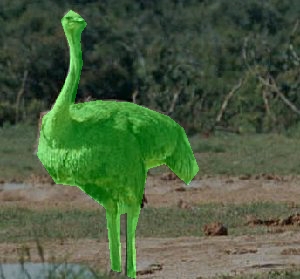}
	\end{subfigure}
	\begin{subfigure}{0.115\textwidth}
	\includegraphics[width=\textwidth, frame]{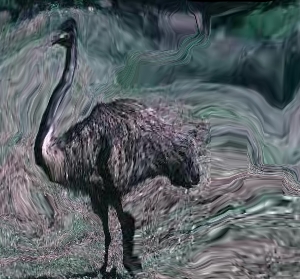}
	\end{subfigure}
	\begin{subfigure}{0.115\textwidth}
	\includegraphics[width=\textwidth, frame]{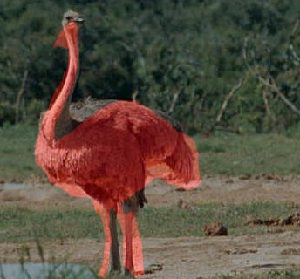}
	\end{subfigure}

	\vspace{0.05cm}
	\begin{subfigure}{0.115\textwidth}
	\includegraphics[width=\textwidth, frame]{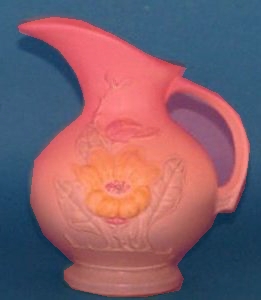}
	\vspace{-0.25cm}
	\caption*{Source image.}
	\end{subfigure}
	\begin{subfigure}{0.115\textwidth}
	\includegraphics[width=\textwidth, frame]{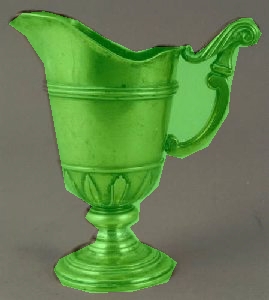}
	\vspace{-0.2cm}
	\caption*{Target image.}
	\end{subfigure}
	\begin{subfigure}{0.115\textwidth}
	\includegraphics[width=\textwidth, frame]{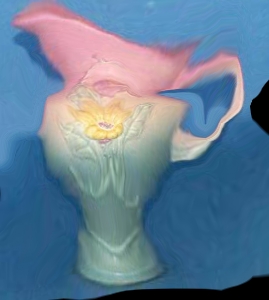}
	\vspace{-0.2cm}
	\caption*{Alignment.}
	\end{subfigure}
	\begin{subfigure}{0.115\textwidth}
	\includegraphics[width=\textwidth, frame]{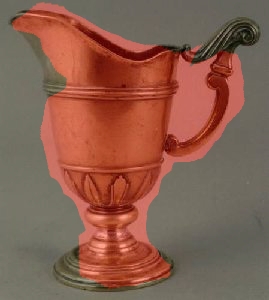}
	\vspace{-0.2cm}
	\caption*{Label transfer.}
	\end{subfigure}
	\vfill
\vspace{-0.2cm}
\caption{Alignment and label transfer examples on the Caltech-101 dataset~\cite{fei2006one}. The source and target masks are overlaid on the corresponding images. We transfer pixel labels of the source images to the target ones using established correspondences. We show label transfer results overlaid on target images. (Best viewed in color.)}
\label{fig:qualitative_Caltech}	
\end{figure}

\subsubsection{Runtime Analysis}
Table~\ref{tab:inference_time} shows runtime comparisons of state-of-the-art methods. For comparison, we run the original source codes implemented using $\texttt{PyTorch}$\cite{paszke2017automatic}. The average runtime is measured on the same machine with a NVIDIA Titan RTX GPU. The table shows that our model is fastest among the state of the art. Semantic alignment methods~\cite{rocco2017convolutional,rocco2018end,Seo2018AttentiveSA} estimate parameters of affine and thin place spline sequentially, which degrades runtime performance. Semantic flow methods involve 4-D convolutions~\cite{rocco2018neighbourhood} or a Hough voting process~\cite{min2019hyperpixel} on top of the correlation volume, requiring additional computations to establish pixel-level correspondences. Our model, on the other hand, simply assigns the best matches from the correlation volume in a single stage. Most computation time is spent extracting features~(23.7 milliseconds). Computing matching scores and establishing correspondences using the kernel soft argmax just take 1.2 milliseconds.

\begin{table}[t] % runtime
	\caption{Runtime comparison per image pair on the test split of the PF-PASCAL dataset~\cite{ham2017proposal,han2017scnet} in milliseconds.}
	\small
	\begin{center}
	\begin{tabular}{@{}C{1.2cm}@{} | L{3.2cm}@{\hspace{1mm}} | @{}C{1.75cm}@{}}
	
	Type & \multicolumn{1}{c|}{Methods}  & Time~($\textit{ms}$)    \\ 
	\cmidrule{1-3}\morecmidrules
	\hline
	A& (T)~CNNGeo$_\textrm{res101}$~\cite{rocco2017convolutional} & 34.2\\
	A& (T+P)~WS-SA$_\textrm{res101}$~\cite{rocco2018end} & 34.4\\
	A& (T)~A2Net$_\textrm{res101}$~\cite{Seo2018AttentiveSA} & 61.2\\
	%F& PF-LOM$_\textrm{HOG}$~\cite{ham2017proposal} & >1000\\
	%F& SCNet$_\textrm{vgg16}$~\cite{han2017scnet} & >1000\\
	\hline
	F& (P)~NCN$_\textrm{res101}$~\cite{rocco2018neighbourhood} & 284.2\\
	F& (C+P)~HPF$_\textrm{res101}$~\cite{min2019hyperpixel} & 48.9\\
	F& (M)~Ours$_\textrm{res101}$ & \textbf{24.9}\\
	\hline
	\end{tabular}
	\end{center}
	\label{tab:inference_time}
	\vspace{-0.3cm}
\end{table}

\subsubsection{Qualitative Results}

Figure~\ref{fig:comparison_PF_PASCAL} shows a visual comparison of alignment results between source and target images with the state of the art on the test split of PF-PASCAL~\cite{ham2017proposal,rocco2018end}. To this end, the source images are warped to the target images using the dense flow fields computed by each method. We can see that our method is robust to local non-rigid deformation~(\eg,~bird beaks and horse legs in the first two rows), scale changes between objects~(\eg,~front wheels in the third row), and clutter~(\eg,~wheels in the last row), while semantic alignment methods~\cite{rocco2017convolutional,rocco2018end} are not. In particular, the fourth example clearly shows that our method gives more discriminative correspondences, cutting off matches for non-common objects. For example, it does not establish correspondences between a person and background regions in the source and target images, respectively, while CNNGeo~\cite{rocco2017convolutional} and WS-SA~\cite{rocco2018end} fail to cut off matches on these regions. We can also see that all methods do not establish correspondences for occluded regions~(\eg,~a bicycle saddle in the last row). We also show in Fig.~\ref{fig:qualitative_PF_TSS} the top 60 matches chosen according to matching probabilities on the PF-WILLOW~\cite{ham2016proposal}, PF-PASCAL~\cite{ham2017proposal}, and TSS~\cite{taniai2016joint} datasets. We can see that most strong matches are established between prominent objects, and matches between foreground and background regions have low matching probabilities.

\begin{table}[t] % caltech-101
%\vspace{-2mm}
\caption{Quantitative comparison on the Caltech-101 dataset~\cite{fei2006one}. All numbers but ours are taken from~\cite{ham2017proposal,rocco2018end,Seo2018AttentiveSA,min2019hyperpixel}.}
%\vspace{-0.3cm}
\small
\begin{center}
\begin{tabular}{@{}C{0.5cm}@{} | @{}C{0.5cm}@{} | L{4.0cm}@{\hspace{1mm}} | @{}C{1.4cm}@{} | @{}C{1.4cm}@{}}

\multicolumn{2}{c|}{Type} & \multicolumn{1}{c|}{Methods}  & LT-ACC    & IoU     \\ 
\cmidrule{1-5}\morecmidrules
\hline
\multirow{6}{*}{\rb{Hand-crafted}}
&F &DeepFlow$_\textrm{SIFT}$~\cite{revaud2016deepmatching} & 0.74 & 0.40\\
&F &GMK$_\textrm{SIFT}$~\cite{duchenne2011graph} & 0.77 & 0.42\\
&F &SIFT Flow$_\textrm{SIFT}$~\cite{liu2011sift} & 0.75 & 0.48 \\
&F &DSP$_\textrm{SIFT}$~\cite{kim2013deformable} & 0.77 & 0.47 \\
&F &PF-LOM$_\textrm{HOG}$~\cite{ham2017proposal} & 0.78 & 0.50\\
&F&OADSC$_\textrm{HOG}$~\cite{yang2017object} & 0.81 & 0.55\\
\hline
\multirow{8}{*}{\rb{CNN-based}}
&A&(T)~A2Net$_\textrm{vgg16}$~\cite{Seo2018AttentiveSA} & 0.80 & 0.57\\
&A&(T)~CNNGeo$_\textrm{res101}$~\cite{rocco2017convolutional} & 0.83 & 0.61\\
&A&(T+P)~WS-SA$_\textrm{res101}$~\cite{rocco2018end} & 0.85 & \underline{0.63} \\
\cline{2-5}
&F&(B+P)~PF-LOM$_\textrm{FCSS}$~\cite{kim2017fcss} & 0.83 &0.52 \\
&F&(C+P)~SCNet-AG$_\textrm{vgg16}$~\cite{han2017scnet} & 0.79 & 0.51 \\
&F&(P)~NCN$_\textrm{res101}$~\cite{rocco2018neighbourhood} & 0.85 & 0.60 \\
&F&(C+P)~HPF$_\textrm{res101}$~\cite{min2019hyperpixel} & \underline{0.87} & \underline{0.63} \\
&F&(M)~Ours$_\textrm{res101}$ & {\bf 0.88} & {\bf 0.67} \\
\hline
\end{tabular}
\end{center}
\label{tab:eval_caltech}
\end{table}

\begin{figure*}[ht] % visual comparisons
\captionsetup[subfigure]{aboveskip=-0.5pt,belowskip=-0.5pt}
\centering

	\begin{subfigure}{0.126\textwidth} % child / background clutter
		\includegraphics[width=\textwidth, frame]{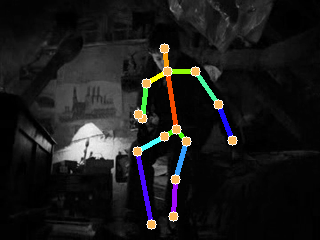}
		\vspace{-0.3cm}
	\end{subfigure}
	\begin{subfigure}{0.42\textwidth}
		\hspace{0.35cm}
		\includegraphics[width=0.3\textwidth, frame]{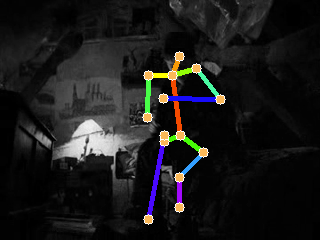}
		\includegraphics[width=0.3\textwidth, frame]{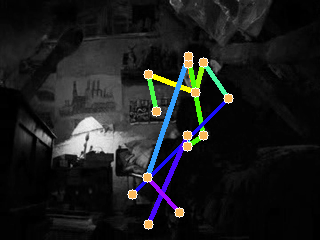}
		\includegraphics[width=0.3\textwidth, frame]{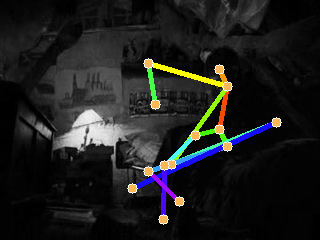}
		\vspace{-0.3cm}
	\end{subfigure}
	\begin{subfigure}{0.42\textwidth}
		\hspace{0.35cm}
		\includegraphics[width=0.3\textwidth, frame]{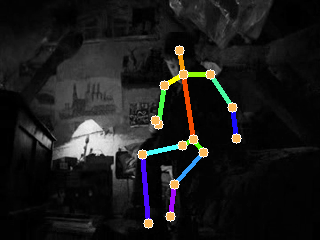}
		\includegraphics[width=0.3\textwidth, frame]{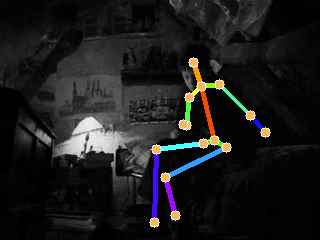}
		\includegraphics[width=0.3\textwidth, frame]{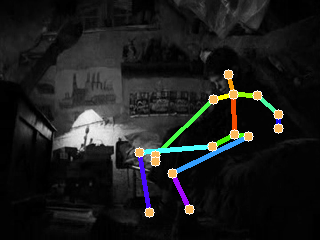}
		\vspace{-0.3cm}
	\end{subfigure}
	
	\begin{subfigure}{0.126\textwidth} % Large moving
		\includegraphics[width=\textwidth, frame]{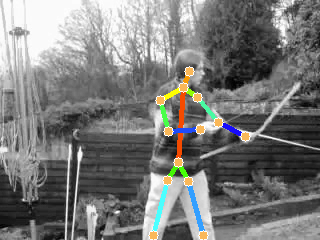}
		\vspace{-0.3cm}
	\end{subfigure}
	\begin{subfigure}{0.42\textwidth}
		\hspace{0.35cm}
		\includegraphics[width=0.3\textwidth, frame]{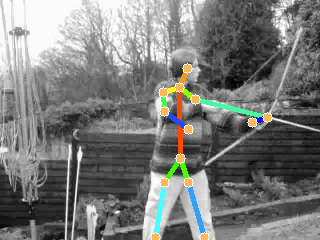}
		\includegraphics[width=0.3\textwidth, frame]{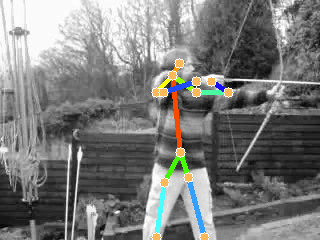}
		\includegraphics[width=0.3\textwidth, frame]{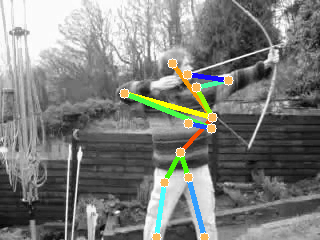}
		\vspace{-0.3cm}
	\end{subfigure}
	\begin{subfigure}{0.42\textwidth}
		\hspace{0.35cm}
		\includegraphics[width=0.3\textwidth, frame]{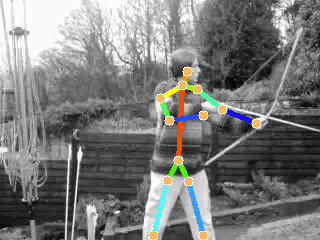}
		\includegraphics[width=0.3\textwidth, frame]{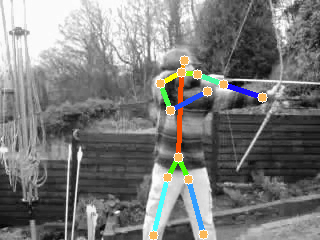}
		\includegraphics[width=0.3\textwidth, frame]{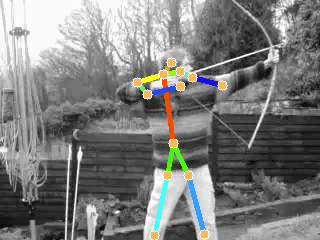}
		\vspace{-0.3cm}
	\end{subfigure}
	
	\begin{subfigure}{0.126\textwidth} % Occlusion
		\includegraphics[width=\textwidth, frame]{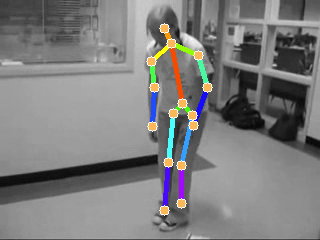}
		\vspace{-0.3cm}
	\end{subfigure}
	\begin{subfigure}{0.42\textwidth}
		\hspace{0.35cm}
		\includegraphics[width=0.3\textwidth, frame]{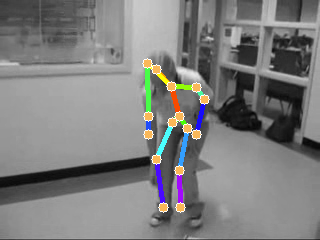}
		\includegraphics[width=0.3\textwidth, frame]{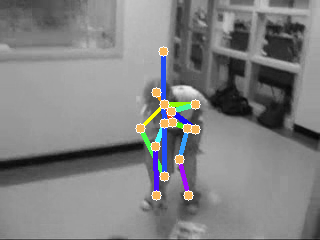}
		\includegraphics[width=0.3\textwidth, frame]{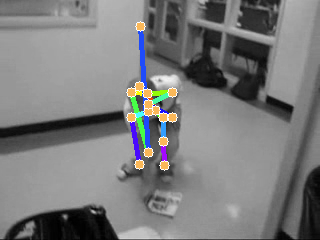}
		\vspace{-0.3cm}
	\end{subfigure}
	\begin{subfigure}{0.42\textwidth}
		\hspace{0.35cm}
		\includegraphics[width=0.3\textwidth, frame]{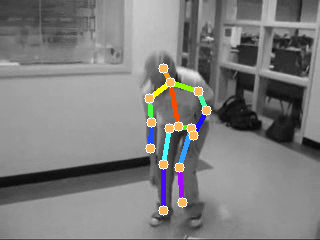}
		\includegraphics[width=0.3\textwidth, frame]{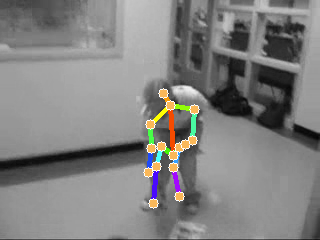}
		\includegraphics[width=0.3\textwidth, frame]{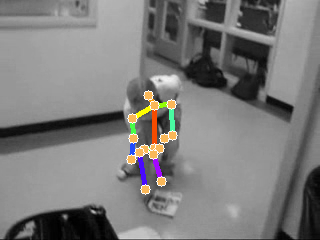}
		\vspace{-0.3cm}
	\end{subfigure}
	
	\begin{subfigure}{0.126\textwidth} % Occlusion
		\includegraphics[width=\textwidth, frame]{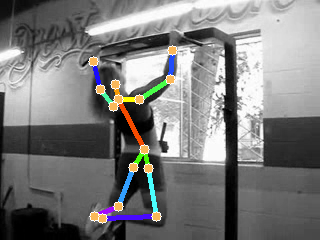}
		\vspace{-0.3cm}
		\caption*{Input.}
	\end{subfigure}
	\begin{subfigure}{0.42\textwidth}
		\hspace{0.35cm}
		\includegraphics[width=0.3\textwidth, frame]{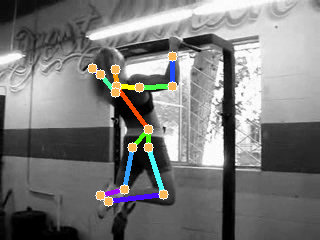}
		\includegraphics[width=0.3\textwidth, frame]{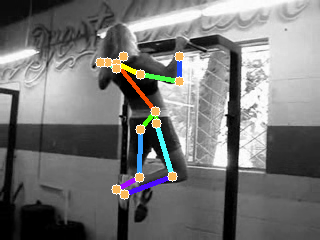}
		\includegraphics[width=0.3\textwidth, frame]{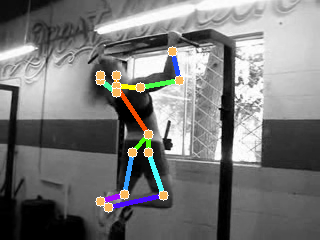}
		\vspace{-0.3cm}
		\caption*{TimeCycle$_\textrm{res50}$~\cite{wang2019learning}.}
	\end{subfigure}
	\begin{subfigure}{0.42\textwidth}
		\hspace{0.35cm}
		\includegraphics[width=0.3\textwidth, frame]{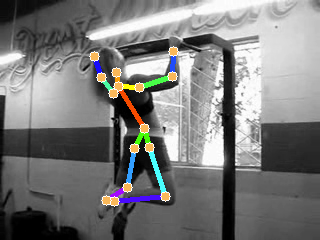}
		\includegraphics[width=0.3\textwidth, frame]{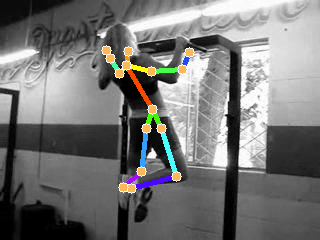}
		\includegraphics[width=0.3\textwidth, frame]{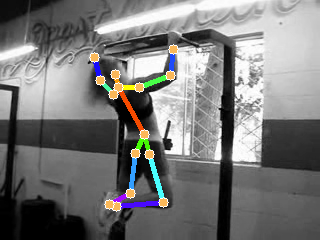}
		\vspace{-0.3cm}
		\caption*{Ours$_\textrm{res50}$.}
	\end{subfigure}

\caption{Visual comparison of keypoint propagation results on the JHMDB dataset~\cite{jhuang2013towards}. Given an input labeled with the ground-truth pose keypoints, we propagate them through video sequences. Compared to the state of the art, our method is more robust to background clutter, large displacements, and occlusion. The keypoints are shown in circles. (Best viewed in color.)}
\label{fig:qualitative_JHMDB}	
%\vspace{-0.3cm}
\end{figure*}

\subsection{Mask Transfer}\label{sec:mask_transfer_results}
We apply our model to the task of mask transfer on the Caltech-101~\cite{fei2006one} dataset. This dataset, originally introduced for image classification, provides pictures of 101 image categories with ground-truth object masks. Unlike the PF~\cite{ham2016proposal,ham2017proposal} and TSS~\cite{taniai2016joint} datasets, it does not provide ground-truth keypoint annotations. For fair comparison, we use 15 image pairs, provided by~\cite{han2017scnet,rocco2018end}, for each object category, and use the corresponding 1,515 image pairs for evaluation. Following the experimental protocol in~\cite{kim2013deformable}, we compute matching accuracy with two metrics using the ground-truth masks: Label transfer accuracy~(LT-ACC) and the intersection-over-union~(IoU) metric. Both metrics count the number of correctly labeled pixels between ground-truth and transformed masks using dense correspondences, where the LT-ACC evaluates the overall matching quality while the IoU metric focuses more on foreground objects. Following~\cite{Seo2018AttentiveSA,rocco2018end}, we exclude the LOC-ERR metric, since it measures the localization error of correspondences using object bounding boxes due to a lack of keypoint annotations, which does not cover rotations, affine or deformable transformations.

The LT-ACC and IoU comparisons on the Caltech-101 dataset are shown in Table~\ref{tab:eval_caltech}. Although this dataset provides ground-truth object masks, we do not retrain or fine-tune our model to evaluate its generalization ability on other datasets. From this table, we can see that (1)~our model generalizes better than other CNN-based methods for other images outside the training dataset; and~(2) it outperforms the state of the art in terms of the LT-ACC and IoU, verifying once more that our model focuses on regions containing objects while filtering out background clutter, even without using object proposals~\cite{ham2017proposal,han2017scnet,yang2017object,kim2017fcss} or inlier counting~\cite{rocco2018end}. In Fig.~ \ref{fig:qualitative_Caltech}, we show alignment and label transfer examples on the Caltech-101~\cite{fei2006one} dataset. We can see that our method is robust against local non-rigid deformations~(\eg,~bird's neck, body, and legs).

\begin{table}[t] 
\caption{Quantitative comparison on the JHMDB dataset~\cite{jhuang2013towards}. All numbers except for HPF~\cite{min2019hyperpixel} are taken from~\cite{wang2019learning}. Identity: copying labels in the first frame; $^{\dag}$: models trained from scratch; V: unlabeled video frames.}
\small
\begin{center}
\begin{tabular}{L{4.5cm}@{\hspace{1mm}} | @{}C{2cm}@{} | @{}C{2cm}@{}}
\multicolumn{1}{c|}{\multirow{2}{*}{Methods}}  & PCK    			 & PCK   \\
													& ($\alpha_{\mathrm{bbox}}=0.1$)  & ($\alpha_{\mathrm{bbox}}=0.2$)     \\
\cmidrule{1-3}\morecmidrules
\hline
Identity & 43.1 & 64.5\\
SIFT Flow$_\textrm{SIFT}$~\cite{liu2011sift} & 49.0 & 68.6\\
(V+C)~Optical Flow$_\textrm{flownet2}$$^{\dag}$~\cite{ilg2017flownet} & 45.2 & 62.9\\
(V)~Video Colorizaiton$_\textrm{3d-res18}$$^{\dag}$~\cite{vondrick2018tracking} & 45.2 & 69.6\\
(V)~TimeCycle$_\textrm{res18}$$^{\dag}$~\cite{wang2019learning} & 57.3 & 78.1\\
%(V)~ResNet-50+TimeCycle$^{\star}$~\cite{wang2019learning} & 0.567 & 0.761\\
(V)~TimeCycle$_\textrm{res50}$$^{\dag}$~\cite{wang2019learning} & 57.7 & 78.5\\
\hline
(V)~TimeCycle$_\textrm{res50}$~\cite{wang2019learning} & 58.4 & 78.4\\
(C+P)~HPF$_\textrm{res101}$~\cite{min2019hyperpixel} & 58.7 & 76.8\\
(M)~Ours$_\textrm{res50}$ & \underline{59.9} & \underline{78.9}\\
(M)~Ours$_\textrm{res101}$ & \bf 61.0 & \bf 80.6\\
\hline
\end{tabular}
\label{tab:eval_jhmdb}	
\end{center}
\vspace{-0.3cm}
\end{table}

\subsection{Pose Keypoint Propagation} \label{sec:tracking_results}
We apply our model to the task of keypoint propagation on the JHMDB~\cite{jhuang2013towards} dataset. We propagate ground-truth pose keypoints in the first frame to subsequent ones by estimating semantic correspondences between them. The JHMDB~\cite{jhuang2013towards} dataset contains 928 clips of 21 action categories with pose keypoints, segmentation masks of humans in action, obtained by a 2D articulated human puppet model~\cite{zuffi2012pictorial}, and provides three splits, where each split consists of training and test sets. Following the experimental protocol of~\cite{wang2019learning,vondrick2018tracking}, we test our model on the test set in the split 1 corresponding to 268 clips of action categories, without retraining or fine-tuning on the dataset. We normalize keypoint coordinates in the range of [0,1] by dividing them with the height and width of the human bounding box size, respectively, and use the PCK score with two threshold values~($\alpha_{\mathrm{bbox}}=0.1,0.2$) for evaluation.

%The results marked with a star~($^{\star}$) are obtained by the official model~\cite{wang2019learning} provided by the authors.
We show in Table~\ref{tab:eval_jhmdb} the average PCK scores for the keypoint propagation task, and compare our method with the state of the art including self-supervised methods~\cite{wang2019learning,vondrick2018tracking}. From this table, we can see that our model based on ResNet-50~\cite{he2016deep} outperforms the state of the art, even without using video datasets for training. For example, TimeCycle~\cite{wang2019learning} is trained with the VLOG~\cite{fouhey2018lifestyle} dataset that contains 114k videos with the total length of 344 hours. Training networks with such video datasets requires lots of computational resources and training time. We can also see that our model outperforms HPF~\cite{min2019hyperpixel}, demonstrating once again its generalization ability to unseen images during training. Figure~\ref{fig:qualitative_JHMDB} shows a visual comparison of keypoint propagation results with the TimeCycle~\cite{wang2019learning} method on the JHMDB~\cite{jhuang2013towards} dataset. The qualitative results for the comparison have been obtained from the original model~\cite{wang2019learning} provided by the authors. We predict the keypoints in the rest of the videos, by propagating ground truth in the first frame. We can see that our method is more robust to background clutter~(\eg,~body parts in the first row) and large displacements~(\eg,~elbows and wrists in the second row). Moreover, it is not seriously affected by occlusion~(\eg,~ankles and wrists in the last two rows) as the smoothness term regularizes flow fields within prominent objects.

\begin{figure}[t]
	\vspace{-0.1cm}
    \captionsetup[subfigure]{aboveskip=0.5pt,belowskip=0.5pt}
    \centering
        \hspace{-0.15cm}
		\begin{subfigure}{0.118\textwidth}
			\captionsetup{font={footnotesize}}
            \caption*{Input images.}
            \vspace{0.15cm}
            \includegraphics[width=\textwidth, frame]{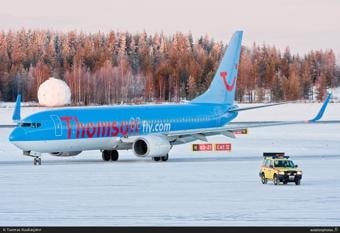}
			\vspace{-0.485cm}
			\captionsetup{font={footnotesize}}
            \caption*{Source.}
        \end{subfigure}
		\begin{subfigure}{0.118\textwidth}
			\captionsetup{font={footnotesize}}
            \caption*{Predicted masks.}
            \vspace{0.15cm}
            \includegraphics[width=\textwidth, frame]{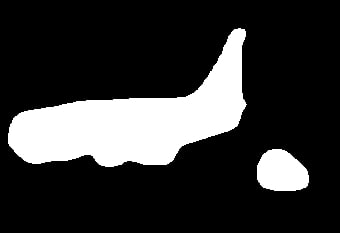}
			\vspace{-0.485cm}
			\captionsetup{font={scriptsize}}
            \caption*{$\tilde{M}^s$.}
        \end{subfigure}
		\begin{subfigure}{0.118\textwidth}
			\captionsetup{font={footnotesize}}
            \caption*{Warped masks.}
            \vspace{0.15cm}
            \includegraphics[width=\textwidth, frame]{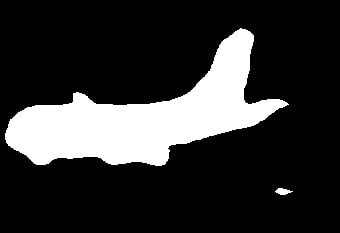}
			\vspace{-0.485cm}
			\captionsetup{font={scriptsize}}
            \caption*{$\mathcal{W}(\tilde{M}^t;\mathcal{F}^s)$.}
        \end{subfigure}
		\begin{subfigure}{0.118\textwidth}
			\captionsetup{font={footnotesize}}
            \caption*{Co-segmentation.}
            \vspace{0.15cm}
            \includegraphics[width=\textwidth, frame]{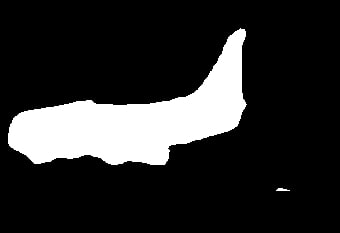}
			\vspace{-0.485cm}
			\captionsetup{font={scriptsize}}
            \caption*{$\tilde{M}^s\odot\mathcal{W}(\tilde{M}^t;\mathcal{F}^s)$.}
        \end{subfigure}
		
		\hspace{-0.15cm}
		\begin{subfigure}{0.118\textwidth}
			\vspace{0.08cm}
            \includegraphics[width=\textwidth, frame]{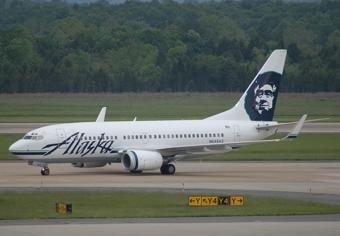}
			\vspace{-0.3cm}
			\captionsetup{font={footnotesize}}
            \caption*{Target.}
		\end{subfigure}
		\begin{subfigure}{0.118\textwidth}
			\vspace{0.08cm}
            \includegraphics[width=\textwidth, frame]{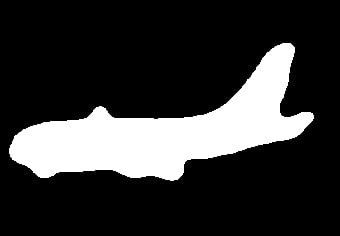}
			\vspace{-0.3cm}
			\captionsetup{font={scriptsize}}
            \caption*{$\tilde{M}^t$.}
		\end{subfigure}
		\begin{subfigure}{0.118\textwidth}
			\vspace{0.08cm}
            \includegraphics[width=\textwidth, frame]{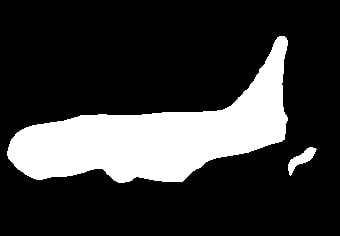}
			\vspace{-0.3cm}
			\captionsetup{font={scriptsize}}
            \caption*{$\mathcal{W}(\tilde{M}^s;\mathcal{F}^t)$.}
		\end{subfigure}
		\begin{subfigure}{0.118\textwidth}
			\vspace{0.08cm}
            \includegraphics[width=\textwidth, frame]{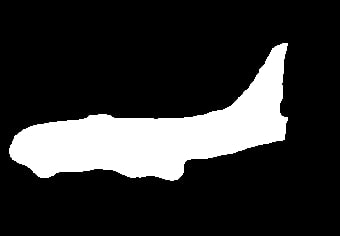}
			\vspace{-0.3cm}
			\captionsetup{font={scriptsize}}
            \caption*{$\tilde{M}^t\odot\mathcal{W}(\tilde{M}^s;\mathcal{F}^t)$.}
        \end{subfigure}
        \vfill
        \vspace{-0.15cm}
	\caption{A visual example of predicting co-segmentation masks. We extract common objects between source and target images by element-wise multiplication between predicted and warped masks.}
	\vspace{-0.15cm}
    \label{fig:coseg}	
\end{figure}

\subsection{Object Co-segmentation} \label{sec:coseg}
We apply our model to the task of object co-segmentation on the TSS dataset~\cite{taniai2016joint}. To this end, we add a mask regression layer on top of each feature extractor in Fig.~\ref{fig:proposed-met}. It inputs source or target feature maps and predicts binary foreground masks. We train adaptation and mask regression layers from scratch with three loss terms in Eq.~\eqref{eq:total_loss} and an additional mask regression loss~(\ie,~using L2 distances between predicted and ground-truth masks). Since our model predicts source and target masks independently, and uses synthetic pairs generated from a single image for training, mask regression layers output binary masks for all foreground objects at test time. To convert them into co-segmentation masks, we simply select the regions consistent with both predicted and warped masks,~\ie,~$\tilde{M}^s \odot \mathcal{W}(\tilde{M}^t;\mathcal{F}^s)$ in case of predicting a co-segmentation mask for the source image~(see Fig.~\ref{fig:coseg}), where we denote by $\tilde{M}^s$ and $\tilde{M}^t$ predicted masks for source and target images, respectively. Table~\ref{tab:coseg} compares IoU scores of co-segmentation masks on the TSS dataset. We can see that our model outperforms other methods using hand-crafted features, suggesting that it is feasible to leverage our model for object co-segmentation.
\vspace{0.07cm}

\begin{table}[t]
    \caption{Quantitative comparison of object co-segmentation on TSS~\cite{taniai2016joint} in terms of IoU. All numbers except ours are taken from~\cite{taniai2016joint}.}
    \vspace{-0.15cm}
    \small
    \begin{center}
        \setlength{\tabcolsep}{0.3em}
        \begin{tabular}{L{3.2cm}@{\hspace{1mm}} | @{}C{1.15cm}@{} | @{}C{1.15cm}@{} | @{}C{1.15cm}@{} | @{}C{1.15cm}@{}}
        \multicolumn{1}{c|}{Methods}  & FG3D. & JODS & PASC. & Avg.  \\ 
        \cmidrule{1-5}\morecmidrules
        \hline
        SIFT Flow$_\textrm{SIFT}$~\cite{liu2011sift} & 0.42 & 0.24 & 0.41 & 0.36 \\
        DSP$_\textrm{SIFT}$~\cite{kim2013deformable} & 0.29 & 0.22 & 0.34 & 0.28 \\
        DFF$_\textrm{DAISY}$~\cite{yang2014daisy} & 0.33 & 0.21 & 0.21 & 0.25\\
        Faktor and Irani$_\textrm{HOG}$~\cite{faktor2013co} & 0.69 & \underline{0.54} & 0.50 & 0.58\\
        Joulin et al.$_\textrm{SIFT}$~\cite{joulin2010discriminative} & 0.46 & 0.32 & 0.40 & 0.39\\
        TSS$_\textrm{HOG}$~\cite{taniai2016joint} & \underline{0.76} & 0.50 & \underline{0.65} & \underline{0.63}\\
        Ours$_\textrm{res101}$ & \bf{0.88} & \bf{0.78} & \bf{0.76} & \bf{0.81}\\
        % Hati \etal~[A] & 0.47 & 0.31 & 0.31 & 0.36\\
        % Chang \etal~[B] & 0.67 & 0.52 & 0.40 & 0.53\\
        % MRW~[C] & 0.63 & 0.46 & 0.66 & 0.58\\
        % Jerripothula \etal~[D] & 0.76 & 0.57 & 0.56 & 0.63\\
        % Jerripothula \etal~[A] & 0.78 & 0.65 & 0.73 & 0.72\\
        % Faktor \etal~[F] & 0.69 & 0.54 & 0.50 & 0.58\\
        % Joulin \etal~[G] & 0.46 & 0.32 & 0.40 & 0.39\\
        % Chen \etal~[B] & \bf{0.90} & \underline{0.77} & \bf{0.86} & \bf{0.84}\\
        \hline
        \end{tabular}
        \label{tab:coseg}	
    \end{center}
    \vspace{-0.3cm}
\end{table}

\subsection{Ablation Study and Effect of Training Data} \label{sec:ablation}
We show an ablation analysis on different components and losses in our model. We measure a PCK score~($\alpha_{\mathrm{bbox}}=0.1$), which is a more strict metric compared to $\alpha_{\mathrm{img}}$, and report the results of semantic correspondence on the test split of PF-PASCAL~\cite{ham2017proposal,rocco2018end}. We also study the effect of using different training datasets on performance.

\subsubsection{Training Loss}
We show the average PCK for three variants of our model in Table~\ref{tab:ablation_loss}. The mask consistency term encourages establishing correspondences between prominent objects. Our model trained with this term only, however, may not yield spatially distinctive correspondences, resulting in the worst performance. The flow consistency term, which spreads flow fields over foreground regions, overcomes this problem, but it does not differentiate correspondences between background and objects. Accordingly, these two terms are complementary each other and exploiting both significantly boosts the performance of our model from 67.5/71.8 to 78.2. An additional smoothness term further boosts performance to 78.7. 
%\todo{already outperforming the state of the art by a large margin~(see Table~\ref{tab:pf_data_bb}).}

\subsubsection{Network Architecture}
\begin{figure}[t]
	\captionsetup[subfigure]{aboveskip=-0.5pt,belowskip=-0.5pt}
	\centering
	\begin{tabular}{cc}
		\begin{tabular}{c}
			\begin{subfigure}{.2\linewidth}
			\includegraphics[width=\linewidth, frame]{./fig/ablation/source_no_kp.png}
			\vspace{-0.25cm}
			\caption*{Source.}
			\vspace{0.5cm}
			\end{subfigure} \\
			\begin{subfigure}{.2\linewidth}
			\includegraphics[width=\linewidth, frame]{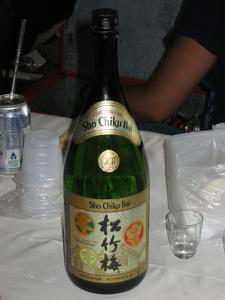}
			\vspace{-0.25cm}
			\caption*{Target.}
			\end{subfigure} \\
		\end{tabular}
		\vline
		\begin{tabular}{cccc}
			\rotatebox[origin=c]{90}{w/o adaptation.} &
			\adjustbox{valign=m,vspace=0.5pt}{\includegraphics[width=.2\linewidth, frame]{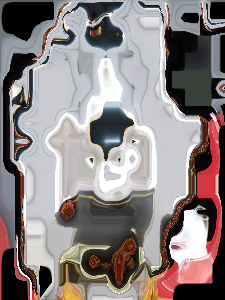}} &
			\adjustbox{valign=m,vspace=0.5pt}{\includegraphics[width=.2\linewidth, frame]{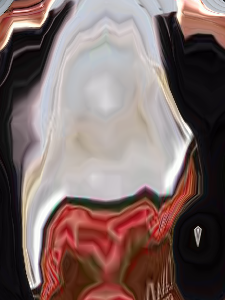}} &
			\adjustbox{valign=m,vspace=0.5pt}{\includegraphics[width=.2\linewidth, frame]{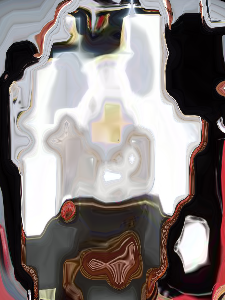}} \\
			
			\rotatebox[origin=c]{90}{w/o multi-level.} & 
			\adjustbox{valign=m,vspace=0.5pt}{\includegraphics[width=.2\linewidth, frame]{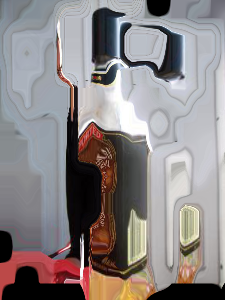}} & 
			\adjustbox{valign=m,vspace=0.5pt}{\includegraphics[width=.2\linewidth, frame]{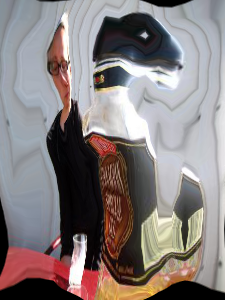}} &
			\adjustbox{valign=m,vspace=0.5pt}{\includegraphics[width=.2\linewidth, frame]{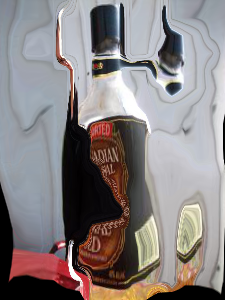}} \\
			
			\rotatebox[origin=c]{90}{All.} & 
			\adjustbox{valign=m,vspace=0.5pt}{\includegraphics[width=.2\linewidth, frame]{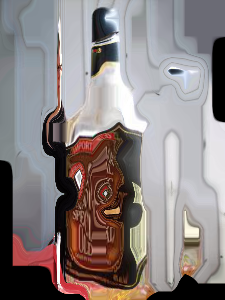}} & 
			\adjustbox{valign=m,vspace=0.5pt}{\includegraphics[width=.2\linewidth, frame]{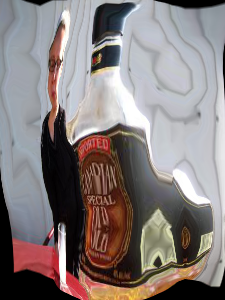}} & 
			\adjustbox{valign=m,vspace=0.5pt}{\includegraphics[width=.2\linewidth, frame]{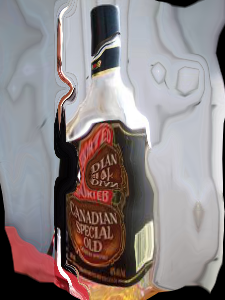}} \\
			
				& Discrete. & Soft. & Kernel soft.
		\end{tabular}
	\end{tabular}
	\vspace{-0.2cm}
	\caption{Visual comparison of different network architectures and argmax operators. We show alignment examples for the variants of our model in Table~\ref{tab:ablation_network}.}
	\label{fig:qualitative_ablation}	
\end{figure}

Table~\ref{tab:ablation_network} compares the performance of networks with different components in terms of average PCK. The baseline models in the first three rows compute matching scores using multi-level features from~$\texttt{conv4-23}$ and $\texttt{conv5-3}$ layers, and estimate correspondences with different argmax operators. They do not involve any training similar to~\cite{long2014convnets} that uses off-the-shelf CNN features for semantic correspondence. We can see that applying the soft argmax directly to the baseline model degrades performance severely, since it is highly susceptible to multi-modal distributions. The results in the next three rows are obtained with a single adaptation layer on top of $\texttt{conv4-23}$. This demonstrates that the adaptation layer extracts features more adequate for pixel-wise semantic correspondences, boosting performance of all baseline models significantly. In particular, we can see that the kernel soft argmax outperforms others by a large margin, since it enables training our model end to end including adaptation layers at a sub-pixel level and is less susceptible to multi-modal distributions. The last three rows suggest that exploiting deeper features is important, and using all components with the kernel soft argmax performs best in terms of the average PCK. We show in Fig.~\ref{fig:qualitative_ablation} alignment examples for the variants of our model in Table~\ref{tab:ablation_network}. This confirms once more the results in Table~\ref{tab:ablation_network} that the adaptation layers and exploiting multi-level features boost the matching performance drastically, regardless of types of argmax operators, and the soft argmax is highly susceptible to multi-modal distributions, \eg,~caused by ambiguous matches between a bottle and a glass in the source and target images, respectively. 

\begin{table}[t]
	%\vspace{-2mm}
	\caption{{Average PCK comparison of different loss functions.}}
	%\vspace{-0.3cm}
	\small
	\begin{center}
	\begin{tabular}{C{1.7cm} C{1.7cm} C{1.7cm} |C{2cm}}
					
	Mask 			& Flow			 	& \multirow{2}{*}{Smoothness}	& PCK	\\
	consistency 	& consistency 		& 				   			 	& ($\alpha_{\mathrm{bbox}}=0.1$)  \\ 
	\cmidrule{1-4}\morecmidrules\hline
	&&& \\[-0.9em]
	\cmark		&	\xmark		&	\xmark		&	67.5 				\\
	\xmark		&	\cmark		&	\xmark		&	71.8				\\
	\cmark		&	\cmark		&	\xmark		&	78.2			 	\\
	\cmark		&	\cmark		&	\cmark		&	\textbf{78.7} 		\\
	\hline
	\end{tabular}
	\end{center}
	\label{tab:ablation_loss}
	\end{table}
	
	\begin{table}[t]
		%\vspace{-2mm}
		\caption{Average PCK comparison of variants of our model. We denote by ``D'', ``S'', and ``KS'' discrete, soft, and kernel soft argmax operators, respectively.}
		%\vspace{-0.4cm}
		\small
		\begin{center}
		\begin{tabular}{C{1.45cm} C{1.45cm} C{0.9cm} C{0.9cm} |C{2cm}}
						
		Adaptation 	& Multi-level 	& \multicolumn{2}{c|}{Argmax}  			&PCK	\\
		layer 		& feature 		& \multicolumn{1}{c|}{Train} & Test     & ($\alpha_{\mathrm{bbox}}=0.1$)	\\ 
		\cmidrule{1-5}\morecmidrules\hline
		&&&& \\[-0.9em]
		\xmark	&	\cmark	&	-		&	D		&	45.8	 		\\
		\xmark	&	\cmark	&	-		&	S		&	8.8    		\\
		\xmark	&	\cmark	&	-		&	KS		&	28.4			\\
		\hline
		
		\cmark	&	\xmark	&	S		&	D		&	72.5			\\
		\cmark	&	\xmark	&	S		&	S		&	71.7			\\
		\cmark	&	\xmark	&	KS		&	KS		&	75.0			\\
		
		\hline
		
		\cmark	&	\cmark	&	S		&	D		&	76.8			\\
		\cmark	&	\cmark	&	S		&	S		&	76.2			\\
		\cmark	&	\cmark	&	KS		&	KS		&	\textbf{78.7}	\\
		\hline
		\end{tabular}
		\end{center}
		\label{tab:ablation_network}
		\vspace{-0.3cm}
\end{table}

%\todo{Alignment examples in Fig.~\ref{fig:qualitative_ablation} confirm the importance of each component. Without adaptation layers~(results in the first row), directly assigning the best matches from a correlation map in~\eqref{eq:correlation_map} does not produce meaningful correspondences. Training adaptation layers improves the quality of correspondences, but the models using a single adaptation layer~(results in the second row) have difficulty of disambiguating matches between nearby objects belonging to different categories~(\eg, person and bottle). We can also see that the discrete argmax causes abrupt changes in flow fields and the soft argmax is highly affected by distractions~(\eg, a glass in the target image), mainly due to the vulnerability to multi-modal distributions. Kernel soft argmax, on the other hand, gives a good compromise between sub-pixel accuracy and localization precision, leading to the best result in case of using adaptation layers on top of multi-level features.}

%\caption{\todo{Alignment examples of ablation studies. In left, we show source and target images. In right, we show alignment results where the source image is warped to the target image using flow fields obtained by models with different components. Note that the models correspond to those in Table~\ref{tab:ablation_network}. We represent types of network architecture and argmax operators used for inference in the left and bottom sides, respectively.}}

\subsubsection{Training with Bounding Boxes}\label{sec:train_bbox}
We train our model using object bounding boxes themselves as binary masks. The generated masks are noisy, but are less expensive to annotate than ground-truth foreground masks. We use the same 2,791 images from the PASCAL VOC 2012 segmentation dataset~\cite{everingham2010pascal} for training, and obtain an average PCK~($\alpha_{\mathrm{bbox}}=0.1$) of $77.9$ on the PF-PASCAL dataset~\cite{ham2017proposal}, which is comparable with the score of $78.7$ using ground-truth masks. This suggests that using bounding boxes might be a less accurate but cheaper alternative.

%\todo{Using flow consistency terms of (8) w.r.t both source and target images penalizes matches between background and foreground regions, making our method robust to noisy labels.} 

\subsubsection{Training on PF-PASCAL}
Semantic correspondence methods based on CNNs use different training sets. For example, the methods of~\cite{rocco2017convolutional,rocco2018end} use the PASCAL VOC 2011~(11,540 images) and Tokyo Time Machine datasets~(20,000 images). In~\cite{han2017scnet,rocco2018end,Jeon2018PARN,kim2018recurrent,rocco2018neighbourhood}, the training split of PF-PASCAL~\cite{ham2017proposal}~(about 700 image pairs for 1,001 images) is used. Following these approaches, we train a network on the training split in the PF-PASCAL dataset. We exclude 302 images in this split that overlap with either target or source images in the test dataset. Note that current methods ignore this bias. For example, removing the bias reduces a PCK score for the WS-SA~\cite{rocco2018end} from 75.8 to 75.67 on the test split of PF-PASCAL. We use object bounding boxes due to the lack of ground-truth foreground masks in the training split. We obtain the average PCK~($\alpha_{\mathrm{bbox}}=0.1$) of 77.8 on PF-PASCAL, which is comparable with the score of 78.7 for the model trained using 1,464 images on the PASCAL VOC 2012 segmentation dataset. This indicates that our model is robust to the size of training data.

% still outperforms the semantic alignment methods in Table~\ref{tab:pf_data_bb}~(\eg,~75.8~for~\cite{rocco2018end} and 71.9~for~\cite{rocco2017convolutional} measured with $\alpha_{\mathrm{img}}=0.1$). \todo{describe the conclusion of this exp.}

%CNN-based methods use different training sets,~\eg,~the training split of PF-PASCAL~\cite{ham2017proposal}~(about 700 image pairs) for~\cite{han2017scnet,rocco2018end,Jeon2018PARN,kim2018recurrent,rocco2018neighbourhood} and PASCAL VOC 2011~(11,540 images) for~\cite{rocco2017convolutional,rocco2018end,Seo2018AttentiveSA}. In~\cite{rocco2017convolutional,rocco2018end}, the Tokyo Time Machine dataset~(20,000 images) is also used. For fair comparison, we train a network on \changed{the smallest training dataset (training split of PF-PASCAL~\cite{ham2017proposal})}. We excluded 302 images in this split that overlap with either target or source images in the test dataset. Note that the methods of~\cite{han2017scnet,rocco2018end} ignore this bias. We use object bounding boxes due to the lack of ground-truth foreground masks in the training split. \changed{Even in the fair comparison, we obtain the average PCK~($\alpha_{\mathrm{bbox}}=0.1$) of 77.8 on PF-PASCAL, which still outperforms the semantic alignment methods in Table~\ref{tab:pf_data_bb}~(\eg,~75.8~for~\cite{rocco2018end} and 71.9~for~\cite{rocco2017convolutional} measured with $\alpha_{\mathrm{img}}=0.1$).}

\subsubsection{Training on Larger Datasets}
We use the training split of MS COCO 2014~\cite{lin2014microsoft} to train our model on a larger dataset. Among 80 object categories, we select 16,624 images of 20 object classes of PASCAL VOC 2012~\cite{everingham2010pascal} using segmentation masks, which is about 6 times the number used in Section~\ref{sec:semantic_correspondence_results}~(2,791 images). We test our model on the PF-PASCAL dataset~\cite{ham2017proposal}, since MS COCO does not provide benchmarks for semantic correspondence. Despite using a larger number of training samples, the average PCK~($\alpha_{\mathrm{bbox}}=0.1$) decreases slightly from $78.7$ to $77.1$, mainly due to domain differences between MS COCO and PASCAL VOC datasets. This, however, demonstrates once more the generalization ability of our approach to samples outside the training domain.

\section{Conclusion}
\label{sec:conclusion}
We have presented a CNN model dubbed SFNet for learning an object-aware semantic flow end to end, with a novel kernel soft argmax layer that outputs differentiable matches at a sub-pixel level. We have proposed to use binary foreground masks that are widely available and can be obtained easily compared to pixel-level annotations to train a model for learning pixel-to-pixel correspondences. The ablation studies clearly demonstrate the effectiveness of each component and loss in our model. Finally, we have shown that the proposed method is robust to distracting details and focuses on establishing dense correspondences between prominent objects, outperforming the state of the art on standard benchmarks for the tasks of semantic correspondence, mask transfer, pose keypoint propagation and object co-segmentation in terms of accuracy and speed. In future work, we will explore a joint learning model for semantic correspondence and object co-segmentation that can assist each other.

\ifCLASSOPTIONcompsoc
  \section*{Acknowledgments}
\else
  \section*{Acknowledgment}
\fi
This work was supported by the National Research Foundation of Korea (NRF) and Institute of Information \& communications Technology Planning \& Evaluation (IITP) grants funded by the Korea government (MSIT) (NRF-2019R1A2C2084816, 2016-0-00197, Development of the high-precision natural 3D view generation technology using smart-car multi sensors and deep learning), and also in part by the Louis Vuitton/ENS chair on artificial intelligence, the Inria/NYU collaboration agreement, and the French government under management of Agence Nationale de la Recherche as part of the "Investissements d'avenir" program, reference ANR-19-P3IA-0001 (PRAIRIE 3IA Institute).

\ifCLASSOPTIONcaptionsoff
  \newpage
\fi



\bibliographystyle{IEEEtran}
\bibliography{SFNet}
%\begin{thebibliography}{1}
%\end{thebibliography}

\begin{IEEEbiography}[{\includegraphics[width=1in,height=1.25in,clip,keepaspectratio]{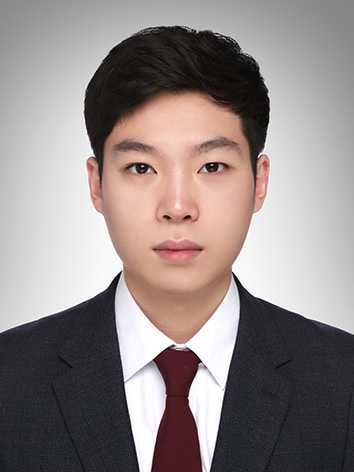}}]{Junghyup Lee}
received the B.S. degree in Electrical and Electronic Engineering from Yonsei University in 2018. He is currently pursuing a Ph.D. degree. He has received the Global PhD Fellowship from the National Research Foundation of Korea (NRF) in 2019. His current current research is focused on computer vision and machine learning, including image matching, super-resolution, and network quantization.
\end{IEEEbiography}

\begin{IEEEbiography}[{\includegraphics[width=1in,height=1.25in,clip,keepaspectratio]{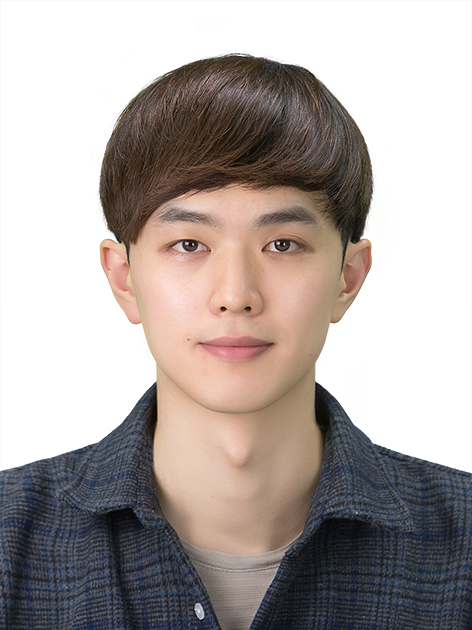}}]{Dohyung Kim}
received the B.S. degree in Electrical and Electronic Engineering from Yonsei University in 2018 where he is currently pursuing the Ph.D. degree. His research interests lie in computer vision, with a focus on image matching and super resolution. He is also interested in an efficient network design using distillation and quantization techniques.
\end{IEEEbiography}

\begin{IEEEbiography}[{\includegraphics[width=1in,height=1.25in,clip,keepaspectratio]{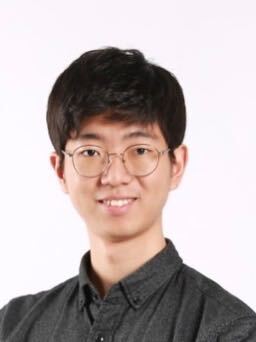}}]{Wonkyung Lee}
received the BE degree in Electrical and Electronic Engineering from Yonsei University, Seoul, South Korea, in 2019. He is currently working toward the PhD degree in Electrical and Electronic Engineering department from Yonsei University. His recent research focuses on building efficient deep learning models for visual tasks. He is also interested in developing alternative self-attention mechanisms for visual recognition.
\end{IEEEbiography}

\begin{IEEEbiography}[{\includegraphics[width=1in,height=1.25in,clip]{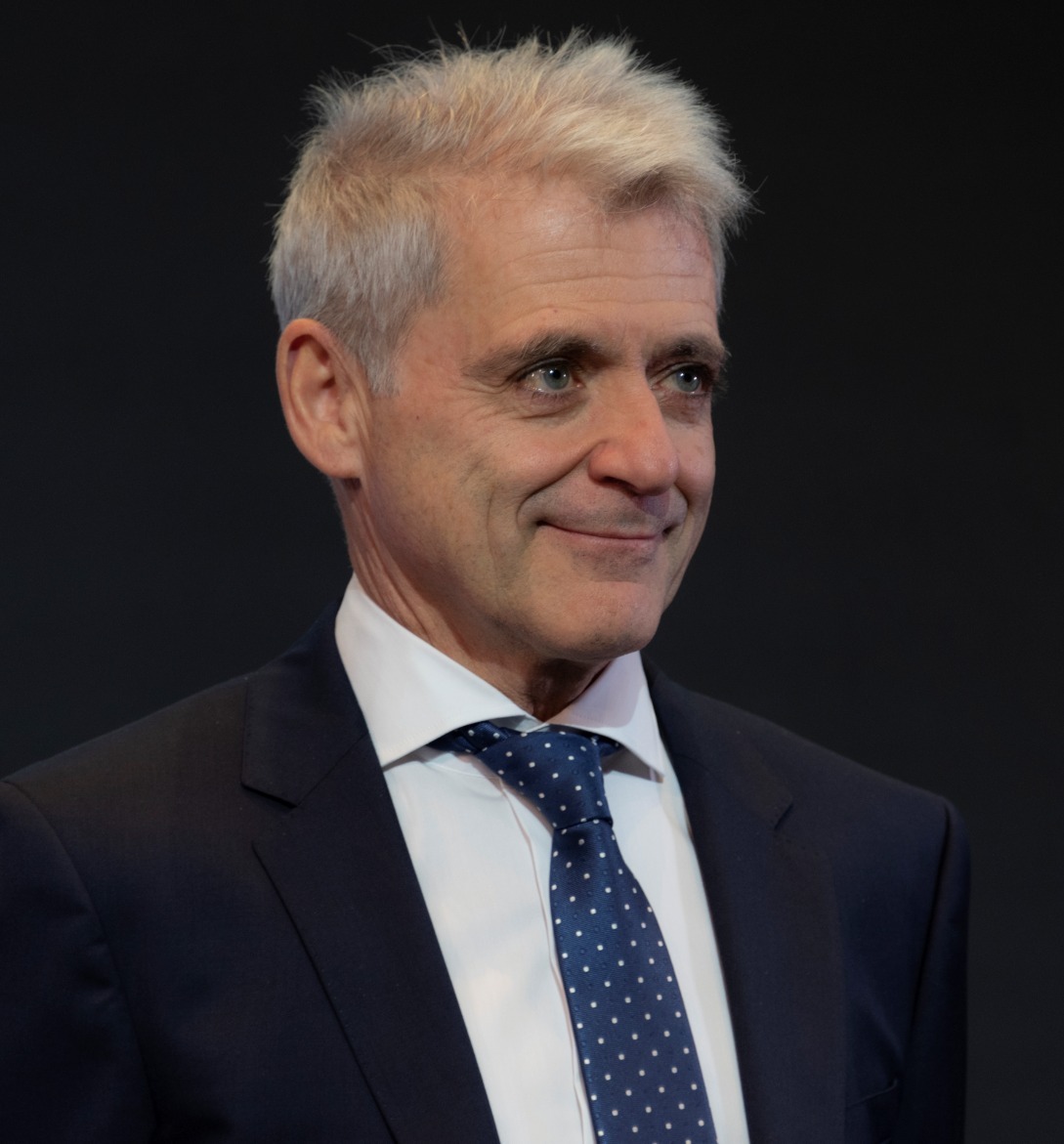}}]{Jean Ponce}
 is a research director at Inria and a visiting researcher
at the NYU Center for Data Science, on leave from Ecole Normale
Superieure (ENS) / PSL Research University, where he is a professor,
and served as director of the computer science department from 2011 to
2017. Before joining ENS and Inria, Jean Ponce held positions at MIT,
Stanford, and the University of Illinois at Urbana-Champaign, where he
was a full professor until 2005. Jean Ponce is an IEEE Fellow and a
Sr. member of the Institut Universitaire de France. He chaired the
IEEE Conference on Computer Vision and Pattern Recognition in 1997 and
2000, and the European Conference on Computer Vision in 2008.  He
currently serves as Sr. Editor-in-Chief of the International
Journal of Computer Vision, and as scientific director of the PRAIRIE
Interdisciplinary AI Research Institute in Paris. Jean Ponce is the
recipient of two US patents, an ERC advanced grant, the 2016 and 2020 IEEE CVPR
Longuet-Higgins prizes, and the 2019 ICML test-of-time award. He is the
author of "Computer Vision: A Modern Approach", a textbook translated
in Chinese, Japanese, and Russian.
\end{IEEEbiography}

\begin{IEEEbiography}[{\includegraphics[width=1in,height=1.25in]{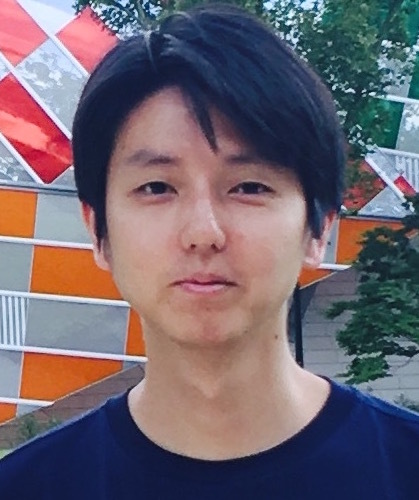}}]{Bumsub Ham} 
is an an Assistant Professor of Electrical and Electronic Engineering at Yonsei University in Seoul, Korea. He received the B.S. and Ph.D. degrees in Electrical and Electronic Engineering from Yonsei University in 2008 and 2013, respectively. From 2014 to 2016, he was Post-Doctoral Research Fellow with Willow Team of INRIA Rocquencourt, {\'E}cole Normale Sup{\'e}rieure de Paris, and Centre National de la Recherche Scientifique. His research interests include computer vision, computational photography, and machine learning, in particular, regularization and  matching, both in theory and applications.
\end{IEEEbiography}

%\begin{IEEEbiography}{John Doe}

%\newpage

%\begin{IEEEbiography}{Jane Doe}


%\vfill


\end{document}